\documentclass[10pt,twocolumn,letterpaper]{article}

\usepackage{iccv}
\usepackage{times}
\usepackage{epsfig}
\usepackage{graphicx}
\usepackage{amsmath}
\usepackage{amssymb}
\usepackage{booktabs}
\usepackage[dvipsnames,table,xcdraw]{xcolor}
\usepackage[utf8]{inputenc} %
\usepackage[T1]{fontenc}    %
\usepackage{url}            %
\usepackage{amsfonts}       %
\usepackage{nicefrac}       %
\usepackage{microtype}      %
\usepackage{multirow}
\usepackage{bbm}
\usepackage{rotating}
\usepackage[english]{babel}
\usepackage{pifont}%

\usepackage[pagebackref=true,breaklinks=true,letterpaper=true,colorlinks,bookmarks=false]{hyperref}

\definecolor{green}{RGB}{0,150,10}

\newcommand{\figLabel}{Figure~}
\newcommand{\eqLabel}[1]{{Eq (#1)}}
\newcommand{\secLabel}{Section~}

\newcommand{\mysection}[1]{\noindent\textbf{#1.}}
\newcommand{\supp}{supplementary material\xspace}

\newcommand{\sota}{state-of-the-art\xspace}

\iccvfinalcopy %

\def\iccvPaperID{***} %

\ificcvfinal\fi

\begin{document}

\title{SPARF: Large-Scale Learning of 3D Sparse Radiance Fields from Few Input Images}

\author{Abdullah Hamdi $^{1,2,3}$ \quad\quad Bernard Ghanem $^2$ \quad\quad Matthias Nie{\ss}ner $^1$\\  
\normalsize$^1$Technical University of Munich (TUM) \hspace{5pt} \normalsize$^2$King Abdullah University of Science and Technology (KAUST)\\
\normalsize$^3$Visual Geometry Group, University of Oxford \\
\tt\small{abdullah.hamdi@eng.ox.ac.uk}
}

\maketitle
\ificcvfinal\thispagestyle{empty}\fi

\begin{abstract}
Recent advances in Neural Radiance Fields (NeRFs) treat the problem of novel view synthesis as Sparse Radiance Field (SRF) optimization using sparse voxels for efficient and fast rendering \cite{plenoxels,instantneural}. In order to leverage machine learning and adoption of SRFs as a 3D representation, we present \textit{SPARF}, a large-scale ShapeNet-based synthetic dataset for novel view synthesis consisting of $\sim$ 17 million images rendered from nearly 40,000 shapes at high resolution (400 $\times$ 400 pixels). The dataset is orders of magnitude larger than existing synthetic datasets for novel view synthesis and includes more than one million 3D-optimized radiance fields with multiple voxel resolutions. Furthermore, we propose a novel pipeline (\textit{SuRFNet}) that learns to generate sparse voxel radiance fields from only few views. This is done by using the densely collected SPARF dataset and 3D sparse convolutions. SuRFNet employs partial SRFs from few/one images and a specialized SRF loss to learn to generate high-quality sparse voxel radiance fields that can be rendered from novel views. Our approach achieves \sota results in the task of unconstrained novel view synthesis based on few views on ShapeNet as compared to recent baselines. 
The SPARF dataset is made public with the code and models on the project website \href{https://abdullahamdi.com/sparf/}{abdullahamdi.com/sparf}.
\end{abstract}
\linespread{0.98}
\section{Introduction} \label{sec:introduction}
\vspace{-4pt}
Although we observe the surrounding world only as a stream of 2D images, it is undeniably 3D.
The goal of recovering this underlying 3D from 2D observations has been a longstanding goal of computer vision. The task of inverting the rendering process that creates the 2D projections we observe by trying to construct the 3D world is known as Vision as Inverse Graphics (VIG) \cite{vig-bid-r,vig-cinvg,vig-inverse-render-net,vig-reinforce}. With the emergence of deep learning applications in computer graphics and the availability of 3D datasets, several approaches address the 3D generation task directly from 3D data, without relying on appearance \cite{polygen,meshconv,arapreg,pc-ae,foldingnet,spgan}. However, recent developments in differentiable rendering have refueled the VIG direction, which facilitates using gradients of the rendering process to optimize for the underlying 3D setup based on image observations \cite{pytorch3d,pixel2mesh,meshrcnn,ners,mvtn,hamdi2023voint,difstereopsis,text2mesh,nerf,neuralvolumes,RealFusion,3DFuse,Zero-1-to-3}. 
More specifically, Neural Radiance Fields (NeRFs) \cite{nerf,dnerf,pixelnerf,mvsnerf} show impressive performance on novel view synthesis by optimizing volumetric radiance fields on a large number of posed multi-view images. 

\begin{figure}
    \centering
    \includegraphics[trim= 1.4cm 0.0cm 0.2cm 0cm,clip, width=0.98\linewidth]{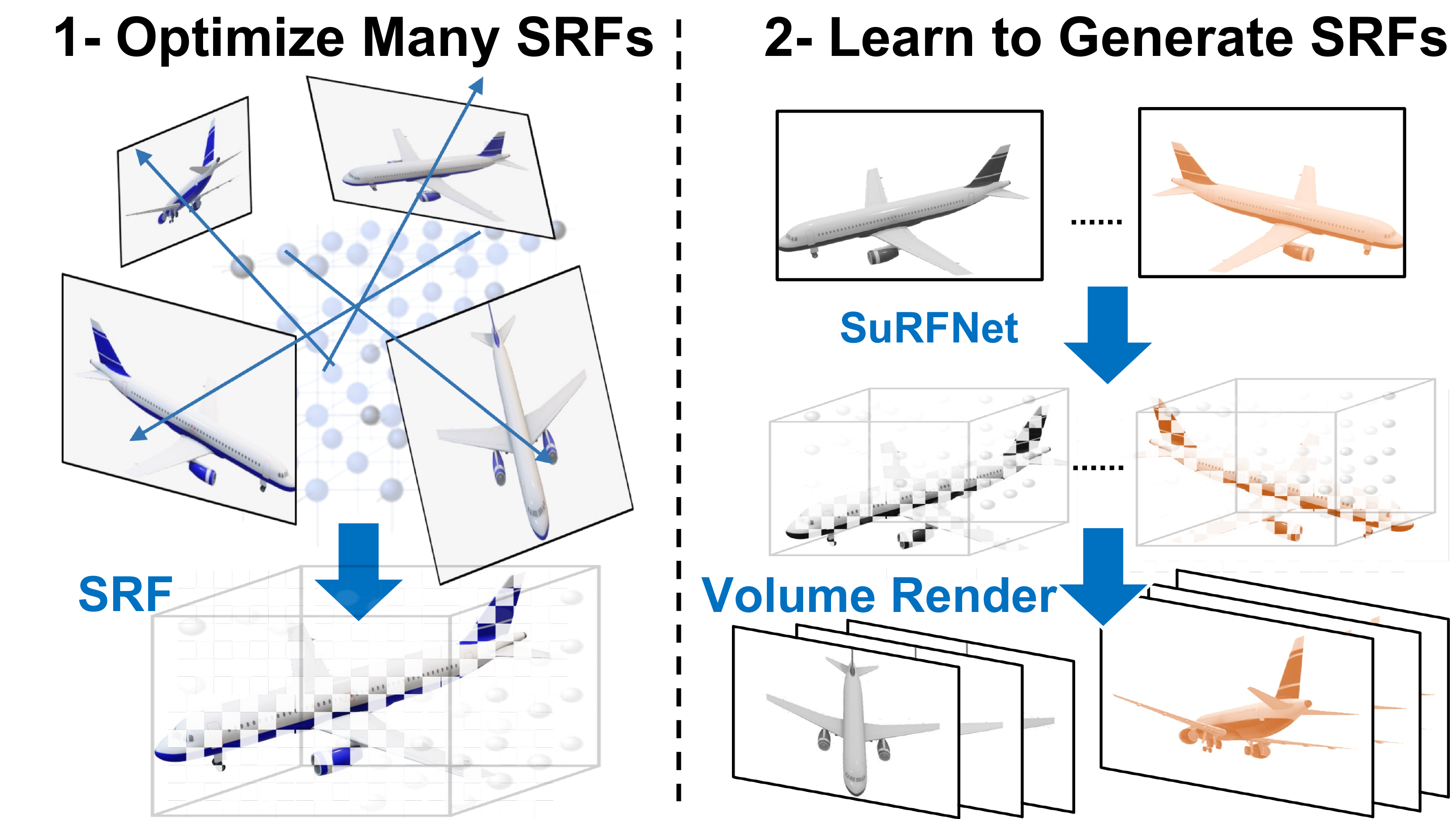}
    \vspace{2pt}
    \caption{\textbf{Distribution of Radiance Fields.} We treat Sparse Radiance Field (SRF) as a 3D data structure and learn the conditional generation of SRFs from few input images for the task of novel view synthesis. In order to do this, we build SPARF, a large-scale dataset of SRFs.
    }
    \label{fig:pullfigure}
\end{figure}
\begin{table*}[t]
\centering
\resizebox{0.88\linewidth}{!}{
\tabcolsep=0.12cm
\begin{tabular}{l|ccccc}
\toprule
 & \multicolumn{5}{c}{\textbf{Posed Multi-View Datasets} } \\
\textbf{Attribute} & \textbf{SRN} \cite{srn} & \textbf{DTU}\cite{dtu} & \textbf{NMR} \cite{dvr}                       & \textbf{RTMV} \cite{rtmv} & ~~\textbf{SPARF (ours)} \\

\midrule
Number of Classes & 2 & N/A   & 13 & N/A             & 13  \\
\rowcolor[HTML]{EFEFEF} 
Number of Scenes/Objects & 3,511  & 124  & 43,756 & 2,000 & 39,705 \\
Image Resolution & 128 & 1,200   & 64 & 1,600  & 400                          \\
\rowcolor[HTML]{EFEFEF} 
Number of Radiance Fields &   0      &   0   &   0     & 2,000  & \textbf{1,072,008}  \\
Real/Synthetic & synthetic & real  & synthetic & synthetic          & synthetic \\
\rowcolor[HTML]{EFEFEF} 
View Setup & sphere  & random  & circle & hemisphere & sphere   \\
Total Number of Images & 265,550           & 4,235   & 1,050,144 & 300,000 & \textbf{17,073,150} \\
\rowcolor[HTML]{EFEFEF} 
Views per Model & 50 & N/A & 24 & N/A & 430 \\
Dataset Size (GB) & 5.8 & 1 & 33 & 2,520 & 3,432 \\
\bottomrule
\end{tabular}
}
\vspace{3pt}
\caption{\small \textbf{Comparison of Different Posed Multi-View Datasets}. We compare posed multi-view datasets to our large-scale SPARF dataset.}
\vspace{-4mm}
    \label{tbl:mvdatasets}
\end{table*}
Various subsequent works addressed NeRFs' shortcomings, as rendering speed \cite{instantneural,plenoxels,plennerf}, training size requirement \cite{pixelnerf,mvsnerf,autorf}, or pose requirements \cite{dnerf,wnerf}. The seminal work of Plenoxels \cite{plenoxels}  showed that the MLP network is not necessary for quick optimization and volumetric rendering of the radiance fields. However, these recent methods are still optimization-based, where a single scene representation is optimized without any generalization to new scenes/objects \cite{instantneural,plenoxels}. In this work, we treat Sparse Radiance Fields (SRFs) as a 3D data structure and try to learn a generative model (dubbed \textit{SuRFNet}) on the distribution of sparse-voxel radiance fields conditioned on a few images to generalize to unseen 3D shapes (see \figLabel{\ref{fig:pullfigure}}).

In order to train deep learning models on 3D data to generalize to unseen examples, the dataset size should be in tens of thousands of samples \cite{modelnet,shapenet}. However, current posed multi-view datasets are not suitable for leveraging the power of deep networks as can be seen in Table \ref{tbl:mvdatasets}. The image resolution is either too low (\eg $64\times 64$ in \cite{dvr}), the samples don't fall into distinct shape categories with similar numbers of views \cite{dtu,co3d,jeong2022perfception}, or lack diversity in the samples and classes \cite{srn}.  For these reasons, we construct a large and high-resolution dataset (SPARF) of posed multi-view images from ShapeNet \cite{shapenet} that correspond to the same 13 classes originally used in the NMR dataset \cite{dvr}, but with an order of magnitude more images and pixels (17M \vs 1M images and 400$\times$400 \vs 64 $\times$ 64 pixels). We also provide more than \textit{one million} optimized sparse radiance fields of spherical harmonics and densities that allow for the novel view synthesis of the 40K models using Plenoxels \cite{plenoxels}.

The idea of learning a prior (2D CNN/ViT) on radiance fields in order to enhance the few-view setup of novel view synthesis is previously investigated by several works \cite{pixelnerf,sharf,visionnerf}. However, we propose SuRFNet to \textit{directly} learn from the 3D sparse radiance fields, by optimizing \textit{partial} SRFs from the few images and training a generalizable network that converts these partial SRFs to \textit{whole} SRFs in a supervised fashion. Such a 3D setup benefits from grid-based 3D learning, creating a 3D prior that ensures multi-view consistency, especially when rendering from out-of-distribution views. Also, this 3D sparse voxel setup benefits from the advancements in fast volume rendering \cite{instantneural,plenoxels}, allowing for end-to-end deep learning pipelines that harness volume rendering. 
To the best of our knowledge, our SURFNet is the first model that learns to generate 3D radiance fields for unseen objects at test time with only a few/single views by learning from the distribution of radiance fields in 3D. 

\noindent\textbf{Contributions:} 
\textbf{(i)} To facilitate the application of deep learning on radiance fields, we provide a new Posed Multi-view dataset (SPARF) that is an order of magnitude larger than others (around 40K 3D models). The dataset includes a total of one million optimized Sparse Radiance Fields (SRFs) with multiple voxel resolutions, which allows for high-quality novel view synthesis and will be made publicly available.
\textbf{(ii)} We propose a novel architecture and a pipeline (SuRFNet) equipped with a specialized SRF-loss to generate voxel-based radiance fields from a few images based on learning to complete partial radiance fields. SuRFNet improves the performance of unconstrained novel view synthesis based on few views compared to \sota methods. 

\section{Related Work} \label{sec:related}
\vspace{-4pt}
\mysection{Learning 3D Shapes}
Several works aim to predict the geometry of 3D shapes given several input images, by directly optimizing the vertices of a template mesh  through differentiable projections or through fitting a network \cite{pixel2mesh,meshrcnn,ners,difstereopsis,text2mesh,get3d,magic123}. 
Other works use MLPs as a deep prior to the optimized mesh \cite{point2mesh,deephybridmesh}. 
Alternately, some methods try to learn the distribution of 3D meshes by optimizing 3D generators independent of how the meshes look when rendered, solely based on the available 3D data and heuristic regularizers \cite{polygen,meshconv,scan2mesh}. Point cloud methods offer an alternative to the mesh complex topology  by learning generative models on the point clouds themselves, \eg by using an Auto Encoder \cite{pc-ae,foldingnet} or a GAN framework \cite{pc-ae,spgan}. The implicit representation paradigm offers an alternative to meshes for smooth and detailed shape representation. These methods learn a continuous implicit representation of shapes by learning the Signed Distance Functions or occupancy of the object through MLPs \cite{deepsdf,occupancynet,volsurf,girraffe,pigan,pix2nerf,voxgraf,neuralvolumes}. 
In this work, the scope focuses on the quality of the rendering from novel views and not on 3D reconstruction.

\mysection{Neural Radiance Fields (NeRFs)}
NeRFs \cite{nerf} proved to be a successful popularizing in implicit volume representation and novel view synthesis. They define an implicit field and learn an MLP that predicts the RGB and density value of that 3D field given a set of posed images. NeRFs shoot rays on the volume and integrate the predictions to obtain individual pixel values. This formulation, however, has many drawbacks including large memory and compute requirements, inability to model dynamic scenes, posed image requirements, and the limitation to small 3D objects or rooms \cite{wnerf,dnerf,plennerf,girraffe,pixelnerf,li2022compnvs}.
To address the speed limitation, PlenOctreeNeRF \cite{plennerf} stores the precomputed RGB, density, and spherical harmonics in the 3D volume as an Octree data structure for fast inference. Plenoxels \cite{plenoxels} optimize the density and spherical harmonics on sparse voxels with a TV loss and perform ray marching for rendering from novel views. Similarly, INGP \cite{instantneural} uses multi-resolution voxel hashing to perform a real-time rendering of radiance fields, demonstrating that the redundancy of the MLP in NeRFs. We build on these observations and build the SPARF dataset of sparse voxel radiance fields in order to facilitate learning on these SRFs as 3D data structures instead of as just side outcomes of a volumetric optimization.   
\begin{figure}
    \centering
    \includegraphics[trim= 0cm 0cm 0cm 0cm,clip, width=0.19\linewidth]{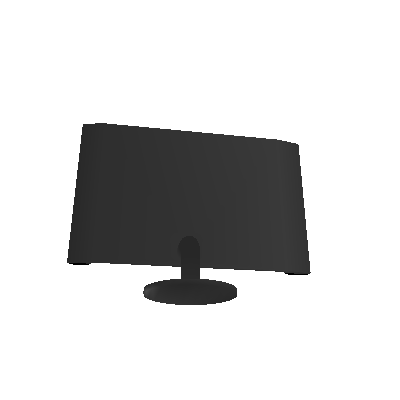}
        \includegraphics[trim= 0cm 0cm 0cm 0cm,clip, width=0.19\linewidth]{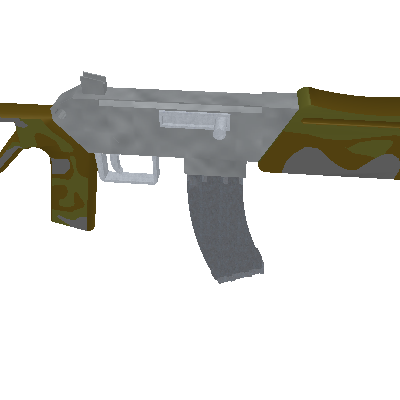}
    \includegraphics[trim= 0cm 0cm 0cm 0cm,clip, width=0.19\linewidth]{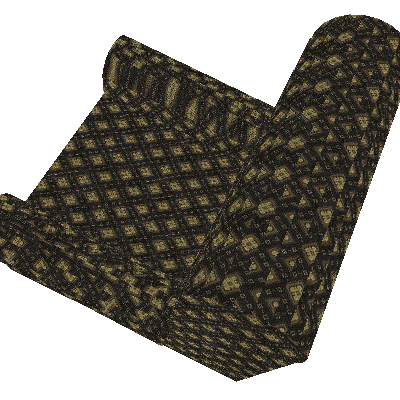}
    \includegraphics[trim= 0cm 0cm 0cm 0cm,clip, width=0.19\linewidth]{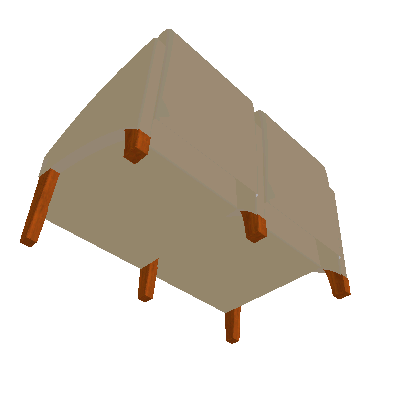}
    \includegraphics[trim= 0cm 0cm 0cm 0cm,clip, width=0.19\linewidth]{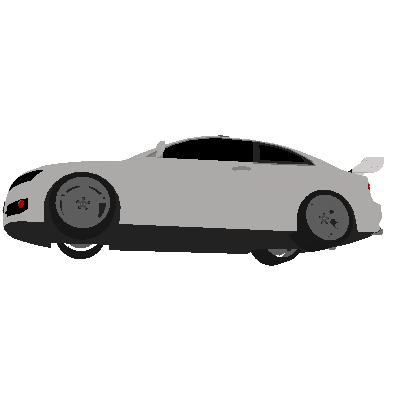}
\includegraphics[trim= 0cm 0cm 0cm 0cm,clip, width=0.19\linewidth]{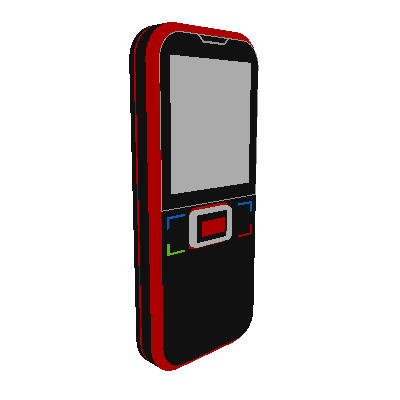}
    \includegraphics[trim= 0cm 0cm 0cm 0cm,clip, width=0.19\linewidth]{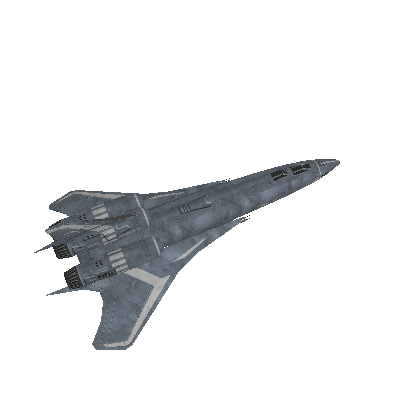}
        \includegraphics[trim= 0cm 0cm 0cm 0cm,clip, width=0.19\linewidth]{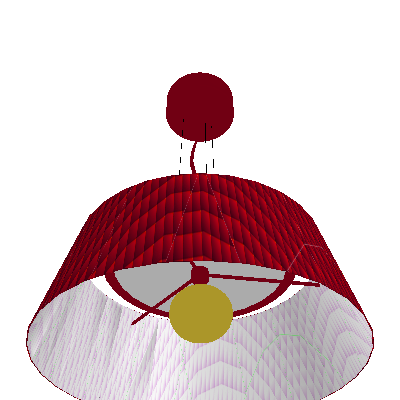}
    \includegraphics[trim= 0cm 0cm 0cm 0cm,clip, width=0.19\linewidth]{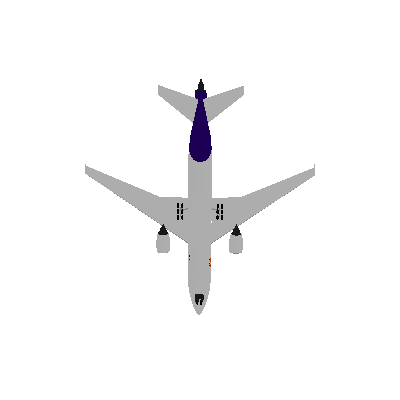}
    \includegraphics[trim= 0cm 0cm 0cm 0cm,clip, width=0.19\linewidth]{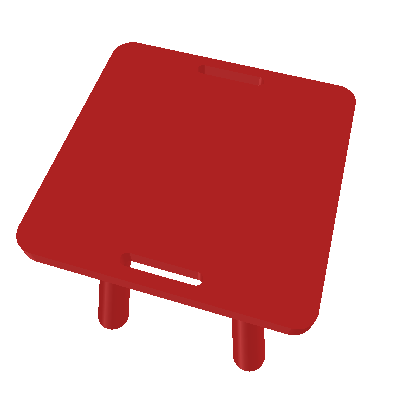}
    \includegraphics[trim= 0cm 0cm 0cm 0cm,clip, width=0.19\linewidth]{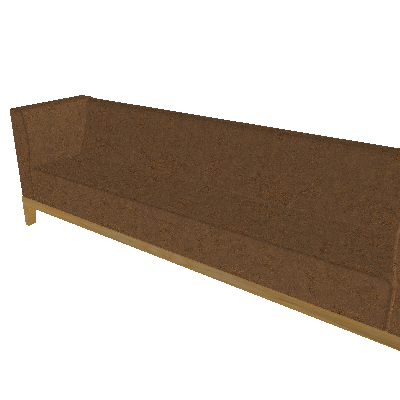}
        \includegraphics[trim= 0cm 0cm 0cm 0cm,clip, width=0.19\linewidth]{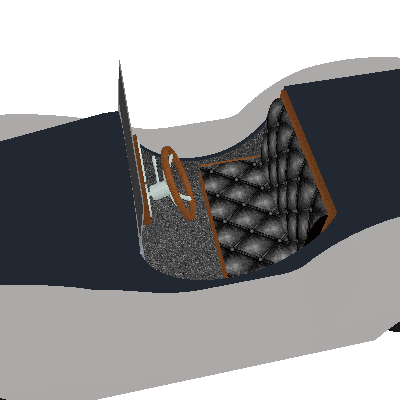}
        \includegraphics[trim= 0cm 0cm 0cm 0cm,clip, width=0.19\linewidth]{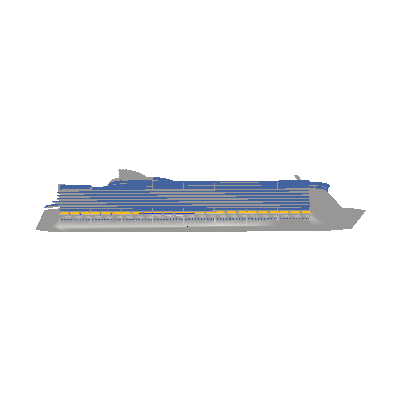}
    \includegraphics[trim= 0cm 0cm 0cm 0cm,clip, width=0.19\linewidth]{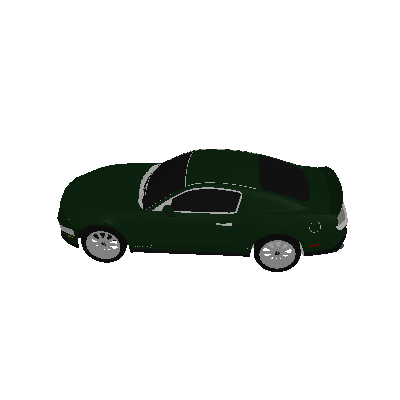}
    \includegraphics[trim= 0cm 0cm 0cm 0cm,clip, width=0.19\linewidth]{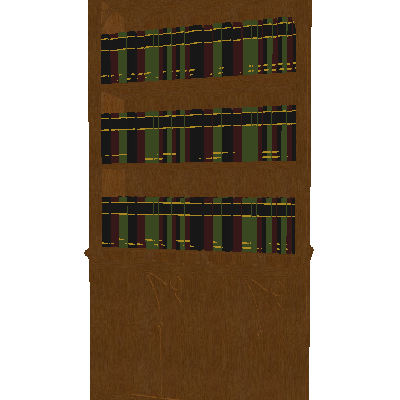}
\vspace{2pt}
    \caption{\textbf{SPARF: a Large Dataset for 3D Shapes Radiance Fields and Novel Views Synthesis}.  
    }
    \label{fig:sparf}
    \vspace{-4mm}
\end{figure}
\begin{figure}[t]
    \centering
        \tabcolsep=0.03cm
\resizebox{0.85\linewidth}{!}{
\begin{tabular}{ccc}
SPARF (ours) & SRN \cite{srn}  & NMR \cite{dvr} \\
     \includegraphics[trim= 1.0cm 1.5cm 1.0cm 1.9cm,clip, width=0.333\linewidth]{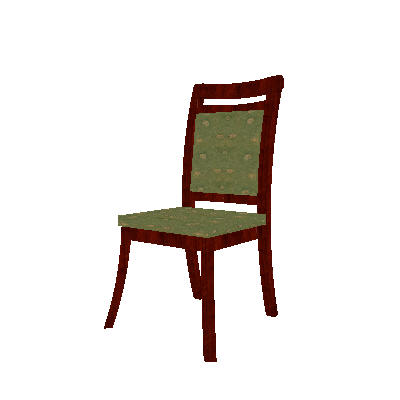} &
     \includegraphics[trim= 0.6cm 1cm 0.6cm 0.8cm,clip, width=0.35\linewidth]{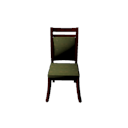} &
     \includegraphics[trim= 0.25cm 0.3cm 0.25cm 0.3cm,clip, width=0.30\linewidth]{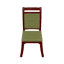} \\
\bottomrule
\end{tabular}
}
\vspace{3pt}
    \caption{\textbf{SPARF \vs other Datasets}. SPARF offers a large-scale high-resolution dataset compared to other posed multi-view datasets. We show the same chair here on SPARF, SRN, and NMR (please zoom in for differences). This highlights the huge quality gap between SPARF and other ShapeNet-based datasets.
    }
    \label{fig:datasets}
    \vspace{-4mm}
\end{figure}

\mysection{Few-Image NeRFs}
To address the original NeRF's requirement of  many posed images, several methods were proposed. The seminal work PixelNerf \cite{pixelnerf} is the first to reduce the image data requirements in order to learn a NeRF by using a trained CNN prior that can allow for transferable representation between scenes. Similarly, MVSNerf \cite{mvsnerf,fastandexplicit}, AutoRF \cite{autorf}, and ShaRF \cite{sharf} learn a CNN prior to generalize across scenes. IBRNet \cite{ibrnet} learns to render novel views based on neighboring views and optimized neural volume representation. More recently, VisionNerf \cite{visionnerf} proposes to use a ViT \cite{vit} to extract global features from the input images to enhance the capability of the NeRF MLP to predict the radiance field when one image is used as input. Unlike these works, we propose SuRFNet to directly learn from the 3D sparse radiance fields. Such a 3D setup benefits from structured 3D learning, creating a 3D prior that guarantees multi-view consistency while benefiting from the speed of recent voxel-based methods. A concurrent work by Guo \etal \cite{fastandexplicit} learns a 3D prior based on a perceptual loss, but does not use 3D supervision and only uses dense voxels ( limiting the pipeline to the low resolution of $64^3$ ).

\mysection{Datasets for novel view synthesis} Several datasets were proposed to support the task of novel view synthesis. NeRF \cite{nerf} introduced 8 synthetic scenes with 360-degree views.
Wang \etal \cite{mvs_google} introduced Google Scanned Objects. Other datasets for training multi-view algorithms include DTU \cite{dtu}, LLFF \cite{mvs_life}, Tanks and Temples \cite{mvs_tanks}, Spaces \cite{mvs_spaces}, RealEstate10K \cite{mvs_real}, SRN \cite{srn}, Transparent Objects \cite{mvs_transparent}, ROBI \cite{mvs_robi}, CO3D \cite{co3d}, SAPIEN \cite{mvs_sapian}, and BlendedMVS \cite{mvs_blender}. Recently, RTMV \cite{rtmv} introduced a ray-traced posed multi-view dataset with 2000 scenes and high-resolution images. Unfortunately, the current posed  multi-view datasets commonly used in NeRF research are either small in image resolution or the number of posed images (SRN \cite{srn} and NMR \cite{dvr}), small in the number of scenes/shapes and classes (Synthetic NeRFs \cite{nerf}), or lack structure (DTU \cite{dtu}, RTMV \cite{rtmv}) and Objeverse \cite{Objaverse}. A detailed multi-attribute comparison is provided  in Table \ref{tbl:mvdatasets}. 

\section{SPARF: a Large Dataset of 3D Shapes Radiance Fields} \label{sec:dataset}
\vspace{-4pt}
\subsection{Motivations for SPARF}
One of the goals of this work is to learn to generate high-quality SRFs in one forward pass of a deep network to enable fast novel view synthesis. In order to do this, harnessing the power of deep networks would require a large dataset of SRFs that follow a distinct distribution (\eg different class categories). 
 Datasets similar to SPARF (SRN \cite{srn}, NMR \cite{dvr}) form the basis of many advancements that pushed the community of neural radiance fields forward \cite{pixelnerf,visionnerf}. As can be seen in Table \ref{tbl:mvdatasets} and Figures \ref{fig:sparf} and \ref{fig:datasets}, SPARF naturally extends both datasets in orders of magnitude in resolution, scale, and diversity, benefiting the entire research community. Furthermore, the density of views ( $\sim$ 17M ) is higher than in previous datasets. This setup helps in studying out-of-distribution generalization of the views as a new benchmark that was neglected in previous works (see \secLabel{\ref{sec:results}}). 
\begin{figure*}
    \centering
    \tabcolsep=0.03cm
\resizebox{0.9\linewidth}{!}{
\begin{tabular}{cccccc}
      \includegraphics[trim= 0.0cm 0cm 0cm 0cm,clip, width=0.2\linewidth]{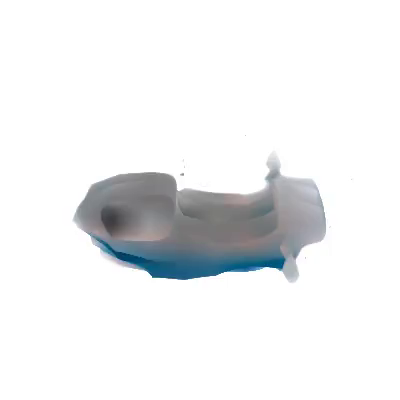}  &
     \includegraphics[trim= 0.0cm 0cm 0cm 0cm,clip, width=0.2\linewidth]{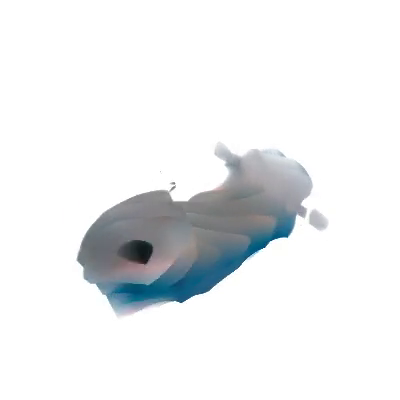}  &
     \includegraphics[trim= 0.0cm 0cm 0cm 0cm,clip, width=0.2\linewidth]{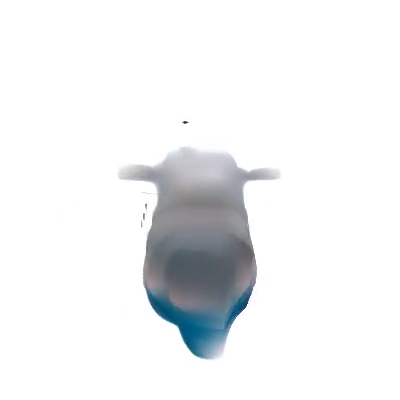}  &
     \includegraphics[trim= 0.0cm 0cm 0cm 0cm,clip, width=0.2\linewidth]{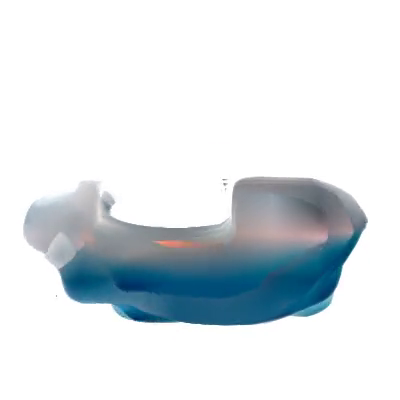}  &
     \includegraphics[trim= 0.0cm 0cm 0cm 0cm,clip, width=0.2\linewidth]{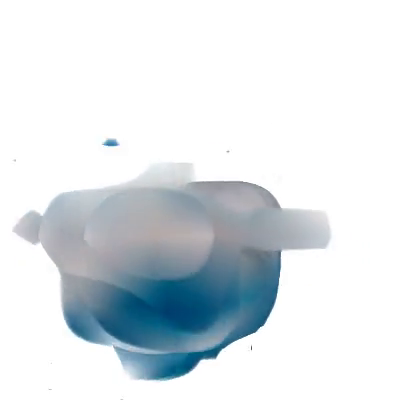}  & 
     \includegraphics[trim= 0.0cm 0cm 0cm 0cm,clip, width=0.2\linewidth]{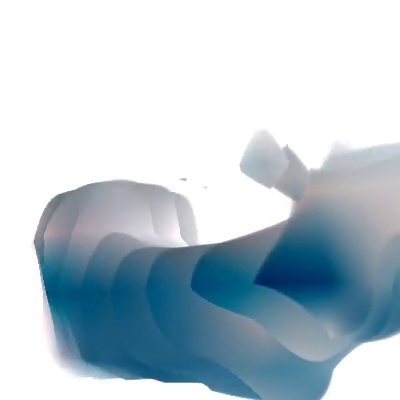}   \\
     \includegraphics[trim= 0.0cm 0cm 0cm 0cm,clip, width=0.2\linewidth]{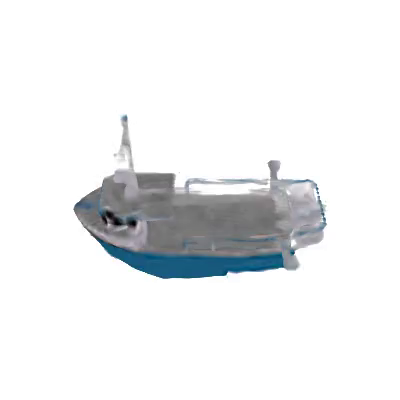}  &
     \includegraphics[trim= 0.0cm 0cm 0cm 0cm,clip, width=0.2\linewidth]{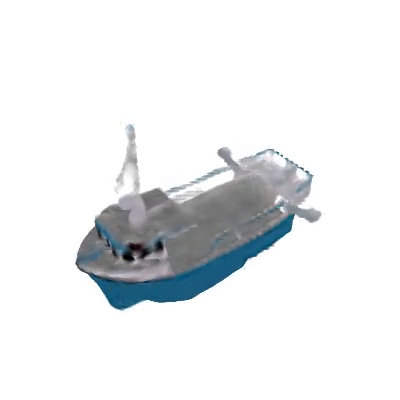}  &
     \includegraphics[trim= 0.0cm 0cm 0cm 0cm,clip, width=0.2\linewidth]{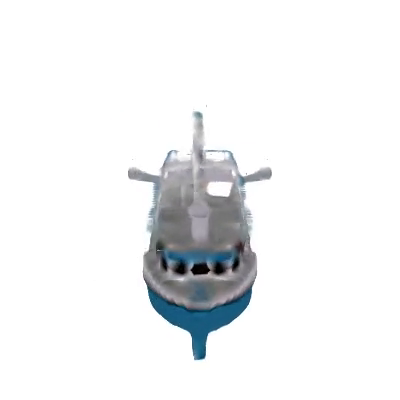}  &
     \includegraphics[trim= 0.0cm 0cm 0cm 0cm,clip, width=0.2\linewidth]{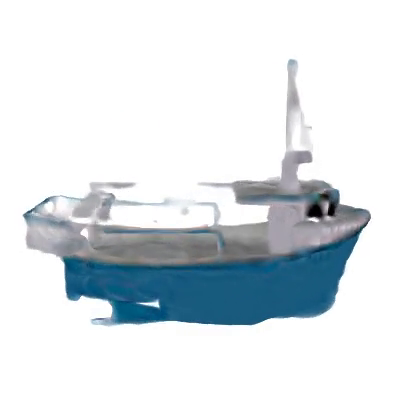}  &
     \includegraphics[trim= 0.0cm 0cm 0cm 0cm,clip, width=0.2\linewidth]{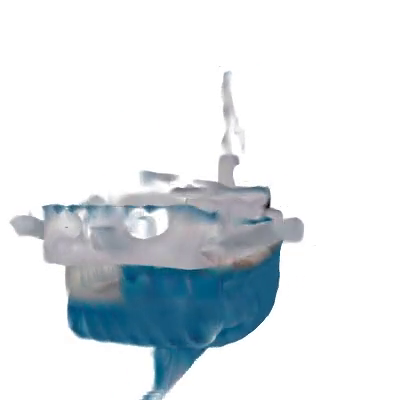}  & 
     \includegraphics[trim= 0.0cm 0cm 0cm 0cm,clip, width=0.2\linewidth]{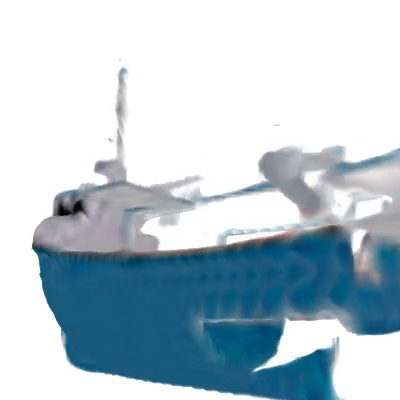}   \\
     \includegraphics[trim= 0.0cm 0cm 0cm 0cm,clip, width=0.2\linewidth]{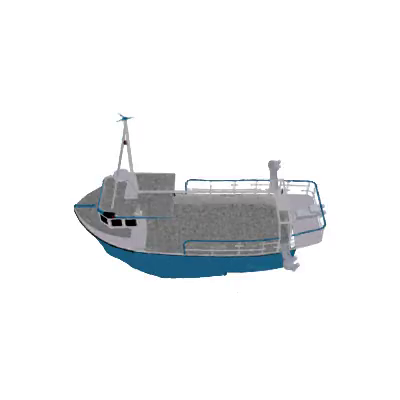}  &
     \includegraphics[trim= 0.0cm 0cm 0cm 0cm,clip, width=0.2\linewidth]{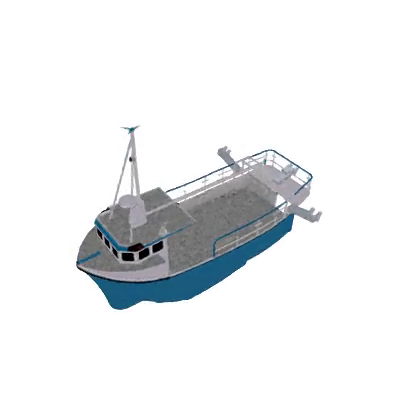}  &
     \includegraphics[trim= 0.0cm 0cm 0cm 0cm,clip, width=0.2\linewidth]{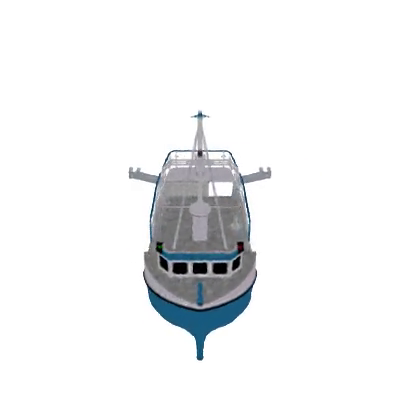}  &
     \includegraphics[trim= 0.0cm 0cm 0cm 0cm,clip, width=0.2\linewidth]{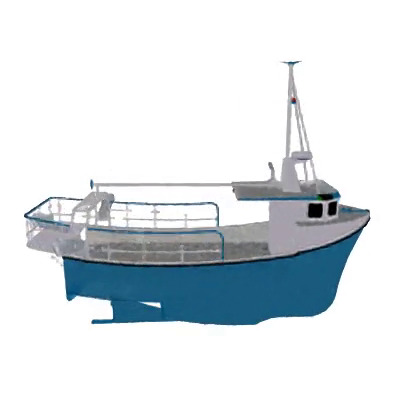}  &
     \includegraphics[trim= 0.0cm 0cm 0cm 0cm,clip, width=0.2\linewidth]{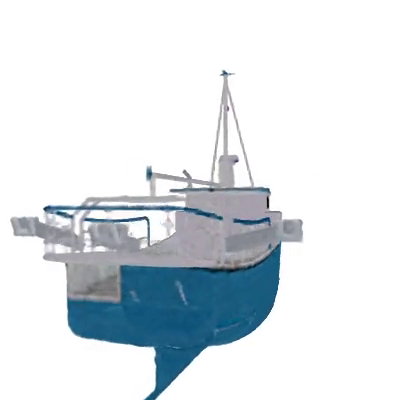}  & 
     \includegraphics[trim= 0.0cm 0cm 0cm 0cm,clip, width=0.2\linewidth]{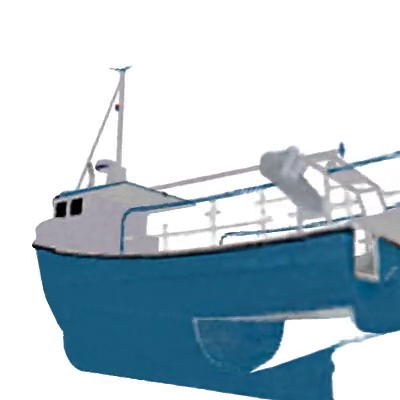}  \\
\bottomrule
\end{tabular}
}
\vspace{3pt}
    \caption{\textbf{SRFs: The optimized Sparse Radiance Fields in SPARF}.  A total of one million SRFs have been collected in SPARF, including on multiple voxel resolutions: 32 (\textit{\textit{top}}), 128 (\textit{middle}), and 512 (\textit{bottom}).
    }
    \label{fig:srf}
        \vspace{-4mm}
\end{figure*}
\subsection{Dense Posed Multi-view Image Dataset}
\vspace{-4pt}
The first step in collecting the desired large high-quality radiance field dataset is to collect a synthetic posed multi-view dataset. We use ShapeNet Core 55 \cite{shapenet} as the data of choice for 3D shapes. For rendering, we used Pyglet \cite{pyglet} API through Trimesh library \cite{trimesh}. The renderer is based on OpenGL \cite{opengl} rasterizer to render over 17 million images of around 40,000 shapes from 13 different classes at a high resolution of $400 \times 400$. Every shape is rendered equidistantly  from 400 views distributed in a spherical configuration surrounding the object, including from the bottom (see  \figLabel{\ref{fig:sparf}} for examples). An additional 20 views are rendered from random views from the same distance as test views for novel view synthesis tasks. Furthermore, an additional 10 views are rendered randomly from random distances bounded by a reasonable range, such that at least a part of the object is guaranteed to be visible. This last set is aimed at robustness purposes to test whether novel view synthesis methods can generalize to out-of-distribution posed views. 
\begin{figure}[t]
    \centering
    \tabcolsep=0.03cm
\resizebox{0.9\linewidth}{!}{
\begin{tabular}{cccc}
\multicolumn{2}{c}{SRF Renderings } &  \multicolumn{2}{c}{Extracted Mesh }  \\
     \includegraphics[trim= 0.0cm 1cm 0cm 0cm,clip, width=0.25\linewidth]{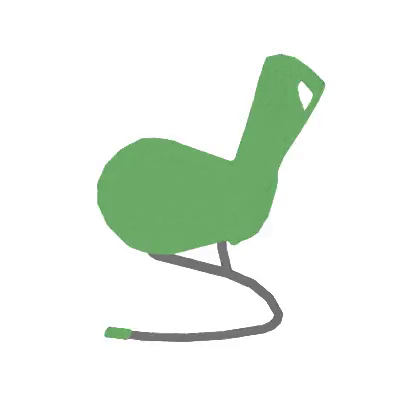} & \includegraphics[trim= 0.0cm 1cm 0cm 0cm,clip, width=0.25\linewidth]{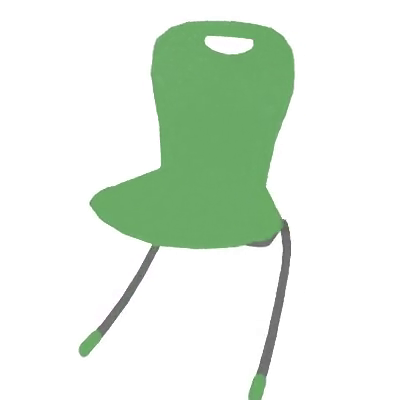}   & \includegraphics[trim= 0.0cm 1cm 0cm 0cm,clip, width=0.25\linewidth]{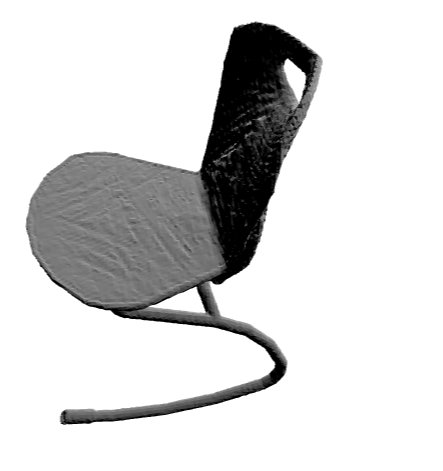} & \includegraphics[trim= 0.0cm 1cm 0cm 0cm,clip, width=0.25\linewidth]{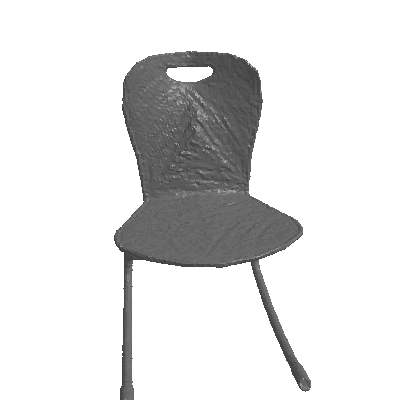}  \\
\bottomrule
\end{tabular}
}
\vspace{2pt}
    \caption{\textbf{Extracting 3D Meshes from SRFs}. Since SPARF and SuRFNet lie on the 3D voxel's space, extracting the mesh is straightforward with one pass of MarchingCubes \cite{mcubes}.   
    }
    \vspace{-4mm}
    \label{fig:mesh}
\end{figure}
\subsection{Multi-Resolution 3D SRFs}
\vspace{-4pt}
Sparse Radiance Field (SRF) can be defined as a voxel grid of dimension $1+d$, where $d$ is the dimension of radiance colors $\boldsymbol{\rho}_{i,j,k} \in \mathbb{R}^{d} $ at that specific $(i,j,k) $ indexed voxel in addition to one dimension for density $\alpha_{i,j,k} \in \mathbb{R}$. We assume that the grid is of size $H$ in each of the three dimensions: $\mathcal{X} \in \mathbb{R}^{H^3 \times (1+d)}$. Since the SRF is sparse, it can be represented with the COO format \cite{minkosky} as a set of $M$ tuples of positive integer coordinates $\mathbf{c}\in \mathbb{Z}^{+}$ and features $\mathbf{f}\in \mathbb{R}^{d+1}$ with the sparsity of $1-\frac{M}{H^3}$ as follows: 
\begin{equation}
\begin{aligned} 
 \mathcal{X}_{\text{non-empty}}= \{(\mathbf{c_m},\mathbf{f_m}) \}_{m=1}^{M}
\label{eq:srf}
\end{aligned} 
\end{equation}
The ordering of the set of tuples is arbitrary, but the features $\mathbf{f}_m$ consist of the density $\alpha_{i,j,k}$ and radiance colors $\boldsymbol{\rho}_{i,j,k}$ at that location $\mathbf{c}_m= (i,j,k)$. For the radiance field colors, we use the spherical harmonics proposed in Plenoxels \cite{plenoxels} for view-dependent learning of radiance common in NeRFs \cite{nerf}. The SRF can be viewed as a encoding of the NeRF MLP into sparse voxels for efficient optimization and volume rendering.  
In many 3D object tasks, a coarse-to-fine approach is followed \cite{magic3d}, demanding multiple resolutions. Hence, we collect the SPARF with multiple resolutions $H\in\{32,128,512\}$, as shown in \figLabel{\ref{fig:srf}}. We used an adaptation of Plenoxels \cite{plenoxels} to collect the dataset of a total of one million SRFs as we detail next. In order to scale up the Plenoxels optimization for this huge number of shapes and variants, we utilize a large number of images in the SPARF dataset to reduce the iterations to a minimum number while maintaining a high average PSNR across the dataset for the collected SRFs across the multiple resolutions.
highly detailed 3D meshes can be extracted easily from the collected SRFs as can be seen in \figLabel{\ref{fig:mesh}}.

\subsection{Representing Images with Partial SRFs} \label{sec:partials}
\begin{figure}[t]
    \centering
    \tabcolsep=0.03cm
\resizebox{0.85\linewidth}{!}{
\begin{tabular}{ccc}
Whole SRF & Partial (3 Views)  & Partial (1 View) \\
     \includegraphics[trim= 0.0cm 0cm 0cm 0cm,clip, width=0.3\linewidth]{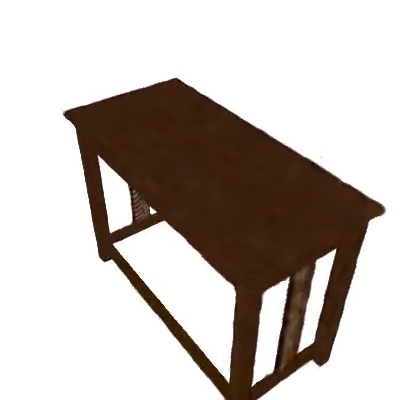}  &
     \includegraphics[trim= 0.0cm 0cm 0cm 0cm,clip, width=0.3\linewidth]{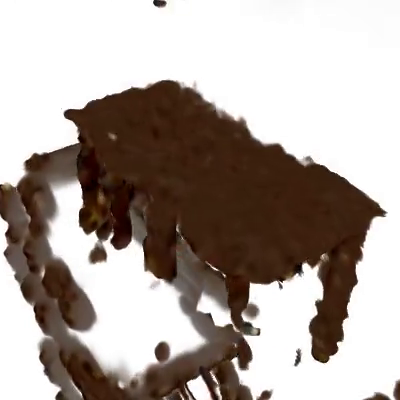}  &
     \includegraphics[trim= 0.0cm 0cm 0cm 0cm,clip, width=0.3\linewidth]{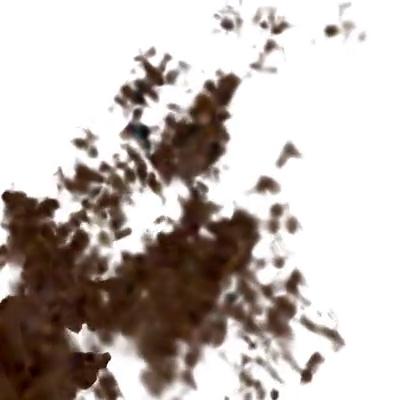}  \\
\bottomrule
\end{tabular}
}
\vspace{2pt}
    \caption{\textbf{Whole \vs Partial SRFs}. The partial SRFs are used instead of the few images that generated them as input to the learning pipeline to generate the whole SRFs 
    }
    \label{fig:partials}
\end{figure}

\vspace{-4pt}
In addition to collecting the ``whole" part of SPARF that utilized all 400 images for every shape in optimizing the SRFs, we collect ``partial" SRFs. These are SRFs that are quickly optimized on only a small number of images (1 or 3) randomly sampled from all 400 images, resulting in multiple partial SRF variants of that shape (see \figLabel{\ref{fig:partials}}). The Million SRFs dataset consists of two types of SRFs, either optimized using the entire set of 400 views to generate one Plenoxel \cite{plenoxels} per object, or optimized using multiple partial views from each object to generate multiple Plenoxel variants (see \figLabel{\ref{fig:stats}}). The partial SRFs were utilized as input to our SuRFNet pipeline (refer to Figure \ref{fig:pipeline}). The release of these partial SRFs should assist in 3D pipelines that consider radiance fields as a 3D data structure \cite{fastandexplicit}, as we demonstrate later in Section \ref{sec:ablation}. 

\section{Learning to Generate SRFs} \label{sec:methodology}
\vspace{-4pt}
\subsection{SuRFNet: Sparse Radiance Fields Network} \label{sec:network}
\begin{figure}[t]
    \centering
    \includegraphics[trim= 0.3cm 0cm 0.3cm 8cm,clip, width=\linewidth]{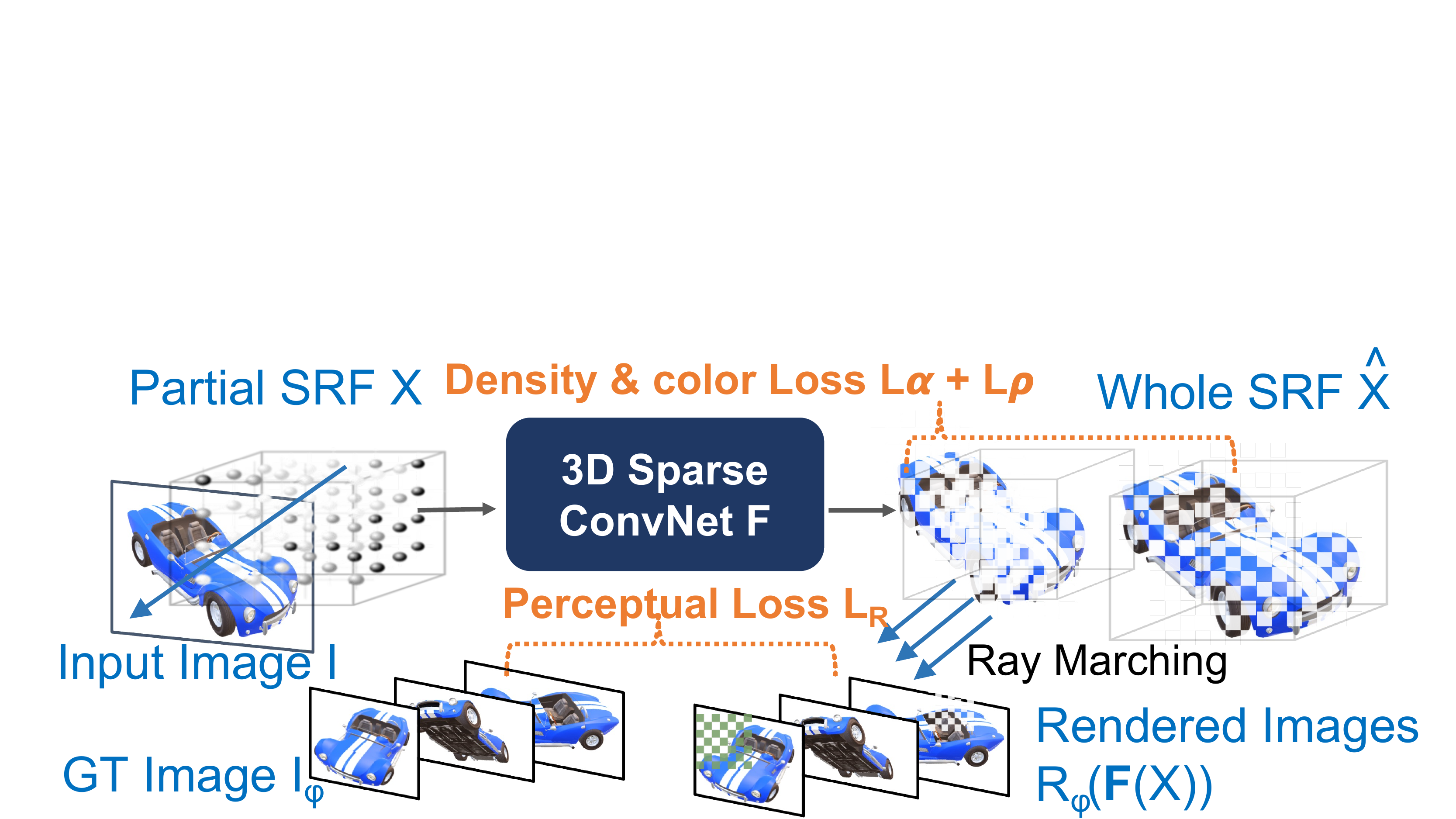}
    \vspace{2pt}
    \caption{\textbf{SuRFNet: Learning to Generate Whole Radiance Fields from Partial Views.} We process the input images into partial SRFs $\mathcal{X}$ before learning a sparse convolutional network to generate the whole SRF. A perceptual loss is employed on the rendered images from poses $\phi$ to enhance the perceptual quality of the generated SRF. The whole SRF $\hat{\mathcal{X}}$ is used to 3D-supervise the SRF generation with density and radiance color losses.     
    }
    \label{fig:pipeline}
\end{figure}

\vspace{-4pt}
\mysection{3D Pipeline}
Previous methods in few-views novel view synthesis (\eg PixelNerf \cite{pixelnerf} and VisionNerf \cite{visionnerf}) learn 2D priors to generalize across different shapes. 
On the other hand, we propose \textit{SuRFNet} to distill the image views into partial 3D SRFs and then perform the learning in the 3D sparse voxel space to generate full SRFs based on the partial SRFs.
Such a 3D Learning setup ensures multi-view consistency, especially when rendering from out-of-distribution views (as we show in \secLabel{\ref{sec:results}}). The input to the pipeline is the input partial SRFs from \secLabel{\ref{sec:partials}}, where the goal is learning a generalizable network that converts partial SRFs to whole SRFs as can be seen in \figLabel{\ref{fig:pipeline}}. 
We leverage the Minkowski Net \cite{minkosky} as the 3D sparse convolution network of choice. However, As this novel setup of learning SRFs poses new challenges, we propose several modifications to the typical pipeline of training MinkoskiNet (the \textit{type} of losses and \textit{where} to define them). 

\mysection{Challenges of Learning SRFs}
As a 3D data structure, SRF is an irregular volumetric representation that does not necessarily reflect the underlying 3D shape/scene, as it results from the optimization of posed images into volume. Many of the non-empty voxels have low densities and do not affect the volume rendering, but include color information that can confuse the network. Also, small errors in predicting the densities or radiance colors can result in large distortions in the rendered images, hurting the overall performance of novel view synthesis.
Another challenge in learning SRFs is vanishing gradients.
The typical sparsity in our setup is $\sim 99\%$, and misalignment between the input SRF coordinates and output SRF coordinates can further harm the gradients and affect the learning process. In order to tackle these issues, we propose three specialized losses detailed next.
\begin{figure}[t]
    \centering
    \includegraphics[trim= 0cm 1.2cm 0cm 0cm,clip, width=0.98\linewidth]{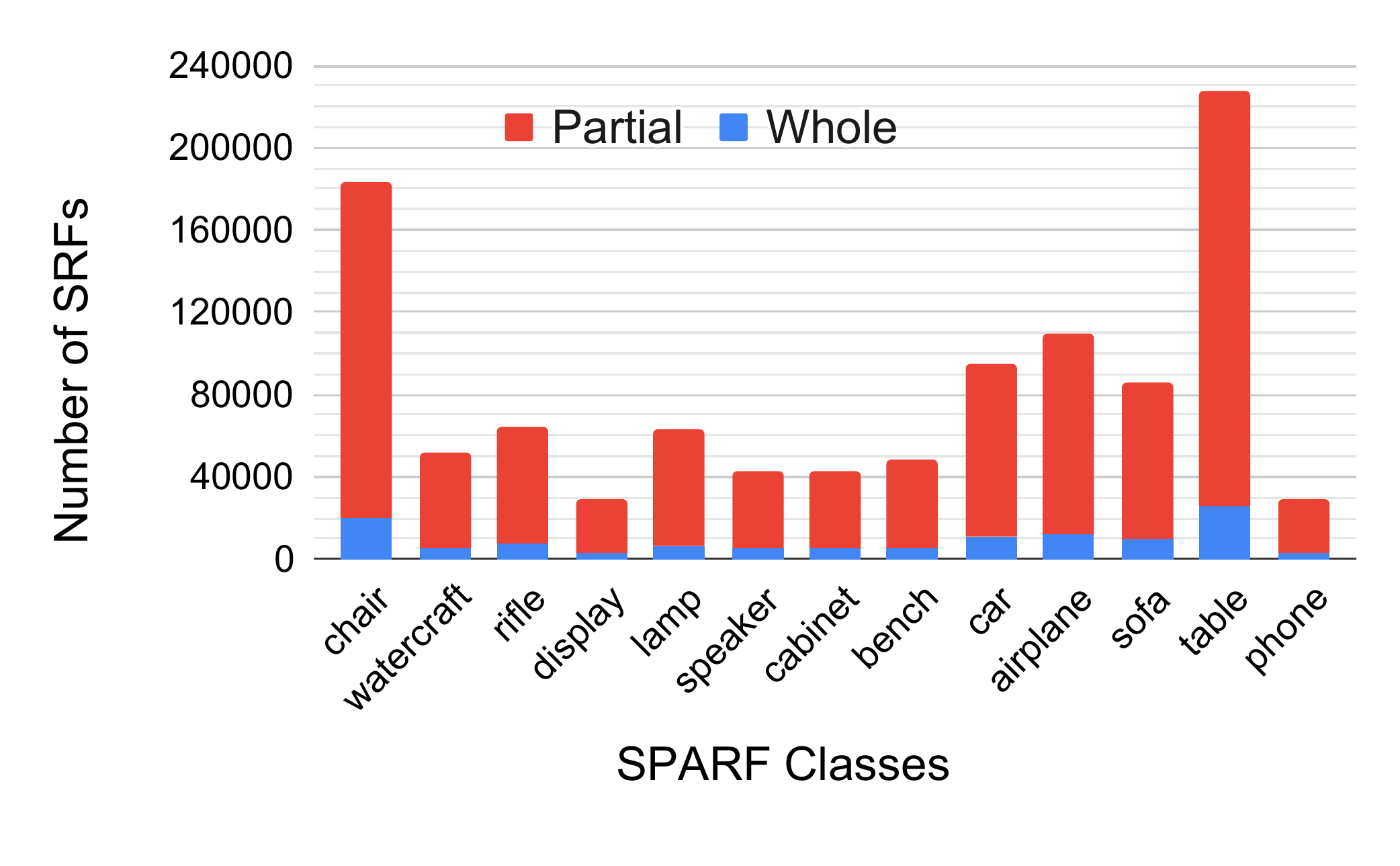}
    \vspace{2pt}
    \caption{\textbf{SPARF Distribution}. We show the distribution of classes in SPARF and how the one million partial and whole SRFs are distributed. The numbers are equally distributed on three voxel resolutions: 512,128. and 32.
    }
    \label{fig:stats}
    \vspace{-4mm}
\end{figure}
\subsection{SRF-Loss} \label{sec:loss}
\vspace{-4pt}
\mysection{Density loss}
The goal of the density loss is to create a dense surface. We propose the following binary cross-entropy loss on the predicted densities $\alpha$ as follows:
\begin{equation}
\begin{aligned} 
 & L_{\alpha}~ \big( \mathcal{X}~,~\hat{\mathcal{X}} \big)~ = -{(\mathbf{\hat{y}}\log(\mathbf{y}) + (\mathbf{1} - \mathbf{\hat{y}})\log(\mathbf{1} - \mathbf{y}))}  \\
 & \text{s. t.} \quad \mathbf{\hat{y}} = ~ \mathbbm{1}\left(\mathcal{S}(\hat{\mathcal{X}}_{\alpha} ) > \alpha_{\text{dense}} \right)~~, ~ \mathbf{y} = \mathcal{S}\left( \mathbf{F}(\mathcal{X})\right)_{\alpha}
\label{eq:density}
\end{aligned} 
\end{equation}
where $\mathcal{X},\hat{\mathcal{X}}$ are the input partial SRF and the ground truth whole SRFs respectively (as defined in \eqLabel{\ref{eq:srf}} ), and $\alpha_{\text{dense}}$ is the density threshold distinguishing dense voxels from the air (usually set to 0). The sampling function $\mathcal{S}$ samples points in the grid space where the loss is defined on the outputs $\mathbf{y}$ and the ground truth densities' labels $\mathbf{\hat{y}}$.
One of the challenges in working with sparse voxels of high resolution is that training the pipeline can not involve  densifying the voxels to the original resolution due to prohibitive memory requirements. The input/output topologies are not necessarily the same, as the sparse convolutional strides and pruning can alter the sparse voxels’ coordinates. This is why the sampling function $\mathcal{S}$ in \eqLabel{\ref{eq:density}} is of utmost importance in guiding the training of SuRFNet. We sample the loss according to Quantized Gaussian sampling (Q-Gaussian) with random coordinates centered at the middle of the voxel grid. 
Simply put, the Q-Gaussian is 3D normal distribution quantized to integer coordinates to give a prior about where the output is expected and where the loss is defined. Further details and alternative configurations are provided in \secLabel{\ref{sec:ablation}}. 

\mysection{Radiance color loss}
To ensure the output SRFs follow the ground truth optimized SRFs in radiance color, we follow the simple L1 loss on the radiance colors $\boldsymbol{\rho}$. However, as mentioned earlier, some of the non-empty voxels contain low density and will not be seen in the rendering, and can contain any random colors. Therefore, we mask these non-empty low-density voxels out of the L1 loss as follows:
\begin{equation}
\begin{aligned} 
 L_{\rho}~& \big( \mathcal{X}~,~\hat{\mathcal{X}} \big)~ = \| \mathbf{M}_{\alpha}\mathbf{F}(\mathcal{X})_{\rho} - \mathbf{M}_{\alpha}\hat{\mathcal{X}}_{\rho}  \|_{1}  \\
 & \text{s. t.} \quad \mathbf{M}_{\alpha} = ~ \mathbbm{1}(\hat{\mathcal{X}}_{\alpha}> \alpha_{\text{dense}} )
\label{eq:color}
\end{aligned} 
\end{equation}

\mysection{Perceptual loss}
Using only the 3D radiance color loss in \eqLabel{\ref{eq:color}} ignores the rendering quality of the generated SRF, and would make it sensitive to hyperparameters (see \figLabel{\ref{fig:loss}}). Hence, we introduce an online perceptual loss that would volume render the generated SRF during training from $M$ random  views that come from the same ground truth image poses $\phi$, and an L1 loss is defined between the generated images and the ground truth posed images $\mathbf{I}_{\phi}$.  
\begin{equation}
\begin{aligned} 
 L_{R}~& \big( \mathcal{X} \big)~ = \| \mathcal{R}_{\phi} \left( \mathbf{F}(\mathcal{X})\right) - \mathbf{I}_{\phi}  \|_{1},
\label{eq:perceptual}
\end{aligned} 
\end{equation}
where $\mathcal{R}_{\phi}$ is the fast volume rendering function that renders SRFs from poses $\phi$ using the trilinear interpolation between voxels proposed in Plenoxels\cite{plenoxels}.  

The final loss to train the network $\mathbf{F}$ would be combining the three losses in \eqLabel{\ref{eq:density},\ref{eq:color},\ref{eq:perceptual}} as follows:
\begin{equation}
\begin{aligned} 
\text{Loss}_{\mathbf{F}} ~=~ L_{\alpha} ~+~  \lambda_{\rho} L_{\rho} ~+~\lambda_{R} L_{R},
\label{eq:final}
\end{aligned} 
\end{equation}
where $\lambda_{\rho},\lambda_{R} $ are hyperparameters to control the radiance colors compared to the density predictions. The network is trained on all $N$ whole SRFs $\hat{\mathcal{X}}$ in the dataset, while the input SRFs $\mathcal{X}$ are randomly chosen from several partial SRFs created by the same number of images from those shapes. 
\begin{table*}[]
\centering
\resizebox{1.0\linewidth}{!}{
\tabcolsep=0.12cm
\begin{tabular}{l|ccccccccccccc|c}
\toprule
 & \multicolumn{13}{c}{\textbf{SPARF Validation PSNR} } & \\
\textbf{Baselines} & chair & watercraft & rifle & display & lamp & speaker & cabinet & bench & car & airplane & sofa & table & phone& mean \\

\midrule
Plenoxels \cite{plenoxels} (1V)   & 9.2  & 11.1   & 11.7  & 8.0 & 13.6 & 8.2 & 10.4  & 10.5 & 7.1 & 12.8 & 9.3 & 9.9 & 8.3 & 10.0    \\ \rowcolor[HTML]{EFEFEF} 
Plenoxels \cite{plenoxels} (3V)   & 10.7  & 13.3   & 14.9  & 9.7 & 15.8 & 10.4 & 12.4  & 11.6 & 7.1 & 14.6 & 11.6 & 10.8 & 9.7 & 11.7    \\ 
PixelNerf \cite{pixelnerf} (1V)   & 13.3  & 16.3   & 16.7  & 11.9 & 17.6 & 11.3 & 14.5  & 14.6 & 13.2 & 19.2 & 13.3 & 13.2 & 13.2 & 14.5    \\ \rowcolor[HTML]{EFEFEF} 
PixelNerf \cite{pixelnerf} (3V)   & 13.5  & 16.6   & 16.9  & 12.2 & 17.9 & 11.9 & 14.9  & 14.8 & 13.4 & \textbf{19.4} & 13.4 & 13.3 & 13.3 & 14.7    \\
VisionNeRF \cite{visionnerf} (1V) & 13.0  & 15.6   & 15.8  & 11.7 & 16.7 & 11.2 & 14.0  & 14.3 & 12.7 & 17.8 & 13.3 & 13.0 & 12.6 & 14.0   \\ \midrule \rowcolor[HTML]{EFEFEF} 
\textbf{SuRFNet (ours) (1V)}      & 11.6	& 16.2 & 	17.0 & 	12.0 & 	16.2 &	12.6 &	17.0 & 	13.5 & 	16.6 & 	17.5 & 	14.1 & 	10.1 & 	15.3 & 	14.6   \\
\textbf{SuRFNet (ours) (3V)}      & \textbf{15.3}	& \textbf{18.3} & 	\textbf{18.8} & 	\textbf{15.0} & 	\textbf{19.0} &	\textbf{16.6} &	\textbf{20.0} & 	\textbf{15.6} & 	\textbf{16.6} & 	18.5 & 	\textbf{18.1} & 	\textbf{14.9} & 	\textbf{17.8} & 	\textbf{17.3}   \\
\bottomrule
\end{tabular}
}
\vspace{2pt}
\caption{\small \textbf{SPARF Benchmark on Out-Of-Distribution View Synthesis}. We compare the validation PSNR of some of the widely used novel view synthesis techniques on the SPARF dataset for the generalization of novel view synthesis beyond a single example and on view tracks completely different from the ones seen in training views. One view (1V) and three views (3V) inputs are reported.}
\vspace{-4mm}
    \label{tbl:robustness}
\end{table*}
\section{Experiments} \label{sec:experiments}
\vspace{-4pt}
\subsection{Collecting SPARF} \label{sec:collecting}
\vspace{-4pt}
The engineering aspect of collecting, storing, and organizing the one million SRFs with multiple resolutions is as challenging as training properly on SRFs. In order to do that in manageable time and memory, while maintaining high quality in the optimized samples, a set of strategies is employed. The dimension of the radiance color $d$ is chosen to be $d=3\times4=12$ of 4 spherical harmonics factors of RGB channels for view-dependent SRF and $d=3\times1=3$ for fixed RGB colors of the SRF. Since the input partial SRFs are noisy, we use $d=3$ while the final output SRFs use $d=12$ for high-quality image generation. 
Using fewer Spherical Harmonics components (from 9 to 4 per RGB channel) reduces the optimization space by 40\% and time by 10\%, while maintaining the same PSNR. Using RGB as colors instead of SH factors reduces PSNR by $\sim 1$ dB, space by 80\%, and time by 20\%. Running Plenoxels \cite{plenoxels} for fewer iterations (3$\times$12K) reduces the time by 30\% while maintaining the same PSNR. Using 400 views/shapes in SPARF to optimize the SRFs keep the time manageable in optimization ($\sim 4$ minutes for the 512 resolution whole SRFs) while maintaining high PSNR ($\sim 30$dB). It takes way less time than that for the rest of the setups (lower resolution or partial SRFs). Other similar sparse voxels optimizations (\eg InstantNGP \cite{instantneural}) can be used to obtain SRFs similarly as fast with similar quality.
A total of four variants of the partial SRFs are collected for all the resolutions and the partials use 1 and 3 images. The anatomy of the distribution of classes and SRFs in SPARF is presented in \figLabel{\ref{fig:stats}}. More details about SPARF and visualizations of some of its samples are available in the \supp . 
\begin{figure}[t]
    \centering
    \tabcolsep=0.03cm
\resizebox{0.9\linewidth}{!}{
\begin{tabular}{ccc}
Output w/o $L_R$ & Output w/ $L_R$  & Whole SRF  \\
     \includegraphics[trim= 0.0cm 1cm 0cm 1cm,clip, width=0.31\linewidth]{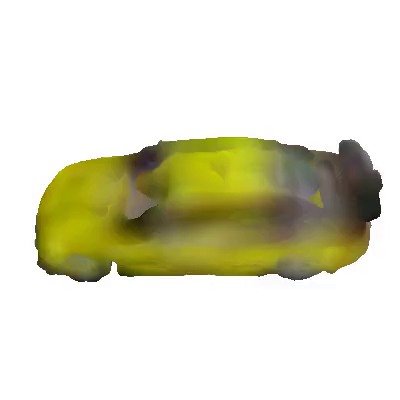} & \includegraphics[trim= 0.0cm 1cm 0cm 1cm,clip, width=0.31\linewidth]{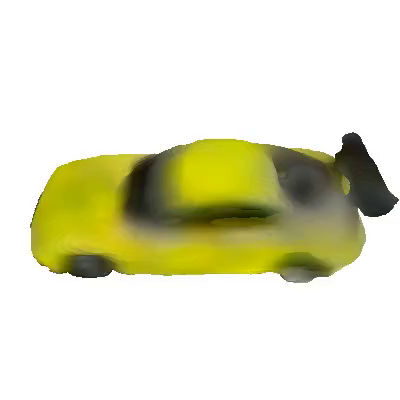}   & \includegraphics[trim= 0.0cm 1cm 0cm 1cm,clip, width=0.31\linewidth]{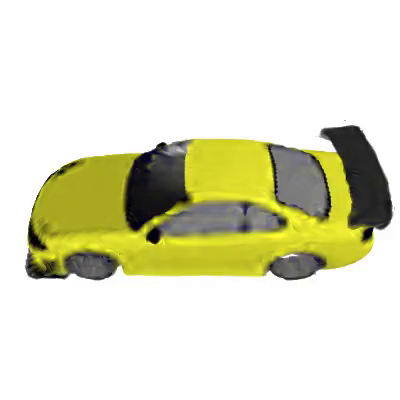}  \\
\bottomrule
\end{tabular}
}
\vspace{2pt}
    \caption{\textbf{Effect of the Perceptual Loss $L_R$}. Adding a perceptual loss on volume-rendered images during training SuRFNet insures the rendered images remain closer to how they should be rendered, as the 3D radiance colors supervision won't guarantee the rendering quality. \textit{(left)}: without perceptual loss , \textit{(middle)}: with the loss.
    }
    \label{fig:loss}
        \vspace{-8pt}
\end{figure}
\subsection{Training Setup} \label{sec:setup}
\vspace{-4pt}
\mysection{Dataset}
We pick our SPARF for the task of predicting whole SRFs for the purpose of novel view synthesis. The other datasets (SRN \cite{srn} and NMR \cite{dvr}) are too small or low in resolution, which prevents optimizing high-quality radiance fields (see \figLabel{\ref{fig:datasets}}). 

\mysection{Evaluation metrics} 
Following the previous novel view synthesis works \cite{pixelnerf,plenoxels,visionnerf}, we use PSNR, SSIM, and LPIPS \cite{lpips} as metrics to evaluate the synthesis. However, one key difference between our work and previous ones is that our setup is a learning setup (with training and validation), while previous works treat it as an optimization problem. Most previous works on novel view synthesis try to only generalize the generated views on the \textit{same} shape, while we aim to generalize \emph{across shapes of the same category} and \textit{across views}. We treat the collected whole SRFs as ground truth labels for the input few images from the training set. We consider the validation PSNR, SSIM, and LPIPS of the input images at the validation SRF set of shapes (on the test images of those shapes) as the main evaluation metrics. Also, we report validation accuracy = $\frac{\text{validation PSNR with test few images}}{\text{whole SRF optimization's PSNR}}$ and propose it as a new metric to evaluate such a learning setup of SRFs. Furthermore, as we describe in \secLabel{\ref{sec:dataset}}, SPARF has 10 posed Out-Of-Distribution (OOD) images for every 3D shape to evaluate the robustness  of novel view synthesis methods in the unconstrained setup. We report these \textit{OOD} PSNR, SSIM, LPIPS, and Accuracies as well. 

\mysection{Basleines} 
We use PixelNeRF \cite{pixelnerf} (ResNet34 backbone), Plenoxels \cite{plenoxels}, and VisionNerf  \cite{visionnerf} (ViT-B backbone) as the main baselines for our work. Our SuRFNet network has two sizes: large (87 million parameters) and small (13 million parameters).
\begin{figure}
    \centering
\tabcolsep=0.03cm
\resizebox{0.85\linewidth}{!}{
\begin{tabular}{cccc}
     
     \includegraphics[trim= 1.0cm 1cm 1cm 1cm,clip, width=0.24\linewidth]{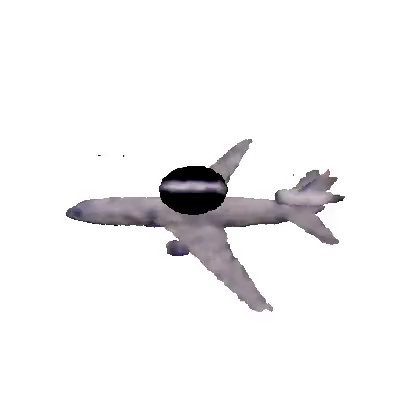}  &
     \includegraphics[trim= 1.0cm 1cm 1cm 1cm,clip, width=0.24\linewidth]{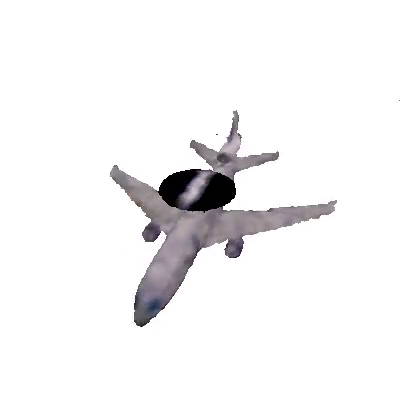}  &
     \includegraphics[trim= 1.0cm 1cm 1cm 1cm,clip, width=0.24\linewidth]{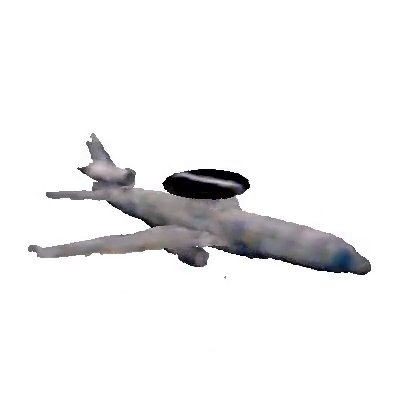} &
          \includegraphics[trim= 1.0cm 1cm 1cm 1cm,clip, width=0.24\linewidth]{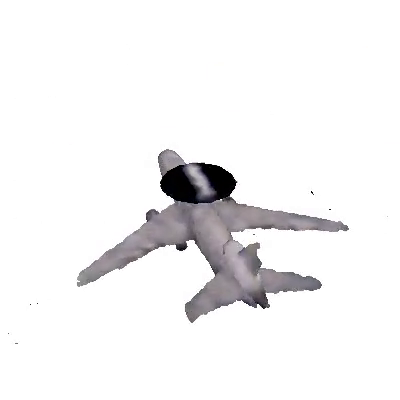} 
     \\
     
 \bottomrule
\end{tabular}
}
\vspace{2pt}
    \caption{\textbf{SuRFNet: Generating High-Resolution Radiance Fields}. We show some volume-rendered sequences based on our SuRFNet voxel radiance field outputs , given only 3 input images.
    }
    \label{fig:interpolation}
        \vspace{-4mm}
\end{figure}
\subsection{Results} \label{sec:results}
\begin{figure*} [t] 
\centering
\tabcolsep=0.03cm
\resizebox{\linewidth}{!}{
\begin{tabular}{c|cccccccc}
Input View & \multicolumn{2}{c}{Target Ground Truth} & \multicolumn{2}{c}{\textbf{SuRFNet (ours)}} & \multicolumn{2}{c}{PixelNeRF \cite{pixelnerf}}  & \multicolumn{2}{c}{VisionNeRF \cite{visionnerf}}     \\
    \includegraphics[trim= 0.0cm 0cm 0cm 0cm,clip, width=0.12\linewidth]{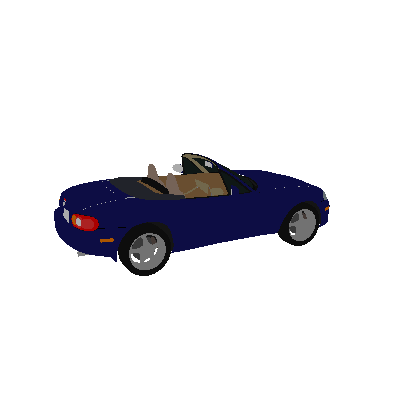} &
        \includegraphics[trim= 0.0cm 0cm 0cm 0cm,clip, width=0.1\linewidth]{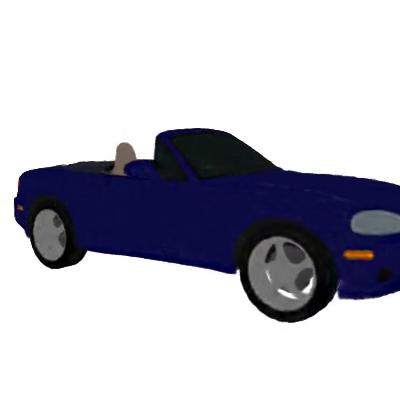}  & 
        \includegraphics[trim= 0.0cm 0cm 0cm 0cm,clip, width=0.1\linewidth]{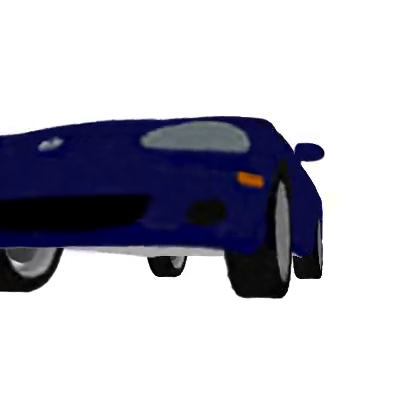} & 
        \includegraphics[trim= 0.0cm 0cm 0cm 0cm,clip, width=0.1\linewidth]{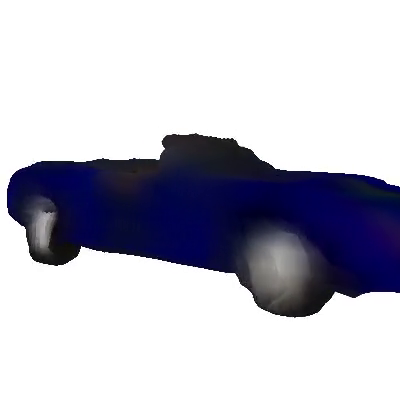}  &
        \includegraphics[trim= 0.0cm 0cm 0cm 0cm,clip, width=0.1\linewidth]{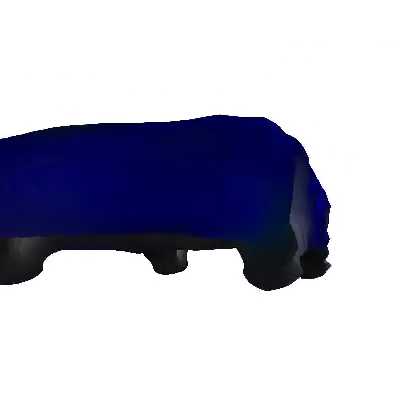} & 
    \includegraphics[trim= 0.0cm 0cm 0cm 0cm,clip, width=0.1\linewidth]{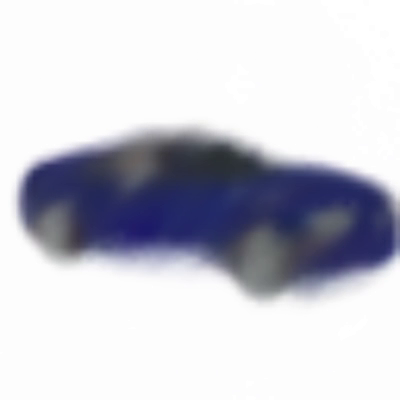}  &
    \includegraphics[trim= 0.0cm 0cm 0cm 0cm,clip, width=0.1\linewidth]{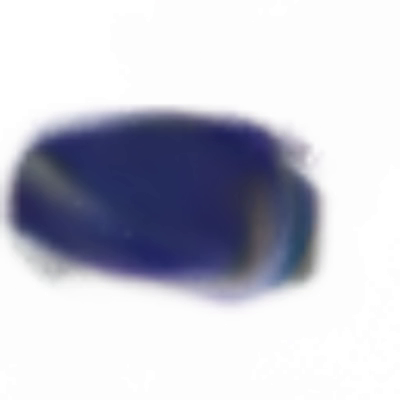}  &
    \includegraphics[trim= 0.0cm 0cm 0cm 0cm,clip, width=0.1\linewidth]{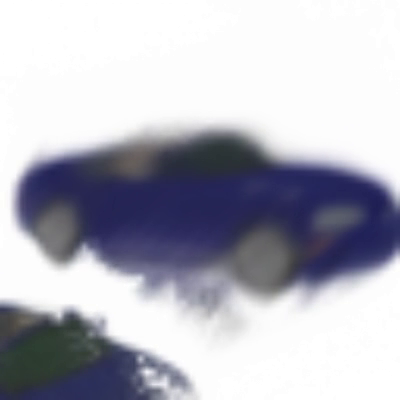}  &
        \includegraphics[trim= 0.0cm 0cm 0cm 0cm,clip, width=0.1\linewidth]{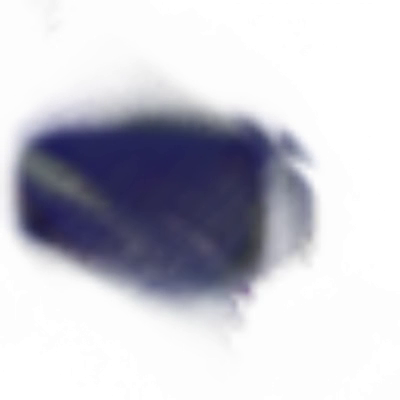}   \\   
 \bottomrule
\end{tabular}
}
\vspace{2pt}
    \caption{\textbf{Qualititve Comparisons}. We show different renderings from our SuRFNet outputs generated from a single image compared to other methods (pixel-NeRF \cite{pixelnerf}, and VisionNeRF \cite{visionnerf} ) and whole SRF ''GT" renderings. Note that the predicted two views lay outside the training views distribution (zoomed in randomly). This test highlights the weakness of the 2D-based baselines \cite{pixelnerf,visionnerf} outside the training track, while our 3D approach maintains multi-view consistency everywhere.
    }
    \label{fig:comparison}
\end{figure*}

\vspace{-4pt}
We show qualitative results of generating novel views from  few input images on unseen shapes in \figLabel{\ref{fig:interpolation}}. We also show qualitative comparisons in \figLabel{\ref{fig:comparison}}. We present a summary of the quantitive evaluations next, where SuRFNet achieves \sota results on unconstrained novel view synthesis from one or few images on unseen shapes.

\begin{figure}[t]
    \centering
    \includegraphics[trim= 0.2cm 1cm 0.2cm 0cm,clip, width=0.99\linewidth]{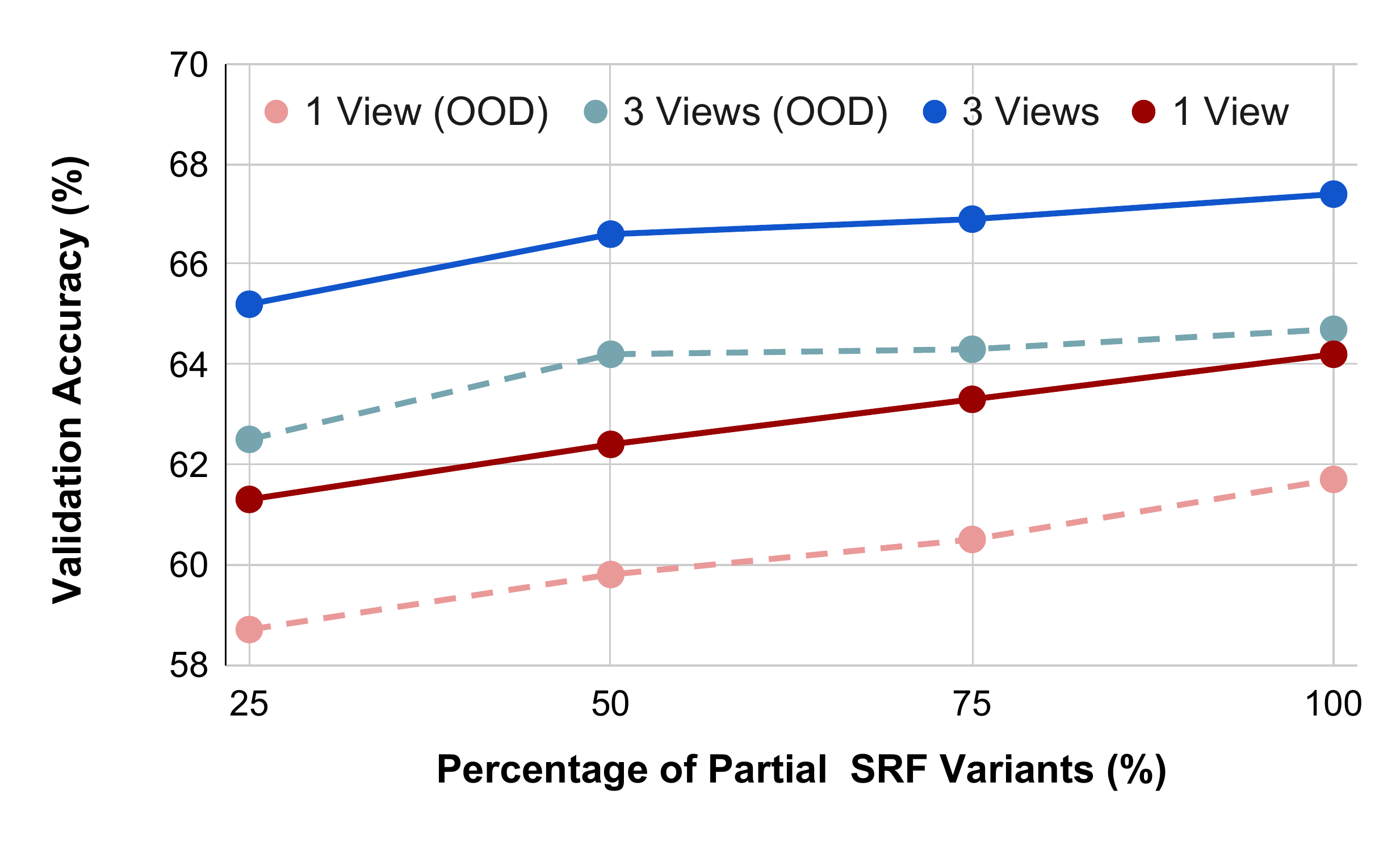}
    \caption{\textbf{Scaling-Up Training on SRFs}. As the training data (partial SRFs) of radiance fields increase, the generalization improves, as can be seen in the car class here. The 3-view and 1-view metrics are reported with test and OOD metrics.
    }
    \label{fig:variants}
        \vspace{-8pt}
\end{figure}
\mysection{SPARF View-Generalization benchmark}
In Table \ref{tbl:robustness}, we report the average PSNR results on the validation set of SPARF for different methods on unseen shapes during training on all 13 different object classes and on out-of-distribution views. It shows that our SuRFNet can generalize to out-of-distribution views on unseen shapes during test time, surpassing \sota PixelNeRF \cite{pixelnerf} and VisionNeRF \cite{visionnerf}. Visual comparisons can be found in \figLabel{\ref{fig:comparison}}. As can be seen from those results, the learned 3D prior results in multi-view consistency, especially when rendering from out-of-distribution views.

\begin{figure}[t]
    \centering
\tabcolsep=0.03cm
\resizebox{0.99\linewidth}{!}{
\begin{tabular}{ccccc}
Real & \multicolumn{4}{c}{Generated} \\
     \includegraphics[trim= 0.0cm 16cm 0cm 10cm,clip, width=0.198\linewidth]{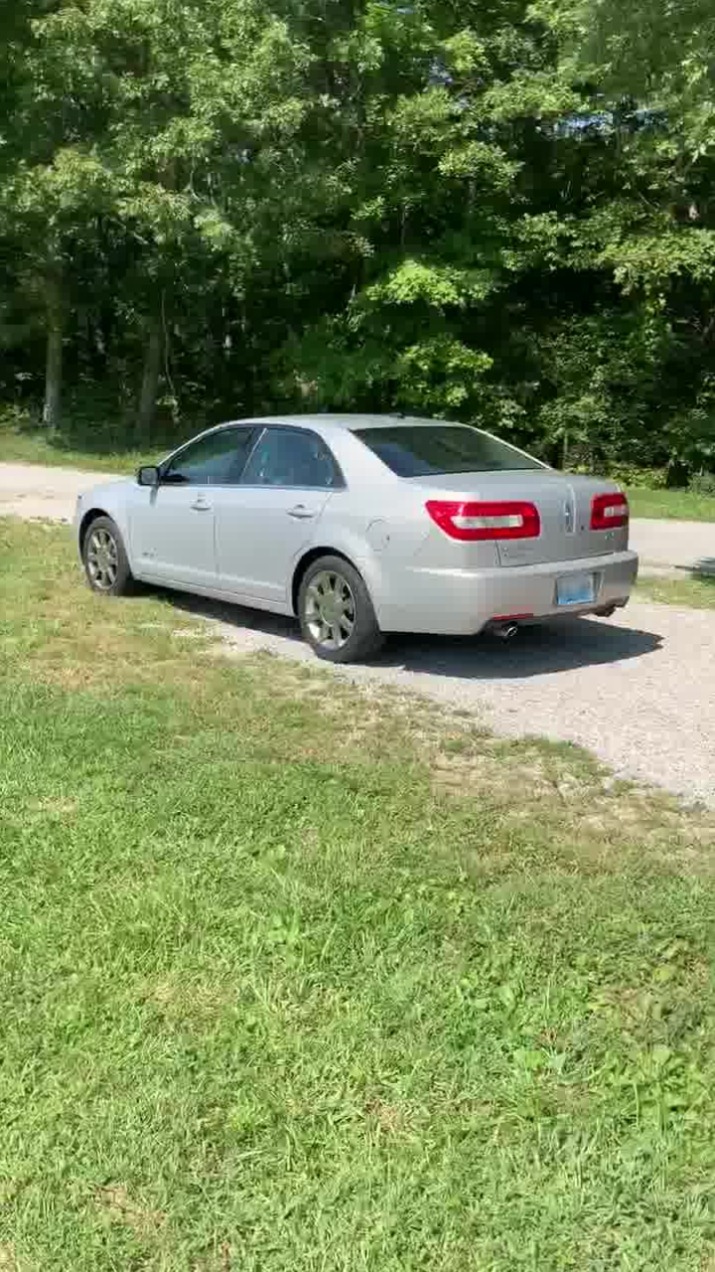}  &
     \includegraphics[trim= 0.0cm 9cm 0cm 9cm,clip, width=0.198\linewidth]{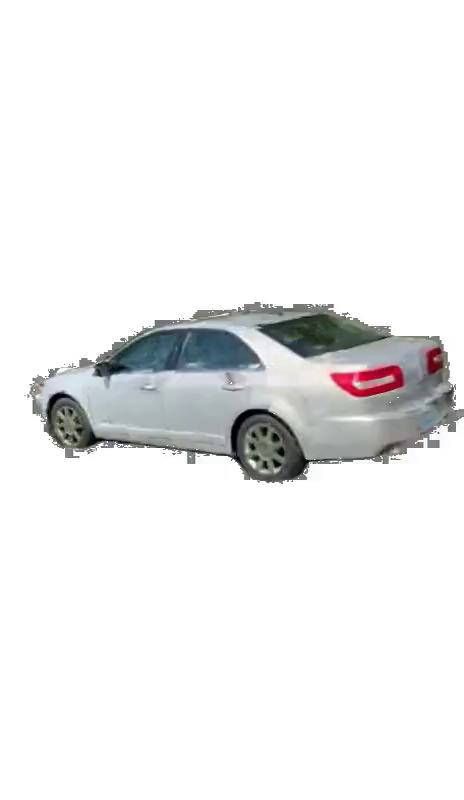}  &
          \includegraphics[trim= 0.0cm 9cm 0cm 9cm,clip, width=0.198\linewidth]{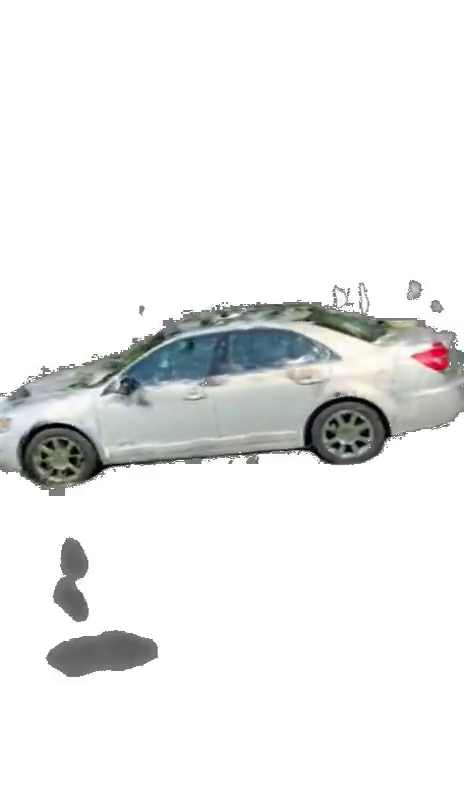} & 
          \includegraphics[trim= 0.0cm 9cm 0cm 9cm,clip, width=0.198\linewidth]{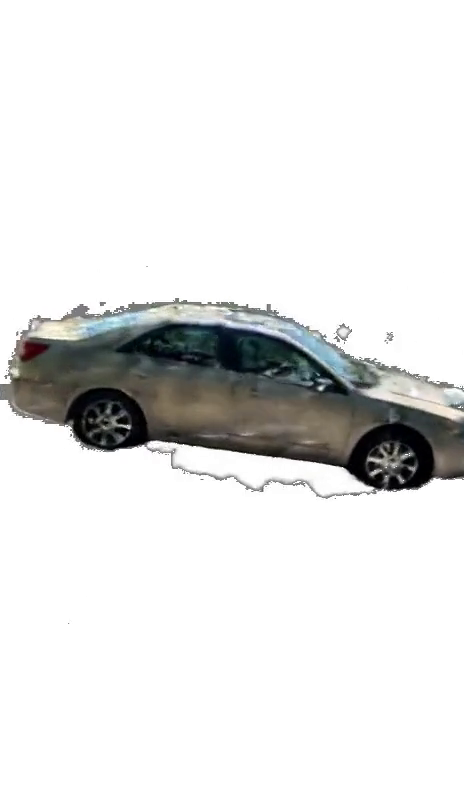}  &
          \includegraphics[trim= 0.0cm 9cm 0cm 9cm,clip, width=0.198\linewidth]{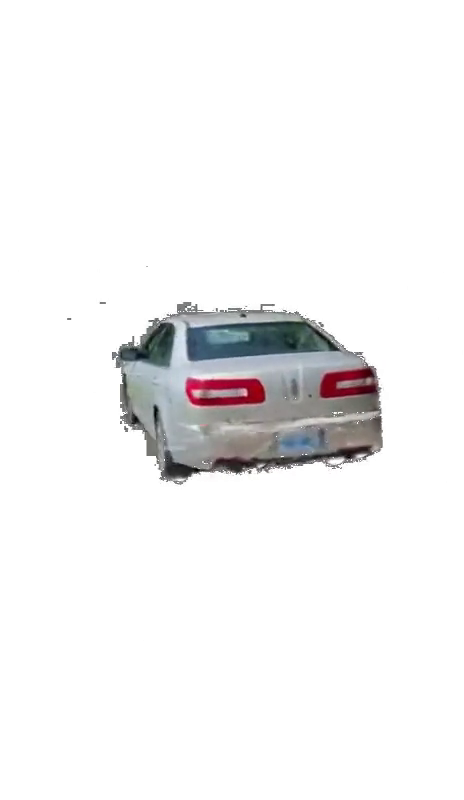}  \\ \hline

    \includegraphics[trim= 0.0cm 16cm 0cm 10cm,clip, width=0.198\linewidth]{images/real/s0/co.jpg}  &
     \includegraphics[trim= 5cm 6cm 5cm 5.5cm,clip, width=0.198\linewidth]{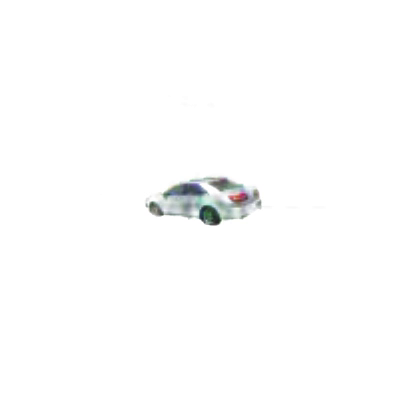}  &
          \includegraphics[trim= 5cm 6cm 5cm 5.5cm,clip, width=0.198\linewidth]{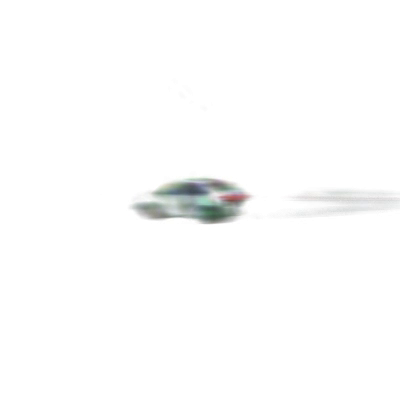} & 
          \includegraphics[trim= 5cm 6cm 5cm 5.5cm,clip, width=0.198\linewidth]{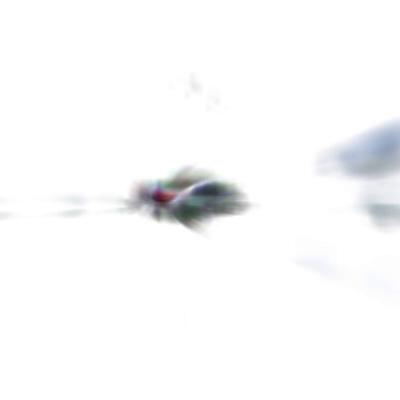}  &
          \includegraphics[trim= 5cm 6cm 5cm 5.5cm,clip, width=0.198\linewidth]{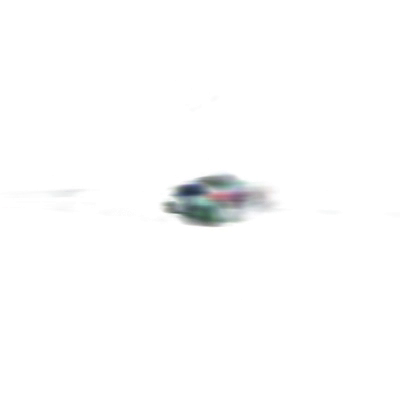}  \\

 \bottomrule
\end{tabular}
}
\vspace{2pt}
    \caption{\textbf{Real images} We show real images of Co3D \cite{co3d} and the corresponding generated views from our SRFs (\textit{top row}) and pretrained PixelNeRF \cite{pixelnerf} (\textit{bottom row}).
    }
    \label{fig:co3d}
    \vspace{-2mm}
    \end{figure}
\section{Analysis and Insights} \label{sec:analysis}
\vspace{-4pt}
\subsection{Ablation Study} \label{sec:ablation}
\vspace{-4pt}
\mysection{Effect of Dataset Size} \label{sec:datasetsize}
We study the effect of increasing the dataset size (Partial SRFs) on the generalization performance of SuRFNet in \figLabel{\ref{fig:variants}}. It shows that as the size increase ( normalized by the total number of shapes in \textit{car} class ), the generalization performance increase. This scalability effect underlines the importance of SPARF. This justifies collecting 4 variants per resolution ( as detailed in \figLabel{\ref{fig:stats}}).

\mysection{Training SuRFNet}  %
We ablate different components of SuRFNet's architectures and the loss configuration choices and report the results in Table \ref{tbl:ablation}. 
The Results show that increasing the size of the network (from 13M to 87M parameters ) helps improve generalization accuracy. Also, they show the importance of the loss components proposed in \eqLabel{\ref{eq:final}}.
The use of only density loss creates a reasonably dense shape but without colors. While combining the density loss with the 3D radiance color loss creates colorful objects, it does not perform well in the task of novel view synthesis as it does not respect how the object renders. Small radiance distortions lead to large image errors (see \figLabel{\ref{fig:loss}} for the importance of the perceptual loss). More ablations on the network, loss sampling, and hyperparameters of training SuRFNet can be found in \supp .

\begin{table}[t]
\footnotesize
\setlength{\tabcolsep}{4pt} %
\renewcommand{\arraystretch}{1} %
\centering
\resizebox{0.88\hsize}{!}{
\begin{tabular}{cc|ccc|c} 
\toprule
\multicolumn{2}{c|}{\textbf{3D Backbone}}& \multicolumn{3}{c|}{\textbf{Loss components}}&    \multicolumn{1}{c}{\textbf{Results}}\\
Small & Large & ($L_\alpha$) & ($L_\rho$) & ($L_R$) & Val. Acc.     \\
\midrule
\checkmark & - & \checkmark & \checkmark & -  & 65.2   \\ \hline
\checkmark & - & \checkmark & \checkmark & \checkmark  & 65.7    \\ \hline
- & \checkmark & \checkmark & \checkmark & -  & 66.4    \\ \hline
- & \checkmark & \checkmark & \checkmark & \checkmark  & \textbf{68.2}   \\ 

 \bottomrule
\end{tabular}
}
\vspace{2pt}
\caption{\small \textbf{Ablation Study}. We ablate different components of in SuRFNet (3D backbone and SRF-Loss) and report validation accuracy of \textit{car} class.}
\label{tbl:ablation}
    \vspace{-4mm}
\end{table}
\subsection{Testing on Real Images} \label{sec:real}
Previous works on similar large-scale learning setups report only the ShapeNet benchmark \cite{visionnerf,sharf,3dim} for standardized evaluations. Testing on real images can provide additional value. Hence, In Fig. \ref{fig:co3d}, we show renderings of whole SRFs (used in our SuRFNet pipeline) trained on \textit{real} Co3D \cite{co3d} images \vs pretrained PixelNeRF renderings. This demonstrates the potential for the proposed method on real images. The small noise is due to background corruption resulting from imperfect masks provided in Co3D. 

\subsection{Speed and Compute Cost} \label{sec:cost}
\vspace{-4pt}
To assess the contributions of the SuRFNet pipeline, we study the time and memory requirements of each element in the pipeline. We record in Table \ref{tbl:cost} the number of floating-point operations (GFLOPs) and the runtime of a forward pass (including rendering) for a single output image from one input image on Titan RTX GPU. Speed and compute details about collecting the SRFs are included in \secLabel{\ref{sec:collecting}}.
\begin{table}[]

\tabcolsep=0.11cm
\centering
\resizebox{\linewidth}{!}{
\begin{tabular}{l|cccc} 
\toprule
\multirow{2}{*}{{Network}}& Network & Network  &   Parameters & Rendering \\
 & FLOP (G) & Infer. (ms) &   Number (M) & Speed (FPS)\\
\midrule
PixelNeRF \cite{pixelnerf} &  7.3 &  5.33 & 21.8 & 1.2 \\
VisionNerf \cite{visionnerf} &  33.7 & 12.5   & 68.6 & 1.2 \\
SuRFNet (small) &  $\sim$15 &  14.4 & 13.4 & 15\\ 
SuRFNet (large) &  $\sim$100 &  90.0 & 87.3 & 15\\ 
\bottomrule
\end{tabular}
}
\vspace{2pt}
\caption{\small \textbf{Time and Memory Requirements}. We assess the computational cost of the main components studied}
\label{tbl:cost}
\end{table}

\section{Conclusions and Future Work} \label{sec:conclusion}
\vspace{-4pt}
We propose a large-scale dataset SPARF of sparse radiance fields that include around one million SRFs and 17 million posed images of 3D shapes. The dataset aims to move the community in the direction of treating radiance fields as a 3D data structure, instead of optimization results and MLP fitting. We domnstrated SPARF's utility using our proposed SuRFNet pipeline. 

One crucial limitation in this work is the large amount of compute and memory necessary to create, store, and process SRFs, especially at high-resolution voxel grids. This creates a bottleneck in training, developing,  and building on SRFs. Developing efficient methods to work and learn from sparse voxel grids would be a viable plan moving forward in order to develop deep and large models in this space.

\mysection{Acknowledgement}\label{sec:ack}
This work was supported by the ERC Starting Grant Scan2CAD (804724) and the German Research Foundation (DFG) Research Unit ``Learning and Simulation in Visual Computing". It was also supported by the KAUST Office of Sponsored Research through the Visual Computing Center funding. Part of the support is also coming from the KAUST Ibn Rushd Postdoc Fellowship program.

{\small
\bibliographystyle{ieee_fullname}
\bibliography{egbib}

\begin{thebibliography}{10}\itemsep=-1pt

\bibitem{pc-ae}
Panos Achlioptas, Olga Diamanti, Ioannis Mitliagkas, and Leonidas Guibas.
\newblock Learning representations and generative models for 3d point clouds.
\newblock {\em International Conference on Machine Learning (ICML)}, 2018.

\bibitem{pyglet}
{Alex Holkner et al.}
\newblock Pyglet.

\bibitem{pix2nerf}
Shengqu Cai, Anton Obukhov, Dengxin Dai, and Luc Van~Gool.
\newblock Pix2nerf: Unsupervised conditional p-gan for single image to neural
  radiance fields translation.
\newblock In {\em Proceedings of the IEEE/CVF Conference on Computer Vision and
  Pattern Recognition (CVPR)}, pages 3981--3990, June 2022.

\bibitem{pigan}
Eric~R Chan, Marco Monteiro, Petr Kellnhofer, Jiajun Wu, and Gordon Wetzstein.
\newblock pi-gan: Periodic implicit generative adversarial networks for
  3d-aware image synthesis.
\newblock In {\em Proceedings of the IEEE/CVF conference on computer vision and
  pattern recognition}, pages 5799--5809, 2021.

\bibitem{shapenet}
Angel~X. Chang, Thomas Funkhouser, Leonidas Guibas, Pat Hanrahan, Qixing Huang,
  Zimo Li, Silvio Savarese, Manolis Savva, Shuran Song, Hao Su, Jianxiong Xiao,
  Li Yi, and Fisher Yu.
\newblock {ShapeNet: An Information-Rich 3D Model Repository}.
\newblock Technical Report arXiv:1512.03012 [cs.GR], Stanford University ---
  Princeton University --- Toyota Technological Institute at Chicago, 2015.

\bibitem{mvsnerf}
Anpei Chen, Zexiang Xu, Fuqiang Zhao, Xiaoshuai Zhang, Fanbo Xiang, Jingyi Yu,
  and Hao Su.
\newblock Mvsnerf: Fast generalizable radiance field reconstruction from
  multi-view stereo.
\newblock In {\em Proceedings of the IEEE/CVF International Conference on
  Computer Vision}, pages 14124--14133, 2021.

\bibitem{vig-bid-r}
Wenzheng Chen, Huan Ling, Jun Gao, Edward Smith, Jaakko Lehtinen, Alec
  Jacobson, and Sanja Fidler.
\newblock Learning to predict 3d objects with an interpolation-based
  differentiable renderer.
\newblock In {\em Advances in Neural Information Processing Systems}, pages
  9609--9619, 2019.

\bibitem{minkosky}
Christopher Choy, JunYoung Gwak, and Silvio Savarese.
\newblock 4d spatio-temporal convnets: Minkowski convolutional neural networks.
\newblock In {\em Proceedings of the IEEE Conference on Computer Vision and
  Pattern Recognition}, pages 3075--3084, 2019.

\bibitem{scan2mesh}
Angela Dai and Matthias Nie{\ss}ner.
\newblock Scan2mesh: From unstructured range scans to 3d meshes.
\newblock In {\em Proceedings of the IEEE/CVF Conference on Computer Vision and
  Pattern Recognition}, pages 5574--5583, 2019.

\bibitem{trimesh}
{Dawson-Haggerty et al.}
\newblock trimesh.

\bibitem{Objaverse}
Matt Deitke, Dustin Schwenk, Jordi Salvador, Luca Weihs, Oscar Michel, Eli
  VanderBilt, Ludwig Schmidt, Kiana Ehsani, Aniruddha Kembhavi, and Ali
  Farhadi.
\newblock Objaverse: A universe of annotated 3d objects.
\newblock In {\em IEEE Conf. Comput. Vis. Pattern Recog.}, pages 13142--13153,
  2023.

\bibitem{vit}
Alexey Dosovitskiy, Lucas Beyer, Alexander Kolesnikov, Dirk Weissenborn,
  Xiaohua Zhai, Thomas Unterthiner, Mostafa Dehghani, Matthias Minderer, Georg
  Heigold, Sylvain Gelly, Jakob Uszkoreit, and Neil Houlsby.
\newblock An image is worth 16x16 words: Transformers for image recognition at
  scale.
\newblock {\em ICLR}, 2021.

\bibitem{mvs_spaces}
John Flynn, Michael Broxton, Paul Debevec, Matthew DuVall, Graham Fyffe, Ryan
  Overbeck, Noah Snavely, and Richard Tucker.
\newblock Deepview: View synthesis with learned gradient descent.
\newblock In {\em Proceedings of the IEEE/CVF Conference on Computer Vision and
  Pattern Recognition}, pages 2367--2376, 2019.

\bibitem{plenoxels}
Sara Fridovich-Keil, Alex Yu, Matthew Tancik, Qinhong Chen, Benjamin Recht, and
  Angjoo Kanazawa.
\newblock Plenoxels: Radiance fields without neural networks.
\newblock In {\em Proceedings of the IEEE/CVF Conference on Computer Vision and
  Pattern Recognition (CVPR)}, pages 5501--5510, June 2022.

\bibitem{get3d}
Jun Gao, Tianchang Shen, Zian Wang, Wenzheng Chen, Kangxue Yin, Daiqing Li, Or
  Litany, Zan Gojcic, and Sanja Fidler.
\newblock Get3d: A generative model of high quality 3d textured shapes learned
  from images.
\newblock In {\em Advances In Neural Information Processing Systems}, 2022.

\bibitem{meshrcnn}
Georgia Gkioxari, Jitendra Malik, and Justin Johnson.
\newblock Mesh r-cnn.
\newblock In {\em Proceedings of the IEEE/CVF International Conference on
  Computer Vision}, pages 9785--9795, 2019.

\bibitem{difstereopsis}
Shubham Goel, Georgia Gkioxari, and Jitendra Malik.
\newblock Differentiable stereopsis: Meshes from multiple views using
  differentiable rendering.
\newblock {\em arXiv preprint arXiv:2110.05472}, 2021.

\bibitem{fastandexplicit}
Pengsheng Guo, Miguel~Angel Bautista, Alex Colburn, Liang Yang, Daniel
  Ulbricht, Joshua~M Susskind, and Qi Shan.
\newblock Fast and explicit neural view synthesis.
\newblock In {\em Proceedings of the IEEE/CVF Winter Conference on Applications
  of Computer Vision}, pages 3791--3800, 2022.

\bibitem{Bigan}
Swaminathan Gurumurthy, Ravi Kiran~Sarvadevabhatla, and R. Venkatesh~Babu.
\newblock Deligan : Generative adversarial networks for diverse and limited
  data.
\newblock In {\em The IEEE Conference on Computer Vision and Pattern
  Recognition (CVPR)}, 2017.

\bibitem{hamdi2019ian}
Abdullah Hamdi and Bernard Ghanem.
\newblock Ian: Combining generative adversarial networks for imaginative face
  generation.
\newblock {\em arXiv preprint arXiv:1904.07916}, 2019.

\bibitem{mvtn}
Abdullah Hamdi, Silvio Giancola, and Bernard Ghanem.
\newblock Mvtn: Multi-view transformation network for 3d shape recognition.
\newblock In {\em Proceedings of the IEEE/CVF International Conference on
  Computer Vision (ICCV)}, pages 1--11, October 2021.

\bibitem{hamdi2023voint}
Abdullah Hamdi, Silvio Giancola, and Bernard Ghanem.
\newblock Voint cloud: Multi-view point cloud representation for 3d
  understanding.
\newblock In {\em The Eleventh International Conference on Learning
  Representations}, 2023.

\bibitem{point2mesh}
Rana Hanocka, Gal Metzer, Raja Giryes, and Daniel Cohen-Or.
\newblock Point2mesh: A self-prior for deformable meshes.
\newblock {\em arXiv preprint arXiv:2005.11084}, 2020.

\bibitem{arapreg}
Qixing Huang, Xiangru Huang, Bo Sun, Zaiwei Zhang, Junfeng Jiang, and
  Chandrajit Bajaj.
\newblock Arapreg: An as-rigid-as possible regularization loss for learning
  deformable shape generators.
\newblock In {\em Proceedings of the IEEE/CVF International Conference on
  Computer Vision}, pages 5815--5825, 2021.

\bibitem{mvs_transparent}
Jeffrey Ichnowski, Yahav Avigal, Justin Kerr, and Ken Goldberg.
\newblock Dex-nerf: Using a neural radiance field to grasp transparent objects.
\newblock {\em arXiv preprint arXiv:2110.14217}, 2021.

\bibitem{dtu}
Rasmus Jensen, Anders Dahl, George Vogiatzis, Engin Tola, and Henrik Aan{\ae}s.
\newblock Large scale multi-view stereopsis evaluation.
\newblock In {\em Proceedings of the IEEE conference on computer vision and
  pattern recognition}, pages 406--413, 2014.

\bibitem{jeong2022perfception}
Yoonwoo Jeong, Seungjoo Shin, Junha Lee, Chris Choy, Anima Anandkumar, Minsu
  Cho, and Jaesik Park.
\newblock Pe{RF}ception: Perception using radiance fields.
\newblock In {\em Thirty-sixth Conference on Neural Information Processing
  Systems Datasets and Benchmarks Track}, 2022.

\bibitem{vig-reinforce}
Danilo Jimenez~Rezende, S.~M.~Ali Eslami, Shakir Mohamed, Peter Battaglia, Max
  Jaderberg, and Nicolas Heess.
\newblock Unsupervised learning of 3d structure from images.
\newblock In D.~D. Lee, M. Sugiyama, U.~V. Luxburg, I. Guyon, and R. Garnett,
  editors, {\em Advances in Neural Information Processing Systems 29}, pages
  4996--5004. Curran Associates, Inc., 2016.

\bibitem{mvs_tanks}
Arno Knapitsch, Jaesik Park, Qian-Yi Zhou, and Vladlen Koltun.
\newblock Tanks and temples: Benchmarking large-scale scene reconstruction.
\newblock {\em ACM Transactions on Graphics (ToG)}, 36(4):1--13, 2017.

\bibitem{vig-cinvg}
Tejas~D Kulkarni, William~F Whitney, Pushmeet Kohli, and Josh Tenenbaum.
\newblock Deep convolutional inverse graphics network.
\newblock In {\em Advances in neural information processing systems (NIPS)},
  pages 2539--2547, 2015.

\bibitem{spgan}
Ruihui Li, Xianzhi Li, Ka-Hei Hui, and Chi-Wing Fu.
\newblock Sp-gan: Sphere-guided 3d shape generation and manipulation.
\newblock {\em ACM Transactions on Graphics (TOG)}, 40(4):1--12, 2021.

\bibitem{li2022compnvs}
Zuoyue Li, Tianxing Fan, Zhenqiang Li, Zhaopeng Cui, Yoichi Sato, Marc
  Pollefeys, and Martin~R Oswald.
\newblock Compnvs: Novel view synthesis with scene completion.
\newblock In {\em Computer Vision--ECCV 2022: 17th European Conference, Tel
  Aviv, Israel, October 23--27, 2022, Proceedings, Part I}, pages 447--463.
  Springer, 2022.

\bibitem{magic3d}
Chen-Hsuan Lin, Jun Gao, Luming Tang, Towaki Takikawa, Xiaohui Zeng, Xun Huang,
  Karsten Kreis, Sanja Fidler, Ming-Yu Liu, and Tsung-Yi Lin.
\newblock Magic3d: High-resolution text-to-3d content creation.
\newblock In {\em IEEE Conf. Comput. Vis. Pattern Recog.}, 2023.

\bibitem{visionnerf}
Kai-En Lin, Lin Yen-Chen, Wei-Sheng Lai, Tsung-Yi Lin, Yi-Chang Shih, and Ravi
  Ramamoorthi.
\newblock Vision transformer for nerf-based view synthesis from a single input
  image.
\newblock In {\em WACV}, 2023.

\bibitem{Zero-1-to-3}
Ruoshi Liu, Rundi Wu, Basile Van~Hoorick, Pavel Tokmakov, Sergey Zakharov, and
  Carl Vondrick.
\newblock Zero-1-to-3: Zero-shot one image to 3d object.
\newblock {\em arXiv preprint arXiv:2303.11328}, 2023.

\bibitem{neuralvolumes}
Stephen Lombardi, Tomas Simon, Jason Saragih, Gabriel Schwartz, Andreas
  Lehrmann, and Yaser Sheikh.
\newblock Neural volumes: Learning dynamic renderable volumes from images.
\newblock {\em ACM Trans. Graph.}, 38(4):65:1--65:14, July 2019.

\bibitem{mcubes}
William~E. Lorensen and Harvey~E. Cline.
\newblock Marching cubes: A high resolution 3d surface construction algorithm.
\newblock {\em SIGGRAPH Comput. Graph.}, 21(4):163–169, aug 1987.

\bibitem{adamw}
Ilya Loshchilov and Frank Hutter.
\newblock Decoupled weight decay regularization.
\newblock {\em arXiv preprint arXiv:1711.05101}, 2017.

\bibitem{wnerf}
Ricardo Martin-Brualla, Noha Radwan, Mehdi~SM Sajjadi, Jonathan~T Barron,
  Alexey Dosovitskiy, and Daniel Duckworth.
\newblock Nerf in the wild: Neural radiance fields for unconstrained photo
  collections.
\newblock In {\em Proceedings of the IEEE/CVF Conference on Computer Vision and
  Pattern Recognition}, pages 7210--7219, 2021.

\bibitem{RealFusion}
Luke Melas-Kyriazi, Christian Rupprecht, Iro Laina, and Andrea Vedaldi.
\newblock Realfusion: 360$\{$$\backslash$deg$\}$ reconstruction of any object
  from a single image.
\newblock In {\em IEEE Conf. Comput. Vis. Pattern Recog.}, 2023.

\bibitem{occupancynet}
Lars Mescheder, Michael Oechsle, Michael Niemeyer, Sebastian Nowozin, and
  Andreas Geiger.
\newblock Occupancy networks: Learning 3d reconstruction in function space.
\newblock In {\em Proceedings of the IEEE/CVF Conference on Computer Vision and
  Pattern Recognition}, pages 4460--4470, 2019.

\bibitem{text2mesh}
Oscar Michel, Roi Bar-On, Richard Liu, Sagie Benaim, and Rana Hanocka.
\newblock Text2mesh: Text-driven neural stylization for meshes.
\newblock {\em arXiv preprint arXiv:2112.03221}, 2021.

\bibitem{mvs_life}
Ben Mildenhall, Pratul~P Srinivasan, Rodrigo Ortiz-Cayon, Nima~Khademi
  Kalantari, Ravi Ramamoorthi, Ren Ng, and Abhishek Kar.
\newblock Local light field fusion: Practical view synthesis with prescriptive
  sampling guidelines.
\newblock {\em ACM Transactions on Graphics (TOG)}, 38(4):1--14, 2019.

\bibitem{nerf}
Ben Mildenhall, Pratul~P Srinivasan, Matthew Tancik, Jonathan~T Barron, Ravi
  Ramamoorthi, and Ren Ng.
\newblock Nerf: Representing scenes as neural radiance fields for view
  synthesis.
\newblock In {\em European conference on computer vision}, pages 405--421.
  Springer, 2020.

\bibitem{autorf}
Norman M\"uller, Andrea Simonelli, Lorenzo Porzi, Samuel~Rota Bul\`o, Matthias
  Nie{\ss}ner, and Peter Kontschieder.
\newblock Autorf: Learning 3d object radiance fields from single view
  observations.
\newblock In {\em Proceedings of the IEEE/CVF Conference on Computer Vision and
  Pattern Recognition (CVPR)}, pages 3971--3980, June 2022.

\bibitem{instantneural}
Thomas M\"uller, Alex Evans, Christoph Schied, and Alexander Keller.
\newblock Instant neural graphics primitives with a multiresolution hash
  encoding.
\newblock {\em ACM Trans. Graph.}, 41(4):102:1--102:15, July 2022.

\bibitem{polygen}
Charlie Nash, Yaroslav Ganin, SM~Ali Eslami, and Peter Battaglia.
\newblock Polygen: An autoregressive generative model of 3d meshes.
\newblock In {\em International Conference on Machine Learning}, pages
  7220--7229. PMLR, 2020.

\bibitem{girraffe}
Michael Niemeyer and Andreas Geiger.
\newblock Giraffe: Representing scenes as compositional generative neural
  feature fields.
\newblock In {\em Proceedings of the IEEE/CVF Conference on Computer Vision and
  Pattern Recognition}, pages 11453--11464, 2021.

\bibitem{dvr}
Michael Niemeyer, Lars Mescheder, Michael Oechsle, and Andreas Geiger.
\newblock Differentiable volumetric rendering: Learning implicit 3d
  representations without 3d supervision.
\newblock In {\em Proceedings of the IEEE/CVF Conference on Computer Vision and
  Pattern Recognition}, pages 3504--3515, 2020.

\bibitem{deepsdf}
Jeong~Joon Park, Peter Florence, Julian Straub, Richard Newcombe, and Steven
  Lovegrove.
\newblock Deepsdf: Learning continuous signed distance functions for shape
  representation.
\newblock In {\em Proceedings of the IEEE/CVF Conference on Computer Vision and
  Pattern Recognition}, pages 165--174, 2019.

\bibitem{dnerf}
Albert Pumarola, Enric Corona, Gerard Pons-Moll, and Francesc Moreno-Noguer.
\newblock D-nerf: Neural radiance fields for dynamic scenes.
\newblock In {\em Proceedings of the IEEE/CVF Conference on Computer Vision and
  Pattern Recognition}, pages 10318--10327, 2021.

\bibitem{magic123}
Guocheng Qian, Jinjie Mai, Abdullah Hamdi, Jian Ren, Aliaksandr Siarohin, Bing
  Li, Hsin-Ying Lee, Ivan Skorokhodov, Peter Wonka, Sergey Tulyakov, et~al.
\newblock Magic123: One image to high-quality 3d object generation using both
  2d and 3d diffusion priors.
\newblock {\em arXiv preprint arXiv:2306.17843}, 2023.

\bibitem{meshconv}
Anurag Ranjan, Timo Bolkart, Soubhik Sanyal, and Michael~J Black.
\newblock Generating 3d faces using convolutional mesh autoencoders.
\newblock In {\em Proceedings of the European Conference on Computer Vision
  (ECCV)}, pages 704--720, 2018.

\bibitem{pytorch3d}
Nikhila Ravi, Jeremy Reizenstein, David Novotny, Taylor Gordon, Wan-Yen Lo,
  Justin Johnson, and Georgia Gkioxari.
\newblock Accelerating 3d deep learning with pytorch3d.
\newblock {\em arXiv:2007.08501}, 2020.

\bibitem{co3d}
Jeremy Reizenstein, Roman Shapovalov, Philipp Henzler, Luca Sbordone, Patrick
  Labatut, and David Novotny.
\newblock Common objects in 3d: Large-scale learning and evaluation of
  real-life 3d category reconstruction.
\newblock In {\em Proceedings of the IEEE/CVF International Conference on
  Computer Vision (ICCV)}, pages 10901--10911, October 2021.

\bibitem{sharf}
Konstantinos Rematas, Ricardo Martin-Brualla, and Vittorio Ferrari.
\newblock Sharf: Shape-conditioned radiance fields from a single view.
\newblock {\em arXiv preprint arXiv:2102.08860}, 2021.

\bibitem{voxgraf}
Katja Schwarz, Axel Sauer, Michael Niemeyer, Yiyi Liao, and Andreas Geiger.
\newblock Voxgraf: Fast 3d-aware image synthesis with sparse voxel grids.
\newblock In {\em Advances in Neural Information Processing Systems (NeurIPS)},
  2022.

\bibitem{3DFuse}
Junyoung Seo, Wooseok Jang, Min-Seop Kwak, Jaehoon Ko, Hyeonsu Kim, Junho Kim,
  Jin-Hwa Kim, Jiyoung Lee, and Seungryong Kim.
\newblock Let 2d diffusion model know 3d-consistency for robust text-to-3d
  generation.
\newblock In {\em IEEE Conf. Comput. Vis. Pattern Recog.}, 2023.

\bibitem{srn}
Vincent Sitzmann, Michael Zollh{\"o}fer, and Gordon Wetzstein.
\newblock Scene representation networks: Continuous 3d-structure-aware neural
  scene representations.
\newblock {\em Advances in Neural Information Processing Systems}, 32, 2019.

\bibitem{rtmv}
Jonathan Tremblay, Moustafa Meshry, Alex Evans, Jan Kautz, Alexander Keller,
  Sameh Khamis, Charles Loop, Nathan Morrical, Koki Nagano, Towaki Takikawa,
  et~al.
\newblock Rtmv: A ray-traced multi-view synthetic dataset for novel view
  synthesis.
\newblock {\em arXiv preprint arXiv:2205.07058}, 2022.

\bibitem{pixel2mesh}
Nanyang Wang, Yinda Zhang, Zhuwen Li, Yanwei Fu, Wei Liu, and Yu-Gang Jiang.
\newblock Pixel2mesh: Generating 3d mesh models from single rgb images.
\newblock In {\em Proceedings of the European conference on computer vision
  (ECCV)}, pages 52--67, 2018.

\bibitem{ibrnet}
Qianqian Wang, Zhicheng Wang, Kyle Genova, Pratul~P Srinivasan, Howard Zhou,
  Jonathan~T Barron, Ricardo Martin-Brualla, Noah Snavely, and Thomas
  Funkhouser.
\newblock Ibrnet: Learning multi-view image-based rendering.
\newblock In {\em Proceedings of the IEEE/CVF Conference on Computer Vision and
  Pattern Recognition}, pages 4690--4699, 2021.

\bibitem{mvs_google}
Qianqian Wang, Zhicheng Wang, Kyle Genova, Pratul~P Srinivasan, Howard Zhou,
  Jonathan~T Barron, Ricardo Martin-Brualla, Noah Snavely, and Thomas
  Funkhouser.
\newblock Ibrnet: Learning multi-view image-based rendering.
\newblock In {\em Proceedings of the IEEE/CVF Conference on Computer Vision and
  Pattern Recognition}, pages 4690--4699, 2021.

\bibitem{3dim}
Daniel Watson, William Chan, Ricardo~Martin Brualla, Jonathan Ho, Andrea
  Tagliasacchi, and Mohammad Norouzi.
\newblock Novel view synthesis with diffusion models.
\newblock In {\em The Eleventh International Conference on Learning
  Representations}, 2023.

\bibitem{diff3d}
Daniel Watson, William Chan, Ricardo~Martin Brualla, Jonathan Ho, Andrea
  Tagliasacchi, and Mohammad Norouzi.
\newblock Novel view synthesis with diffusion models.
\newblock In {\em The Eleventh International Conference on Learning
  Representations}, 2023.

\bibitem{deephybridmesh}
Xingkui Wei, Zhengqing Chen, Yanwei Fu, Zhaopeng Cui, and Yinda Zhang.
\newblock Deep hybrid self-prior for full 3d mesh generation.
\newblock In {\em Proceedings of the IEEE/CVF International Conference on
  Computer Vision}, pages 5805--5814, 2021.

\bibitem{opengl}
Mason Woo, Jackie Neider, Tom Davis, and Dave Shreiner.
\newblock {\em OpenGL Programming Guide: The Official Guide to Learning OpenGL,
  Release 1}.
\newblock Addison-wesley, 1998.

\bibitem{modelnet}
Zhirong Wu, S. Song, A. Khosla, Fisher Yu, Linguang Zhang, Xiaoou Tang, and J.
  Xiao.
\newblock 3d shapenets: A deep representation for volumetric shapes.
\newblock In {\em 2015 IEEE Conference on Computer Vision and Pattern
  Recognition (CVPR)}, pages 1912--1920, 2015.

\bibitem{mvs_sapian}
Fanbo Xiang, Yuzhe Qin, Kaichun Mo, Yikuan Xia, Hao Zhu, Fangchen Liu, Minghua
  Liu, Hanxiao Jiang, Yifu Yuan, He Wang, et~al.
\newblock Sapien: A simulated part-based interactive environment.
\newblock In {\em Proceedings of the IEEE/CVF Conference on Computer Vision and
  Pattern Recognition}, pages 11097--11107, 2020.

\bibitem{mvs_robi}
Jun Yang, Yizhou Gao, Dong Li, and Steven~L Waslander.
\newblock Robi: A multi-view dataset for reflective objects in robotic
  bin-picking.
\newblock In {\em 2021 IEEE/RSJ International Conference on Intelligent Robots
  and Systems (IROS)}, pages 9788--9795. IEEE, 2021.

\bibitem{foldingnet}
Yaoqing Yang, Chen Feng, Yiru Shen, and Dong Tian.
\newblock Foldingnet: Point cloud auto-encoder via deep grid deformation.
\newblock In {\em Proceedings of the IEEE conference on computer vision and
  pattern recognition}, pages 206--215, 2018.

\bibitem{mvs_blender}
Yao Yao, Zixin Luo, Shiwei Li, Jingyang Zhang, Yufan Ren, Lei Zhou, Tian Fang,
  and Long Quan.
\newblock Blendedmvs: A large-scale dataset for generalized multi-view stereo
  networks.
\newblock In {\em Proceedings of the IEEE/CVF Conference on Computer Vision and
  Pattern Recognition}, pages 1790--1799, 2020.

\bibitem{volsurf}
Lior Yariv, Jiatao Gu, Yoni Kasten, and Yaron Lipman.
\newblock Volume rendering of neural implicit surfaces.
\newblock {\em Advances in Neural Information Processing Systems},
  34:4805--4815, 2021.

\bibitem{plennerf}
Alex Yu, Ruilong Li, Matthew Tancik, Hao Li, Ren Ng, and Angjoo Kanazawa.
\newblock {PlenOctrees} for real-time rendering of neural radiance fields.
\newblock In {\em ICCV}, 2021.

\bibitem{pixelnerf}
Alex Yu, Vickie Ye, Matthew Tancik, and Angjoo Kanazawa.
\newblock pixelnerf: Neural radiance fields from one or few images.
\newblock In {\em Proceedings of the IEEE/CVF Conference on Computer Vision and
  Pattern Recognition}, pages 4578--4587, 2021.

\bibitem{vig-inverse-render-net}
Ye Yu and William~AP Smith.
\newblock Inverserendernet: Learning single image inverse rendering.
\newblock {\em arXiv preprint arXiv:1811.12328}, 2018.

\bibitem{ners}
Jason Zhang, Gengshan Yang, Shubham Tulsiani, and Deva Ramanan.
\newblock Ners: Neural reflectance surfaces for sparse-view 3d reconstruction
  in the wild.
\newblock {\em Advances in Neural Information Processing Systems}, 34, 2021.

\bibitem{lpips}
Richard Zhang, Phillip Isola, Alexei~A Efros, Eli Shechtman, and Oliver Wang.
\newblock The unreasonable effectiveness of deep features as a perceptual
  metric.
\newblock In {\em Proceedings of the IEEE conference on computer vision and
  pattern recognition}, pages 586--595, 2018.

\bibitem{mvs_real}
Tinghui Zhou, Richard Tucker, John Flynn, Graham Fyffe, and Noah Snavely.
\newblock Stereo magnification: Learning view synthesis using multiplane
  images.
\newblock {\em arXiv preprint arXiv:1805.09817}, 2018.

\end{thebibliography}
} 
\clearpage \clearpage 
\appendix
\section{Detailed Formulations} \label{secsup:meth}
\subsection{Sparse Convolutions}
Sparse convolutions are a variant of standard convolutions that are used in deep learning. In a sparse convolution, only a subset of the input elements is used in the computation, which allows for more efficient use of computation resources and can improve the performance of the convolutional neural network.
To perform a sparse convolution, we first define a set of indices that specify which input elements should be used in the convolution. This set of indices is called the "support" of the convolution. We then use these indices to select the relevant input elements and compute the convolution using these elements. This is typically done by applying a filter to the selected input elements and summing the results to produce the output of the convolution. 

In the simplest 1 D case, let $\mathbf{x}$ be the input tensor, $\mathbf{w}$ be the convolutional filter, and $c$ be the support of the convolution (i.e. the set of indices specifying which elements of $x$ should be used in the convolution). The output of the sparse convolution, $\mathbf{y}$, can be computed as: $\mathbf{y} = \mathbf{x}[c] * \mathbf{w}$
where $*$ denotes the convolution operation, and $\mathbf{x}[c]$ is the subset of elements from $\mathbf{x}$ specified by the support $c$. This equation applies the convolutional filter $\mathbf{w}$ to the selected input elements and sums the results to produce the output of the convolution. For more detailed formulation and implementation of the Sparse convolutions we used in our work, please refer to MinkowskiNetwork \cite{minkosky}.

\subsection{Q-Gaussian loss sampling} 
In the space of sparse voxels of high resolution, defining \textit{where} the loss is sampled is difficult, especially if the output topology is unknown.
One of the challenges in working with sparse voxels of high resolution (\eg 512) is that training can not involve  densifying the voxels to the original resolution (\ie 512$^3$) due to prohibitive memory requirements. The input/output topologies are not necessarily the same, as the sparse convolutional strides and pruning can alter the sparse voxels’ coordinates. This is why the sampling function $\mathcal{S}$ in \eqLabel{\ref{eq:density}} is of utmost importance in guiding the training of SuRFNet. We sample at random coordinates centered at the center of the voxel grid $\mathcal{S}:~ \mathbf{c} \sim \mathcal{Q}\left( \mathcal{N}( \frac{\mathbf{H}}{2},\, \frac{\mathbf{H}\sigma^{2}}{2}\mathbb{I} )\right) $, where $\mathbf{H} = (H,H,H)$ is the voxel grid resolution vector, $\mathbb{I}$ is the identity matrix, $\sigma$ is a hyperparameter determining the spread of the loss, and $\mathcal{Q}: \mathbb{R}^3 \rightarrow {\mathbb{Z}^{+}}^{3}$ is the quantization-and-cropping function of coordinates that ensure the output coordinates are integers within bounds $\mathbf{c} \in [0,1,...,H-1]^{3}$.  We discuss more details about $\mathcal{S}$ and alternative configurations in \secLabel{\ref{sec:sup-loss}}. Simply put, the Q-Gaussian is 3D normal distribution quantized to integer coordinates to give prior about where the output is expected and where the loss is defined. 

\section{Detailed Setup} \label{secsup:setup}
\subsection{SPARF Dataset}
All the rendered images are of $400\times 400$ resolution with 4 channels (RGB + alpha channel for background). SPARF has three main splits for every 3D shape: training views (400 views), test views (20 views), and an OOD “hard" views (10 views) as can be shown in \figLabel{\ref{fig:ood}}. Regarding the collected SRFs, Plenoxels \cite{plenoxels} is used as the base module. The spherical harmonics dimension of the whole SRFs is $d=4\times 3= 12$, while for partial SRFs, it is $d=1\times 3 = 3$. Most of the hyperparameters used in optimizing the SRFs are the default ones proposed in the Plenoxels paper \cite{plenoxels} (as can be seen in the attached code under Svox2/opt/opt-py). However, the following hyperparameters were engineered in order to scale up the optimization and maintain the quality of the SRFs ( as can be seen in \figLabel{\ref{fig:srf1}, and \ref{fig:srf2}}). The flickering temporal noise introduced in the $32^3$ resolution SRFs is due to the extremely low number of voxels representing the radiance fields while the views are rendered densely from the spiral sequence, hence aliasing occurs. 

Running Plenoxels \cite{plenoxels} for fewer iterations (3$\times$12K) reduces the time by 30\% while maintaining the same PSNR. Using 400 views/shapes in SPARF to optimize the SRFs keep the time manageable in optimization ($\sim 4$ minutes for the 512 resolution) while maintaining high PSNR ($\sim 30$dB). When saving the SRFs, we only save the set of coordinates (integers) and float features (densities and radiance components). The upsampling iteration of Plenoxels is set to 1$\times$12K for faster convergence. The distribution of the collected dataset is detailed in Table \ref{tbl:anatomy}. More examples of the whole \vs partial SRFs collected in SPARF can be found in \figLabel{\ref{figsup:partials}}. A total of 200K GPU hours are used in the optimization process to collect SPARF. The whole SRFs are easily convertible to high-quality meshes using Marching Cubes \cite{mcubes} as shown in \figLabel{\ref{figsup:mesh}}. While the SPARF dataset is indeed larrge in total ( 3.4 TB), its posed-images part is only 360 GB, which makes it manageble for training on other applications that require dense posed images. 

\begin{table}[t]
\footnotesize
\setlength{\tabcolsep}{6pt} %
\renewcommand{\arraystretch}{1.1} %
\centering
\resizebox{0.95\hsize}{!}{
\begin{tabular}{ccc|c} 
\toprule
 \multicolumn{1}{c}{\textbf{SRF}}& \multicolumn{1}{c}{\textbf{Voxewl}}& \multicolumn{1}{c|}{\textbf{Nb. of}}&   \multicolumn{1}{c}{\textbf{Nb. of}}\\
 \textbf{Type} & \textbf{Resolution}  & \textbf{Variants} &  \textbf{SRFs}     \\
\midrule
 \multirow{6}{*}{Partial} & \multirow{2}{*}{32} &  $4\times$1-view  &  158,816 \\ 
  & &  $4\times$3-view  &  158,816  \\ 
 & \multirow{2}{*}{128} & $4\times$1-view  &  158,816   \\ 
 & &  $4\times$3-view   &  158,816  \\  
 & \multirow{2}{*}{512} & $4\times$1-view  &  158,816  \\ 
 & &  $4\times$1-view   & 158,816 \\  \hline

 \multirow{3}{*}{Whole} & \multirow{1}{*}{32} &  $1\times$400-view  &  39,704 \\
 & \multirow{1}{*}{128} & $1\times$400-view  &  39,704   \\  
 & \multirow{1}{*}{512} & $1\times$400-view  &  39,704  \\ 
   \midrule
  Total & - &  -   &  1,072,008  \\ 
   
 \bottomrule
\end{tabular}
}
\caption{\small \textbf{SPARF Anatomy}. We show the distribution of the one million SRFs collected in SPARF between multiple resolutions and between whole and partial SRFS.}
\label{tbl:anatomy}
\end{table}
\begin{figure}[t]
    \centering
    \tabcolsep=0.03cm
\resizebox{0.85\linewidth}{!}{
\begin{tabular}{ccc}
Whole SRF & Partial (3 Views)  & Partial (1 View) \\
     \includegraphics[trim= 0.0cm 0cm 0cm 0cm,clip, width=0.33\linewidth]{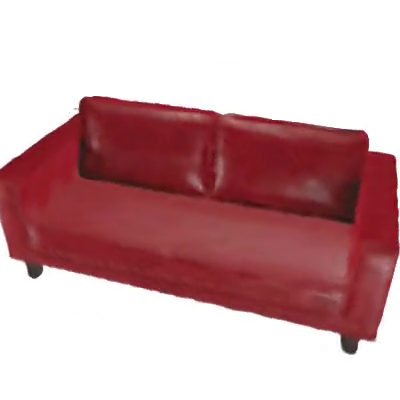}  &
     \includegraphics[trim= 0.0cm 0cm 0cm 0cm,clip, width=0.33\linewidth]{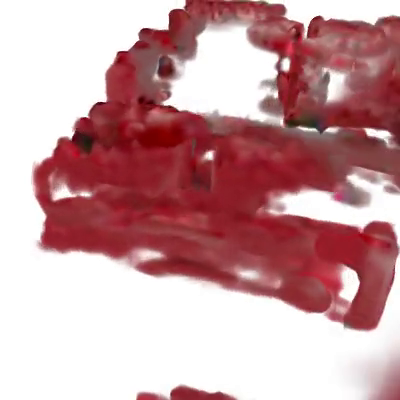}  &
     \includegraphics[trim= 0.0cm 0cm 0cm 0cm,clip, width=0.33\linewidth]{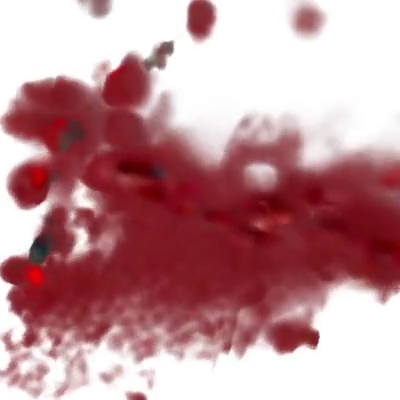}  \\
     \includegraphics[trim= 0.0cm 0cm 0cm 0cm,clip, width=0.33\linewidth]{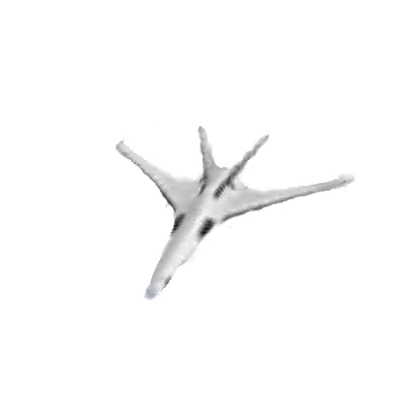}  &
     \includegraphics[trim= 0.0cm 0cm 0cm 0cm,clip, width=0.33\linewidth]{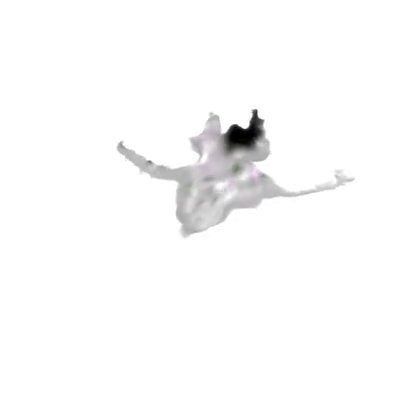}  &
     \includegraphics[trim= 0.0cm 0cm 0cm 0cm,clip, width=0.33\linewidth]{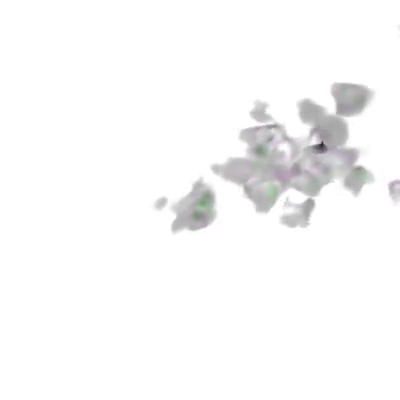}  \\
     \includegraphics[trim= 0.0cm 0cm 0cm 0cm,clip, width=0.33\linewidth]{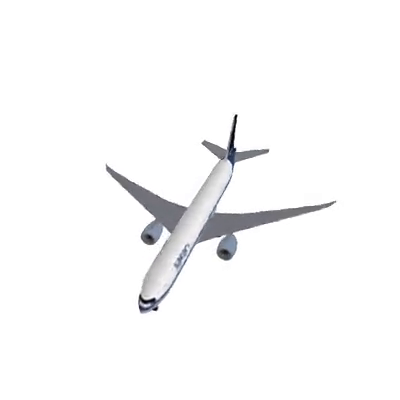}  &
     \includegraphics[trim= 0.0cm 0cm 0cm 0cm,clip, width=0.33\linewidth]{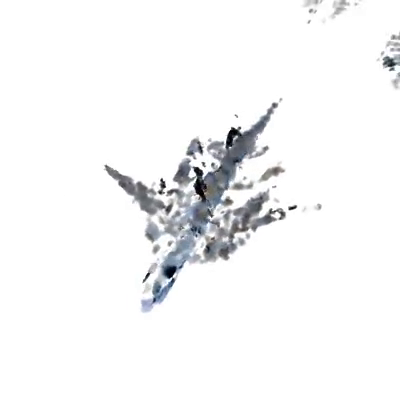}  &
     \includegraphics[trim= 0.0cm 0cm 0cm 0cm,clip, width=0.33\linewidth]{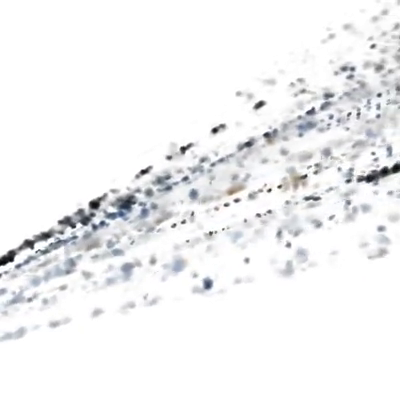}  \\

\bottomrule
\end{tabular}
} %
    \caption{\textbf{Whole \vs Partial SRFs}. The partial SRFs are used instead of the few images that generated them as input to the learning pipeline to generate the whole SRFs 
    }
    \label{figsup:partials}
\end{figure}
\begin{figure}[t]
    \centering
    \tabcolsep=0.03cm
\resizebox{0.9\linewidth}{!}{
\begin{tabular}{ccc|c}
\multicolumn{3}{c}{SRF Renderings } &  Extracted Mesh   \\
     \includegraphics[trim= 0.0cm 1cm 0cm 0cm,clip, width=0.25\linewidth]{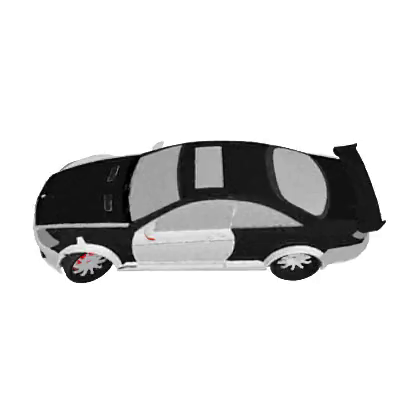} &
     \includegraphics[trim= 0.0cm 1cm 0cm 0cm,clip, width=0.25\linewidth]{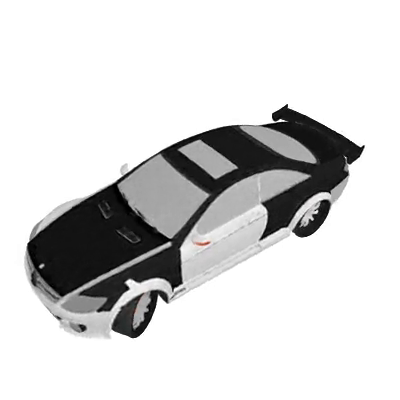}  &
     \includegraphics[trim= 0.0cm 1cm 0cm 0cm,clip, width=0.25\linewidth]{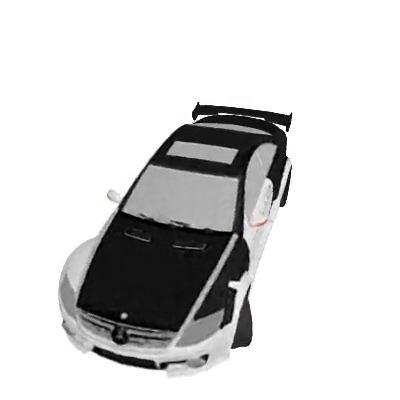}  &
     \includegraphics[trim= 0.0cm 1cm 0cm 0cm,clip, width=0.25\linewidth]{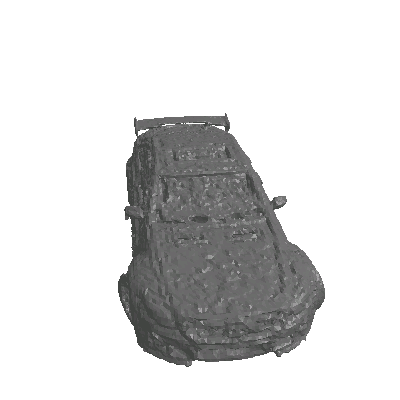}  \\

     \includegraphics[trim= 0.0cm 1cm 0cm 0cm,clip, width=0.25\linewidth]{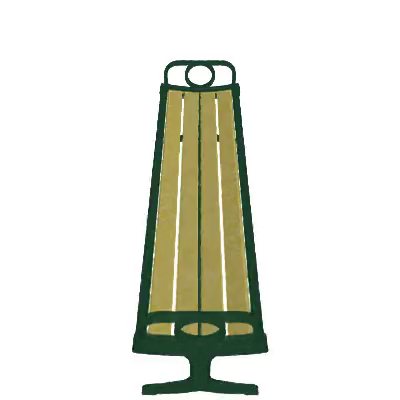} &
     \includegraphics[trim= 0.0cm 1cm 0cm 0cm,clip, width=0.25\linewidth]{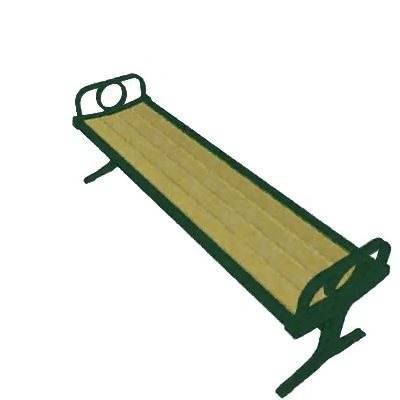}  &
     \includegraphics[trim= 0.0cm 1cm 0cm 0cm,clip, width=0.25\linewidth]{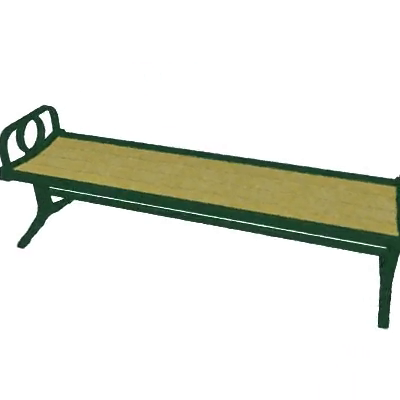}  &
     \includegraphics[trim= 0.0cm 1cm 0cm 0cm,clip, width=0.25\linewidth]{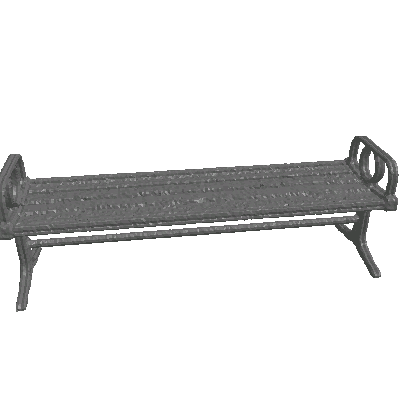}  \\

     \includegraphics[trim= 0.0cm 1cm 0cm 0cm,clip, width=0.25\linewidth]{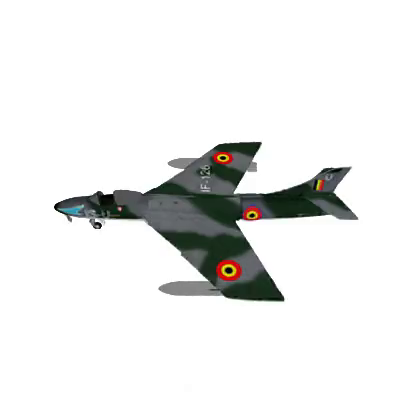} &
     \includegraphics[trim= 0.0cm 1cm 0cm 0cm,clip, width=0.25\linewidth]{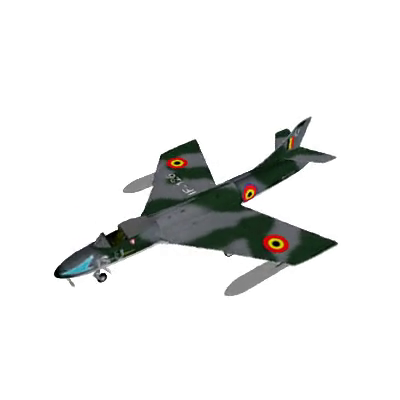}  &
     \includegraphics[trim= 0.0cm 1cm 0cm 0cm,clip, width=0.25\linewidth]{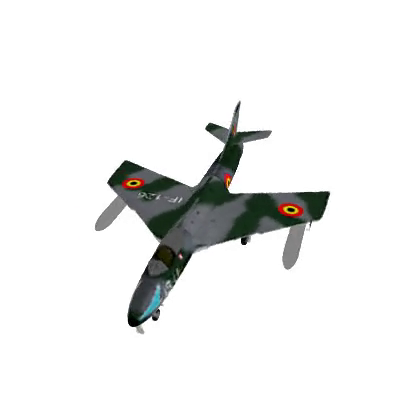}  &
     \includegraphics[trim= 0.0cm 1cm 0cm 0cm,clip, width=0.25\linewidth]{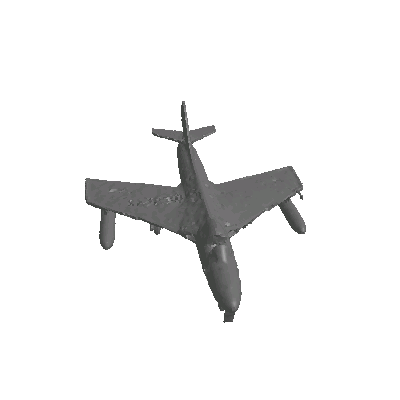}  \\

     \includegraphics[trim= 0.0cm 1cm 0cm 0cm,clip, width=0.25\linewidth]{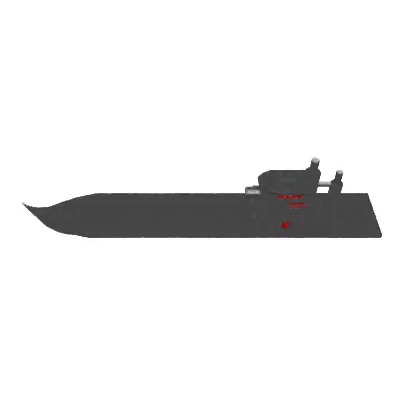} &
     \includegraphics[trim= 0.0cm 1cm 0cm 0cm,clip, width=0.25\linewidth]{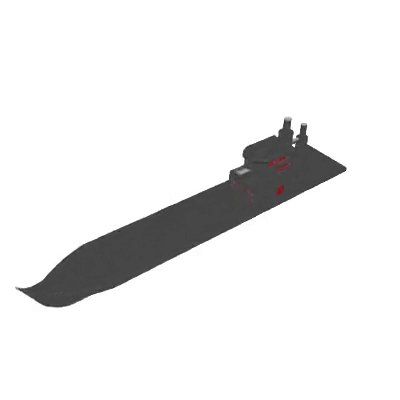}  &
     \includegraphics[trim= 0.0cm 1cm 0cm 0cm,clip, width=0.25\linewidth]{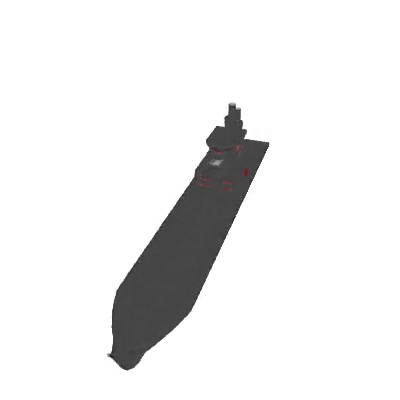}  &
     \includegraphics[trim= 0.0cm 1cm 0cm 0cm,clip, width=0.25\linewidth]{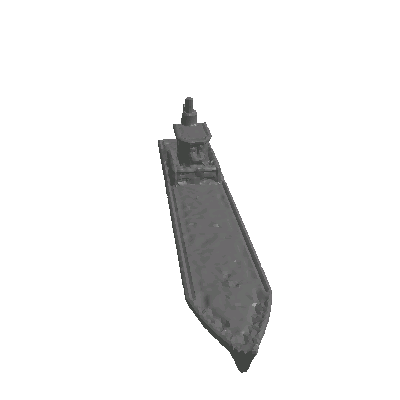}  \\

\bottomrule
\end{tabular}
} %
    \caption{\textbf{Extracting 3D Meshes from SRFs}. Since SPARF and SuRFNet live on the 3D voxel's space, extracting the mesh is straightforward with one pass of MarchingCubes \cite{mcubes}.   
    }
    \label{figsup:mesh}
\end{figure}
\subsection{SuRFNet Training}
We use a voxel resolution of $128^3$ of the SPARF dataset in most of the learning experiments and visualizations in this work, unless otherwise clearly stated. This choice is to reduce the computational cost of training the heavy pipeline and to facilitate the development of proper learning methods on SRFs. The input SRF is normalized with a fixed value of 10,0000 for the density and 10 for the colors, to ensure the distribution lies within -1 to 1. The Q-Gaussian std $\sigma$ is set to 0.444 (studied more in \secLabel{\ref{secsup:analysis}}). The strides for the SuRFNEt are all set 2, while the network depth $l=3$ modules. The batch size for training is 14 when A100 GPUs are used and 6 when V100 GPUs are used. The training saturates at 100 epochs. The optimizer used is AdamW \cite{adamw} with a learning rate of 0.01, a momentum of 0.9, a weight decay of $1e-5$, and a learning rate exponential decay rate of 0.99. The hyperparameters $\lambda_{R}, \lambda_{\alpha}, \lambda_{\rho}$ are all set independently to each class, where a different network is trained on each class separately. Most classes have $\lambda_{\alpha} = 30.0 ,\lambda_{\rho}=1.0, \lambda_{R}  $. We did not prune the output sparse voxel as this leads to harming performance most of the time and increase the problem of vanishing gradients. The background color of the rendered images $\mathcal{R}_{\phi} \left( \mathbf{F}(\mathcal{X})\right)$ is masked out from the perceptual loss and the density component $\alpha$ of the output SRF is also not affected by the perceptual loss, as this can cause excessive densities around the object, leading to deteriorating the SRF output perceptuality. During training with the perceptual loss, three randomly selected images from three different $\phi$ as used as labels for the three rendered images from the output $\mathcal{R}_{\phi} \left( \mathbf{F}(\mathcal{X})\right)$. The SuRFNet is jointly predicting the density and radiance of spherical harmonics colors, but with different heads. More setup details can be found in the attached code and analyzed further in \secLabel{\ref{secsup:analysis}}. We train a separate model for each class, to maintain high-quality generation of 3D SRFs.

\mysection{Retraining Baselines}
To compare to the preprinted baselines PixelNeRF and VisionNerf ( which use $64 \times 64$ resolution), we upsample their resolution at inference at test poses while using their pretrained weights of the NMR dataset. The upsampling is using the bicubic sampling of the Pytorch Transforms library. These baselines are used in this works by default unless otherwise specified.

Retraining the methods from scratch on the high resolution $400 \times 400$ is computationally expensive. 
To illustrate the retraining cost, VisionNeRF \cite{visionnerf} was originally trained on \textit{16 A100 GPUs for a week} just to converge on $64\times64$ resolution. SPARF's $400\times400$ resolution images would need $\sim$ 39 times as much time/compute due to per-pixel sampling. In contrast, our SuRFNet was trained on \textit{a single V100 GPU} for 3 days, which allows for fine-tuning of the learning pipeline for quick convergence. 
However, for a fair comparison, We retrain PixelNeRF \cite{pixelnerf} on the 13 categories and report the 3-view and 1-view PSNR results in Table 2. We see that our SuRFNet indeed surpasses this baseline trained on the same SPARF data. The results of retraining are not very different from the pretrained weights (slight improvement) because training on SPARF high-resolution images is unstable using these 2D-based NeRF networks that need a per-pixel sampling, which diverges the training in some cases.  The full results are shown in Table \ref{tbl:test}

\section{Additional Analysis} \label{secsup:analysis}
\begin{table*}[]
\centering
\resizebox{1.0\linewidth}{!}{
\tabcolsep=0.12cm
\begin{tabular}{l|ccccccccccccc|c}
\toprule
 & \multicolumn{13}{c}{\textbf{SPARF Classes} } & \\
\textbf{Baselines} & chair & watercraft & rifle & display & lamp & speaker & cabinet & bench & car & airplane & sofa & table & phone& mean \\

\midrule
Plenoxels \cite{plenoxels} (1V)   &  10.1  &  12.1  &  12.6  &  8.7  &      14.7  &  8.7  &      10.9  &  11.4  &  7.7  &      14.0  &  9.7  &      10.5  &  9.5   &  10.8 \\ \rowcolor[HTML]{EFEFEF} 
Plenoxels \cite{plenoxels} (3V)   &  10.8  &  13.3  &  15.6  &  9.7  &      16.2  &  10.1  &  12.1  &  12.1  &  9.0   &  15.4  &  11.4  &  10.8  &  10.2  &  12.1 \\ 
PixelNeRF \cite{pixelnerf} (1V)   &  10.8  &  14.1  &  14.2  &  9.0  &      15.6  &  9.2   &  10.5  &  12.4  &  10.1  &  15.7  &  11.1  &  10.6  &  11.1  &  11.9     \\ \rowcolor[HTML]{EFEFEF} 
PixelNeRF \cite{pixelnerf} (3V)   &  11.0  &  14.1  &  14.2  &  9.3   &  15.7  &  9.4  &      10.6  &  12.7  &  10.1  &  15.7  &  11.3  &  10.9  &  11.4  &  12.0 \\
PixelNerf $\mathbf{^*}$ (1V)   & 13.8	& 12.2	& 15.0	& 17.5	& 19.0	& 11.2	& 17.5	& 13.3	& 12.2	& 17.9	& 11.3	& 11.7	& 14.7	& 14.5    \\ \rowcolor[HTML]{EFEFEF} 
PixelNerf $\mathbf{^*}$ (3V)   & 17.5	& 13.6	& 16.2	& 11.7	& 20	& 15.6	& 13.1	& 17.6	& 12.1	& 18.2	& 16.2	& 12.1	& 10.7	& 15.0    \\
VisionNeRF \cite{visionnerf} (1V) &  16.5  &  18.4  &  18.5  &  15.1  &  19.3  &  13.2  &  16.1  &  16.3  &  13.8  &  21.8  &  15.1  &  14.8  &  14.0  &  16.4  \\ \midrule \rowcolor[HTML]{EFEFEF} 
\textbf{SuRFNet (ours) (1V)}      & 15.7	& 15.5 & 	19.1 & 	14.1 & 	18.5 &	14.5 &	18.7 & 	15.6 & 	18.1 & 	20.3 & 	16.3 & 	14.1 & 	17.4 & 	16.8    \\
\textbf{SuRFNet (ours) (3V)}      & \textbf{18.6}	& \textbf{20.7} & 	\textbf{20.9} & 	\textbf{17.1} & 	\textbf{21.2} &	\textbf{18.5} &	\textbf{21.7} & 	\textbf{17.6} & 	\textbf{18.9} & 	\textbf{21.9} & 	\textbf{20.4} & 	\textbf{16.7} & 	\textbf{20.0} & 	\textbf{19.5}  \\

\bottomrule
\end{tabular}
} %
\caption{\small \textbf{SPARF Benchmark on Novel View Synthesis (Normal Test)}. We compare the validation PSNR of some of the widely used novel view synthesis techniques on the SPARF dataset for the generalization of novel view synthesis beyond a single example and on the normal testing-views tracks similar to the ones seen in training views. One view (1V) and three views (3V) inputs are reported, and $\mathbf{^*}$ indicates retraining the baseline backbone on the high-resolution images of the SPARF dataset.}
\vspace{-4mm}
    \label{tbl:test}
\end{table*}
\subsection{Shiny Objects}
\begin{figure}
    \centering
    \includegraphics[trim= 2cm 2cm 2cm 3cm,clip, width=0.499\linewidth]{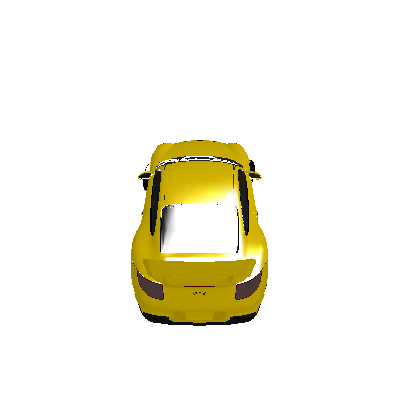}\includegraphics[trim= 1.0cm 1cm 0.0cm 2.0cm,clip, width=0.5\linewidth]{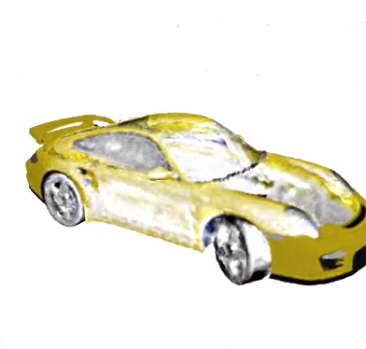} 
    \caption{\textbf{Shiny Objects Corrupts SRFs:}. Optimizing SRFs on shiny objects with a reflective material (\textit{left}) results in distorted radiance fields (\textit{right}). These distorted SRFs (of 76 shapes in total) were separated from the main classes in SPARF.
    }
    \label{fig:shiny}
\end{figure}

Some of the rendered objects have reflective materials, resulting in distorted optimized radiance fields for these shapes despite using all of the views. We separate these distorted SRFs (only 76 shapes in total) from the SPARF dataset (see \figLabel{\ref{fig:shiny}}). 

\subsection{Effect of Dataset Size} \label{sec:sup-datasetsize}
\begin{figure}
    \centering
    \includegraphics[trim= 0.2cm 0cm 0.2cm 0cm,clip, width=0.98\linewidth]{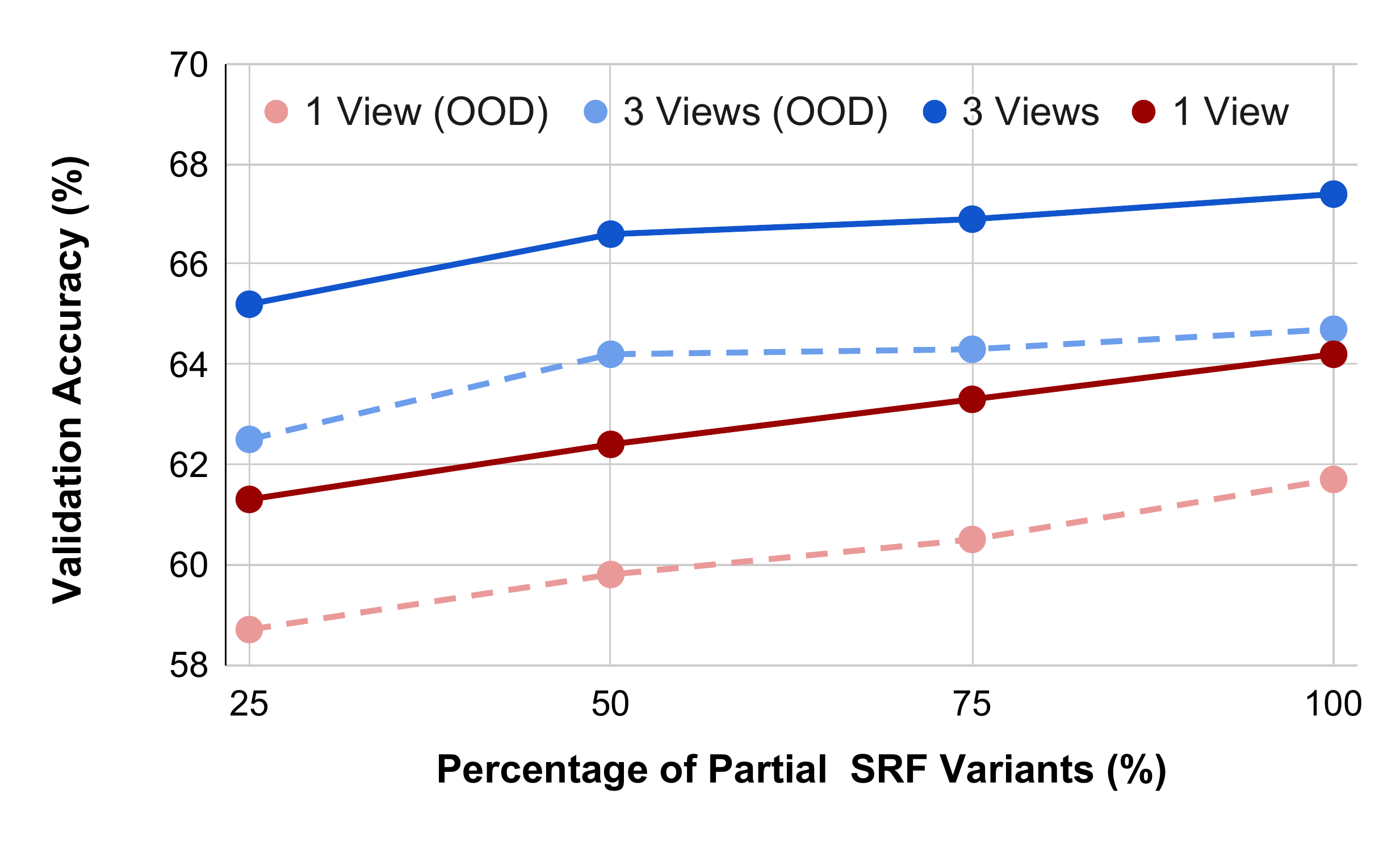}
    \caption{\textbf{Scaling-Up Training on SRFs: Partial SRFs}. As the training data (partial SRFs) of radiance fields increase, the generalization improves, as can be seen in the car class here. The 3-view and 1-view metrics are reported with test and OOD metrics.
    }
    \label{fig:sup-variants}
\end{figure}

\begin{figure}
    \centering
    \includegraphics[trim= 0.2cm 0cm 0.2cm 0cm,clip, width=0.98\linewidth]{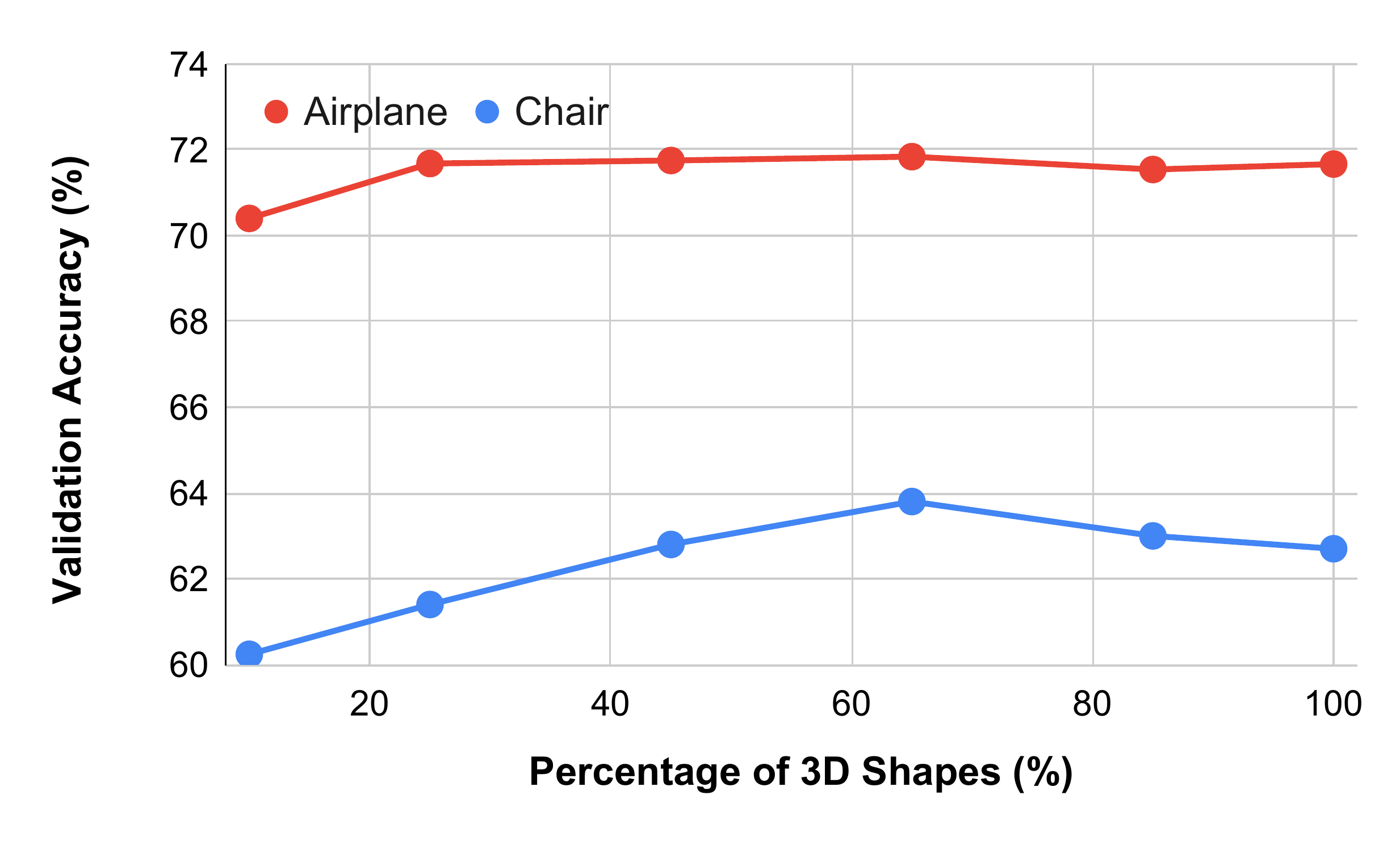}
    \caption{\textbf{Scaling-Up Training on SRFs: Whole SRFs}. As the training data of radiance fields increase, the generalization improves across different classes in SPARF. 
    }
    \label{fig:scale}
\end{figure}

We study the effect of increasing the dataset size (Whole SRFs and Partial SRFs) on the generalization performance of SuRFNet in \figLabel{\ref{fig:scale},\ref{fig:sup-variants}}. It shows that as the dataset size increase ( normalized the number of shapes in each class ), the generalization performance increase. This scalability effect underlines the importance of SPARF. However, as can be seen from these two figures, partial SRFs scalability is more important than increasing whole SRFs, which justifies collecting 4 variants per resolution ( as detailed in Table \ref{tbl:anatomy}). This observation aligns with previous generative models in the literature \cite{Bigan,diff3d,hamdi2019ian} 

\subsection{Loss Ablation Study} \label{sec:sup-loss}
\mysection{Loss Sampling}  %
We study the effect of the sampling strategy with a different number of input images at test time on the performance of SuRFNet in Table \ref{tbl:views}. It shows that using a uniform sampling strategy depletes the learning capacity of the network and can degrade performance. The effect is more evident when the number of views is one, where the partial SRFs are more sparse and the training is delicate.
\begin{table}[t]
\tabcolsep=0.10 cm
\centering
\resizebox{0.95\linewidth}{!}{
\begin{tabular}{lccccccc} 
\toprule
& \multicolumn{3}{c}{\textbf{1-view}} & & \multicolumn{3}{c}{\textbf{3-view}} \\ \cmidrule{2-4} \cmidrule{6-8}
 \multicolumn{1}{c}{\textbf{Strategy}} & PSNR$\uparrow$ & SSIM$\uparrow$ & LPIPS$\downarrow$ & & PSNR$\uparrow$ & SSIM$\uparrow$ & LPIPS$\downarrow$ \\ 
\midrule
uniform & 20.03 & 0.94 &  0.10 & &  21.00 &  0.94 &  0.09 \\
Q-Gaus. & 20.55 & 0.93 &  0.09 & &  21.83 &  0.94 &  0.08 \\ 
\bottomrule
\end{tabular}
}
\caption{\small \textbf{Effect of Loss Sampling Strategy.}We study the effect of Loss sampling strategy (uniform \vs Q-Gaussian) on airplane class.}
\label{tbl:views}
\end{table}
\mysection{Loss Hyperparameters}  %
For the density threshold $\alpha_{\text{dense}}$ defined in Eq 2 and 3, the validation accuracies of SuRFNet on car class are 13, 14.7, 67.8, 67, 67, 67.2, 62.5 for the values of -0.01, -0.001, 0, 0.001, 0.003, 0.01, 0.03 of $\alpha_{\text{dense}}$  respectively. The hyperparameter $\sigma$ which governs the spread of the Q-Gaussian loss is studied as follows. the validation accuracies of SuRFNet on airplane class are 52, 70.4, 71.7, 72.1, 72.2, 72.4, 72.3 for the values of 0.1, 0.2, 0.3, 0.4, 0.6, 0.8, and 1.0 of $\sigma$ respectively. The number of coordinates $\mathbf{c}$ sampled in the Q-Gaussian loss is proportional to the number of coordinates in the input SRFs with multiplier $K=40$. For different values of this multiplier 1, 5, 10, 20, 40, 80, 200, the validation accuracies of SuRFNet trained on airplane class are 72.2, 72.7, 72.9, 72.5, 71.8, 71.7, and 71.6 respectively.

\subsection{Faulty Textures}
\begin{figure}
    \centering
    \includegraphics[trim= 0cm 0cm 0cm 0cm,clip, width=0.45\linewidth]{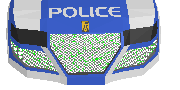}     \includegraphics[trim= 0cm 0cm 0cm 0cm,clip, width=0.4\linewidth]{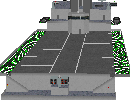} 
    \caption{\textbf{Rare Cases of Faulty Textures}. Some objects in ShapeNet \cite{shapenet} have doubled textrures in some parts, leading to faulty renderings.
    }
    \label{fig:textures}
\end{figure}

In some rare instance of the shapes in ShapeNet \cite{shapenet}, some objects have doubled textures in some areas. This occurs in less than 1\% of the data and leads the renderer to render the background instead in these areas (highlighted with green). See \figLabel{\ref{fig:textures}} for examples of these cases.

\begin{figure}
    \centering
    \tabcolsep=0.03cm
\resizebox{0.9\linewidth}{!}{
\begin{tabular}{ccc}
output w/o $L_R$ & output w/ $L_R$  & whole SRF  \\
                    \includegraphics[trim= 0.0cm 0cm 0cm 0cm,clip, width=0.31\linewidth]{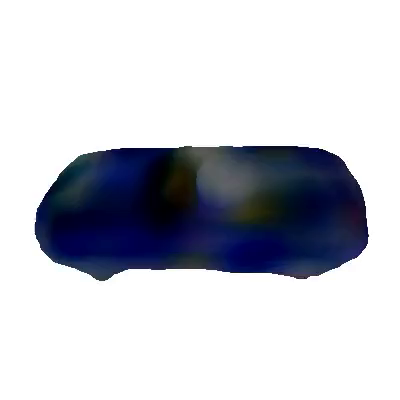} & \includegraphics[trim= 0.0cm 0cm 0cm 0cm,clip, width=0.31\linewidth]{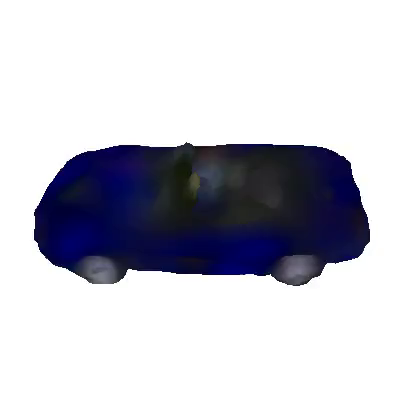}   & \includegraphics[trim= 0.0cm 0cm 0cm 0cm,clip, width=0.31\linewidth]{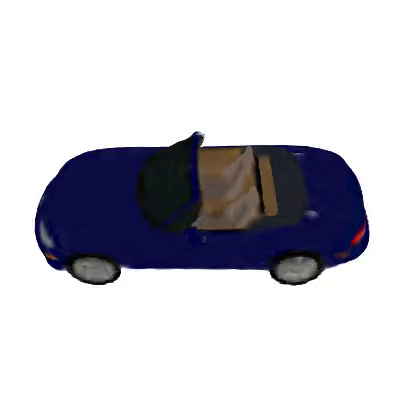}  \\
\bottomrule
\end{tabular}
}%
    \caption{\textbf{Effect of the Perceptual Loss $L_R$}. Adding a perceptual loss on volume-rendered images during training SuRFNet insures the rendered images remain closer to how they should be rendered, as the 3D radiance colors supervision won't guarantee the rendering quality. \textit{(left)}: without perceptual loss , \textit{(middle)}: with the loss.
    }
    \label{figsup:loss}
\end{figure}
\subsection{The Irregularity of SRFs}
\begin{figure}
    \centering
    \tabcolsep=0.03cm
\resizebox{0.9\linewidth}{!}{
\begin{tabular}{ccc}
\includegraphics[trim= 0.0cm 0cm 0cm 0cm,clip, width=0.31\linewidth]{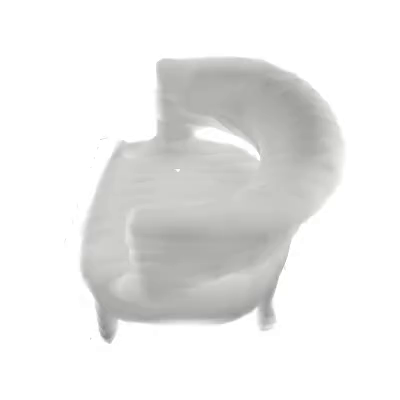} &
\includegraphics[trim= 0.0cm 0cm 0cm 0cm,clip, width=0.31\linewidth]{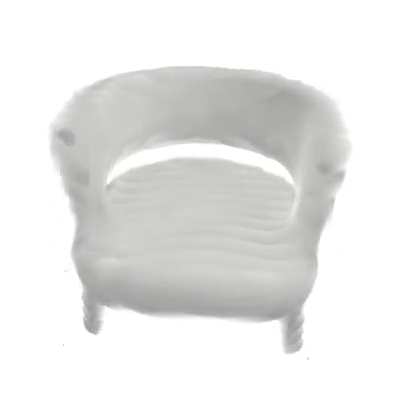}   &
\includegraphics[trim= 0.0cm 0cm 0cm 0cm,clip, width=0.31\linewidth]{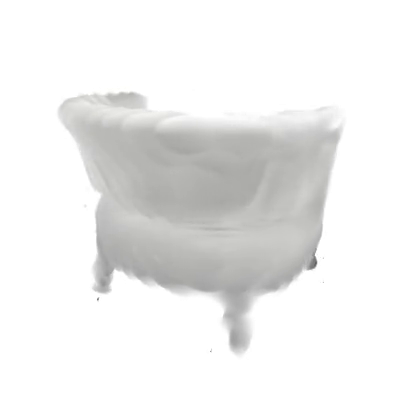}  \\ \midrule

\includegraphics[trim= 0.0cm 0cm 0cm 0cm,clip, width=0.31\linewidth]{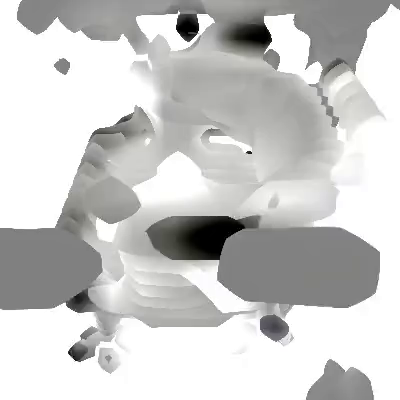} &
\includegraphics[trim= 0.0cm 0cm 0cm 0cm,clip, width=0.31\linewidth]{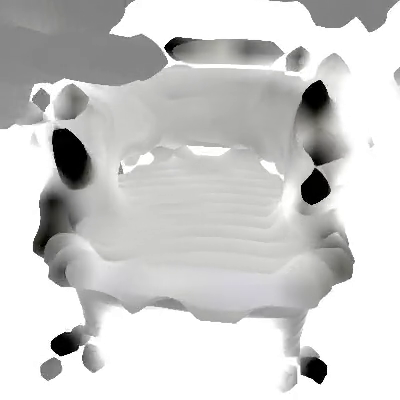}   &
\includegraphics[trim= 0.0cm 0cm 0cm 0cm,clip, width=0.31\linewidth]{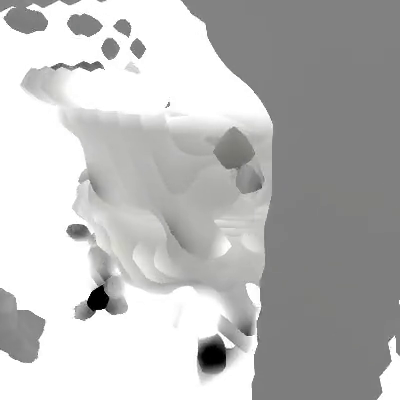}  \\
\bottomrule
\end{tabular}
}%
    \caption{\textbf{The Irregularity of SRFs}. The optimized SRFs used in our training are irregular 3D data structures. they include non-empty voxels that contain low-density radiance information that does not affect rendering. \textit{(top)}: renderings of a whole SRF, \textit{(bottom)}: renderings of the same SRF when densifying non-empty voxels. 
    }
    \label{figsup:messy}
\end{figure}

The optimized whole SRFs used in our training are irregular 3D data structures. They hold many non-empty voxels that contain low-density radiance information that does not affect rendering. As can be seen in \figLabel{\ref{figsup:messy}}, the non-empty and low-density components do not affect rendering but contain radiance information ( \eg black albedo) that affects the 3D learning pipeline. This motivates the use of the specialized losses proposed in this work, in order to tackle these challenges associated with SRFs. 

\section{Additional Results} \label{secsup:results}
\begin{figure}[t]
    \centering
\tabcolsep=0.03cm
\resizebox{0.99\linewidth}{!}{
\begin{tabular}{cccccc}
\multicolumn{6}{c}{$\xrightarrow{\quad \quad\text{Going out-of-distribution of training views }\quad \quad}$} \\
     \includegraphics[trim= 0.0cm 0cm 0cm 0cm,clip, width=0.15\linewidth]{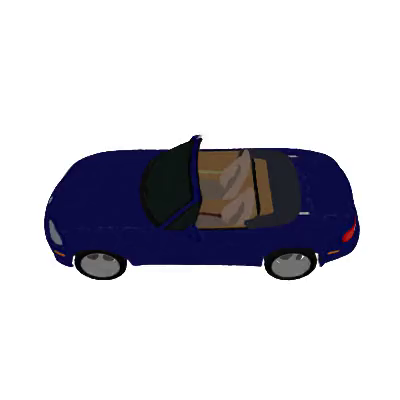}  &
   \includegraphics[trim= 0.0cm 0cm 0cm 0cm,clip, width=0.15\linewidth]{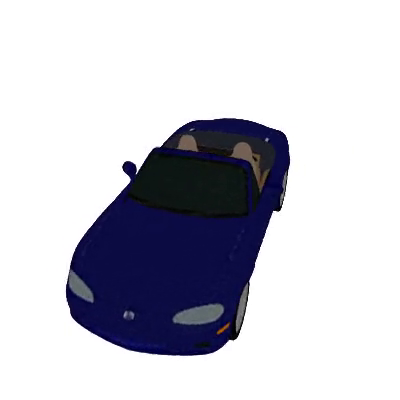}  &
   \includegraphics[trim= 0.0cm 0cm 0cm 0cm,clip, width=0.15\linewidth]{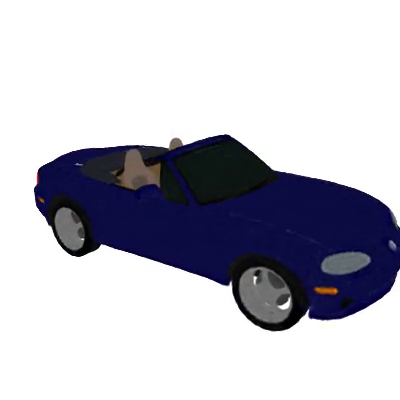}  &
\includegraphics[trim= 0.0cm 0cm 0cm 0cm,clip, width=0.15\linewidth]{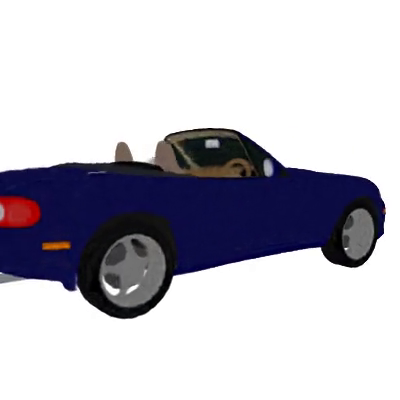} & 
\includegraphics[trim= 0.0cm 0cm 0cm 0cm,clip, width=0.15\linewidth]{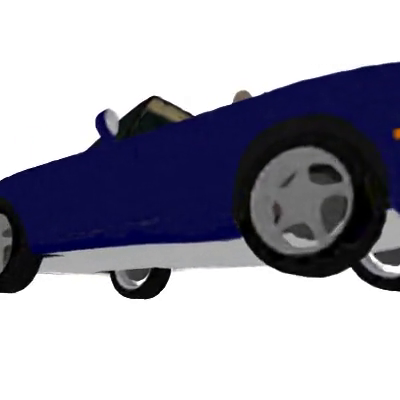} & 
\includegraphics[trim= 0.0cm 0cm 0cm 0cm,clip, width=0.15\linewidth]{images/baselines/s0/100/gt.png}    \\ \hline
     \includegraphics[trim= 0.0cm 0cm 0cm 0cm,clip, width=0.15\linewidth]{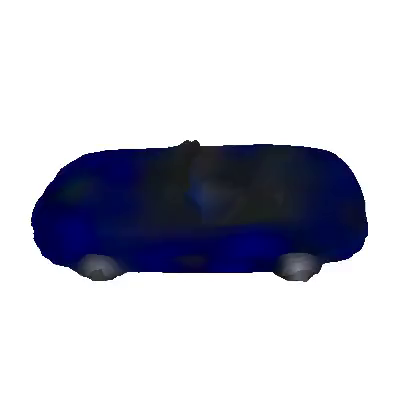}  &
   \includegraphics[trim= 0.0cm 0cm 0cm 0cm,clip, width=0.15\linewidth]{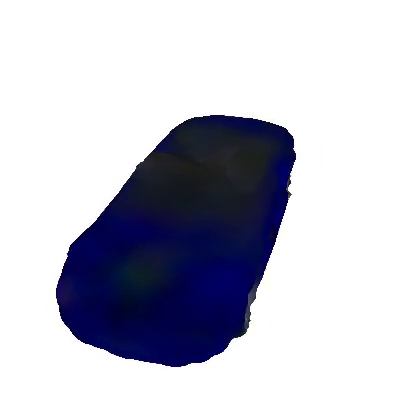}  &
   \includegraphics[trim= 0.0cm 0cm 0cm 0cm,clip, width=0.15\linewidth]{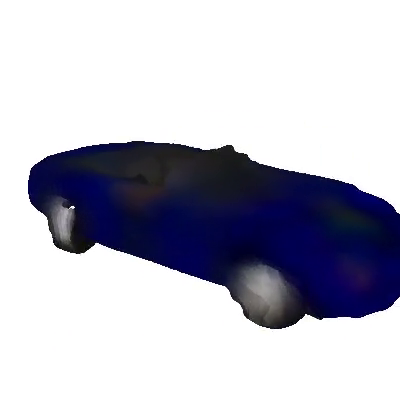}  &
\includegraphics[trim= 0.0cm 0cm 0cm 0cm,clip, width=0.15\linewidth]{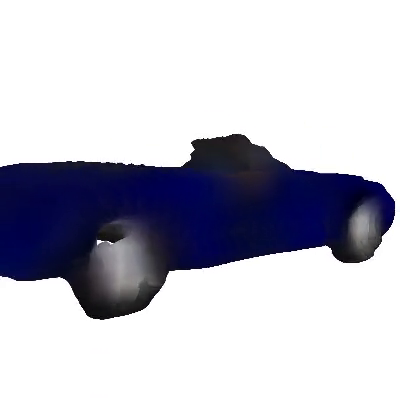} & 
\includegraphics[trim= 0.0cm 0cm 0cm 0cm,clip, width=0.15\linewidth]{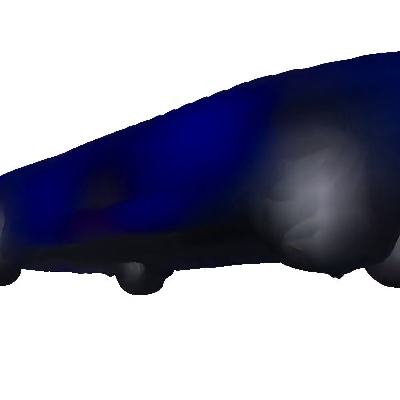} & 
\includegraphics[trim= 0.0cm 0cm 0cm 0cm,clip, width=0.15\linewidth]{images/baselines/s0/100/ours.png}  \\ \hline
     \includegraphics[trim= 0.0cm 0cm 0cm 0cm,clip, width=0.15\linewidth]{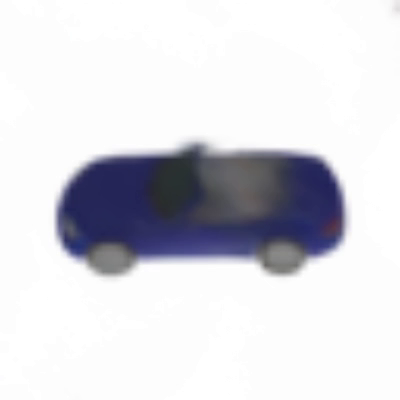}  &
   \includegraphics[trim= 0.0cm 0cm 0cm 0cm,clip, width=0.15\linewidth]{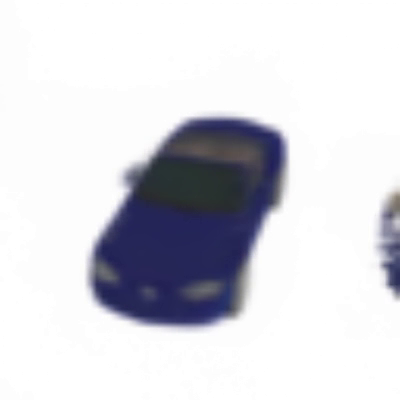}  &
   \includegraphics[trim= 0.0cm 0cm 0cm 0cm,clip, width=0.15\linewidth]{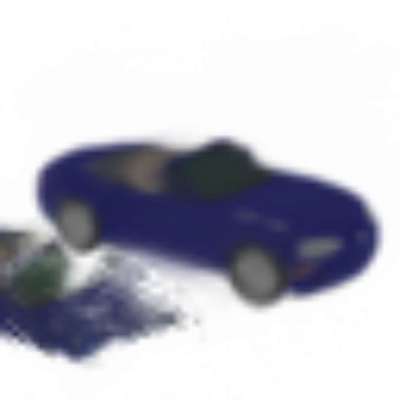}  &
\includegraphics[trim= 0.0cm 0cm 0cm 0cm,clip, width=0.15\linewidth]{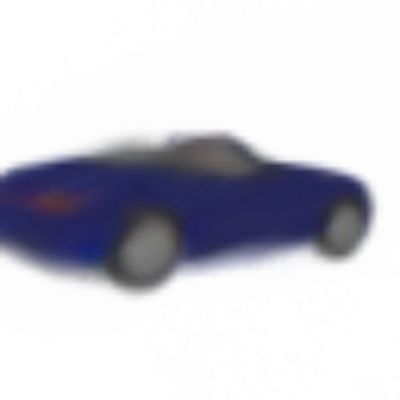} & 
\includegraphics[trim= 0.0cm 0cm 0cm 0cm,clip, width=0.15\linewidth]{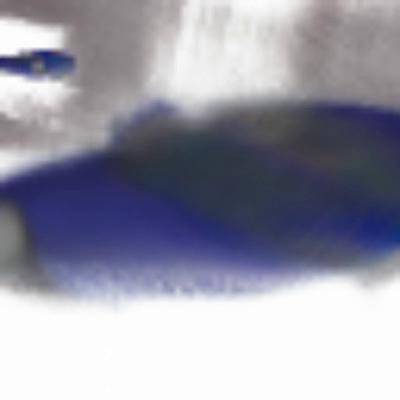} & 
\includegraphics[trim= 0.0cm 0cm 0cm 0cm,clip, width=0.15\linewidth]{images/baselines/s0/100/avision.png} \\ \hline
     \includegraphics[trim= 0.0cm 0cm 0cm 0cm,clip, width=0.15\linewidth]{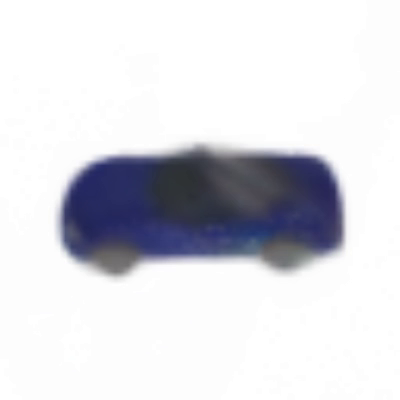}  &
   \includegraphics[trim= 0.0cm 0cm 0cm 0cm,clip, width=0.15\linewidth]{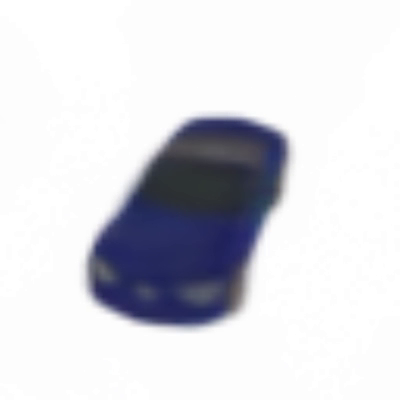}  &
   \includegraphics[trim= 0.0cm 0cm 0cm 0cm,clip, width=0.15\linewidth]{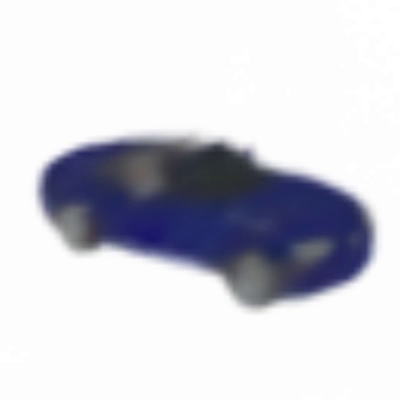}  &
\includegraphics[trim= 0.0cm 0cm 0cm 0cm,clip, width=0.15\linewidth]{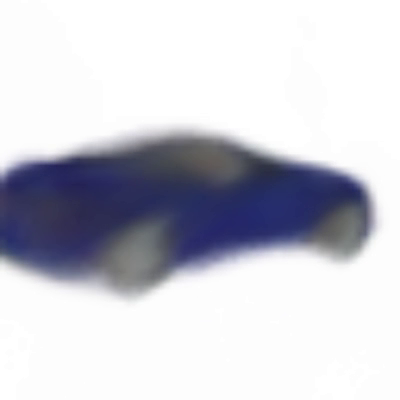} & 
\includegraphics[trim= 0.0cm 0cm 0cm 0cm,clip, width=0.15\linewidth]{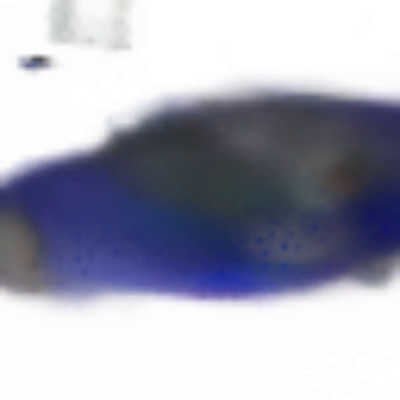} & 
\includegraphics[trim= 0.0cm 0cm 0cm 0cm,clip, width=0.15\linewidth]{images/baselines/s0/100/pixel.png} \\

 \bottomrule
\end{tabular}
}
    \caption{\textbf{Going Out-of-distribution of Training Views}. Renderings are shown of ground truth using whole SRFs (\textit{first row}), SurFNet [ours] (\textit{second row}), VisionNeRF \cite{visionnerf} (\textit{third row}), and PixelNeRF \cite{pix2nerf} (\textit{bottom row}). As rendered views move outside of the training distribution (going right), SuRFNet generally produces more 3D-consistent renderings than previous 2D-based methods.
    }
    \label{fig:arrow}
    \vspace{-4pt}
\end{figure}

\begin{figure*}
    \centering
        \tabcolsep=0.03cm
\resizebox{0.9\linewidth}{!}{
\begin{tabular}{c|c|c}
 \textbf{Train} & \textbf{Test}  & \textbf{Hard Test (OOD)} \\ \hline

          \includegraphics[trim= 5.0cm 3.2cm 3.5cm 1.3cm,clip, width=0.33\linewidth]{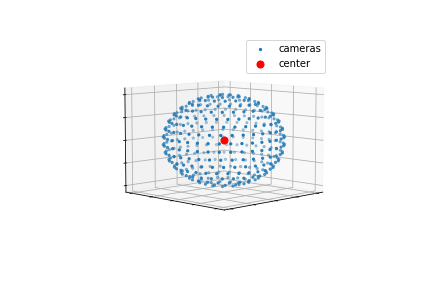} &
     \includegraphics[trim= 5.0cm 3.2cm 3.5cm 1.3cm,clip, width=0.33\linewidth]{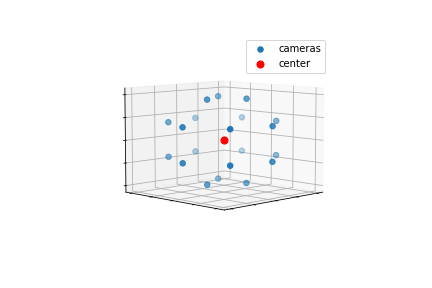}&
     \includegraphics[trim= 5.0cm 3.2cm 3.5cm 1.3cm,clip, width=0.33\linewidth]{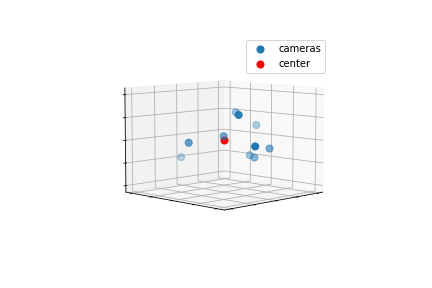} \\

\bottomrule
\end{tabular}
} %
    \caption{\textbf{Cameras Setups for Different SPARF Splits}. Here, we show different visualizations of the camera setups of the three splits of SPARF. (\textit{Train}): 400 determinsitic spherical views, (\textit{Test}): 20 random spherical views, (\textit{hard OOD Test}): 10 random views.
    }
    \label{fig:cameras}
\end{figure*}

\begin{figure*}
    \centering
    \tabcolsep=0.03cm
\resizebox{0.99\linewidth}{!}{
\begin{tabular}{ccccccc}
    \multicolumn{7}{c}{\textbf{Train/Test Track}} \\
     \includegraphics[trim= 0.0cm 0cm 0cm 0cm,clip, width=0.142\linewidth]{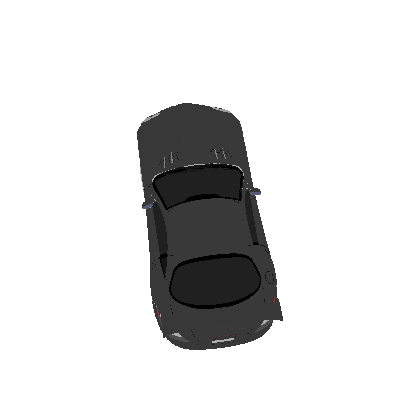}  &
     \includegraphics[trim= 0.0cm 0cm 0cm 0cm,clip, width=0.142\linewidth]{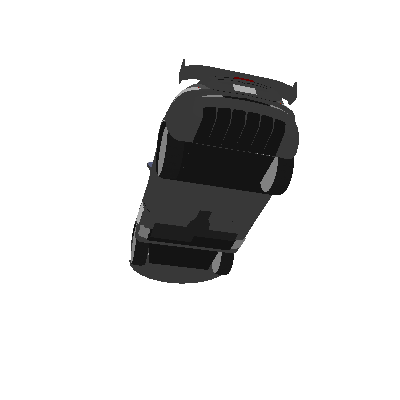}  &
     \includegraphics[trim= 0.0cm 0cm 0cm 0cm,clip, width=0.142\linewidth]{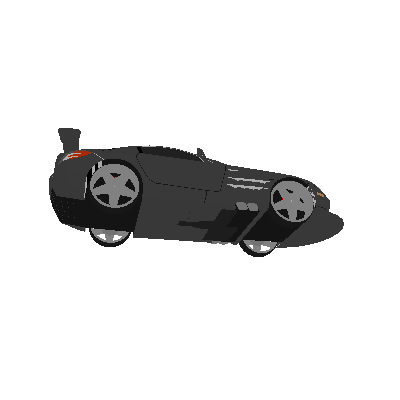}  &
     \includegraphics[trim= 0.0cm 0cm 0cm 0cm,clip, width=0.142\linewidth]{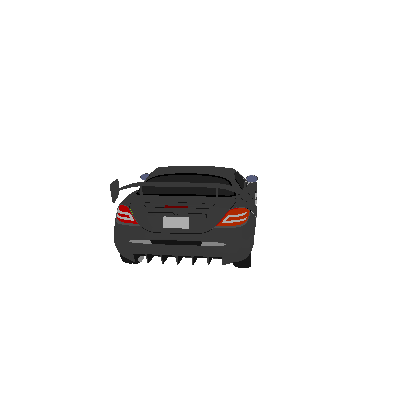}  &
     \includegraphics[trim= 0.0cm 0cm 0cm 0cm,clip, width=0.142\linewidth]{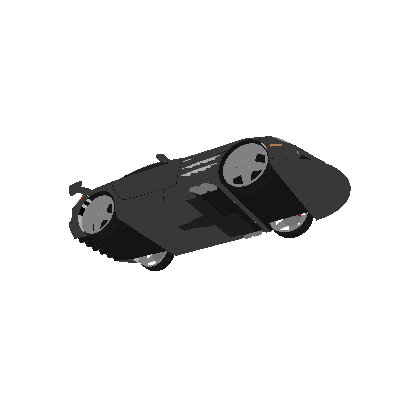}  &
     \includegraphics[trim= 0.0cm 0cm 0cm 0cm,clip, width=0.142\linewidth]{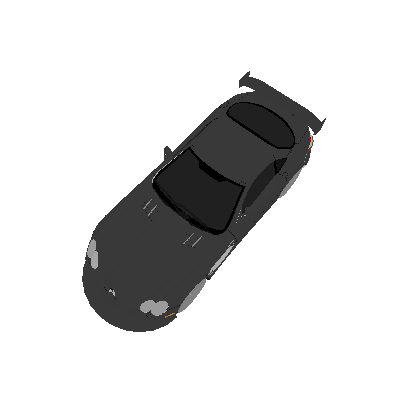}  &
     \includegraphics[trim= 0.0cm 0cm 0cm 0cm,clip, width=0.142\linewidth]{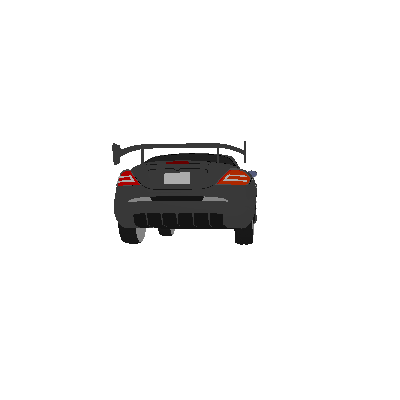}  \\ 
    \multicolumn{7}{c}{\textbf{OOD Hard Test Track}} \\
     \includegraphics[trim= 0.0cm 0cm 0cm 0cm,clip, width=0.142\linewidth]{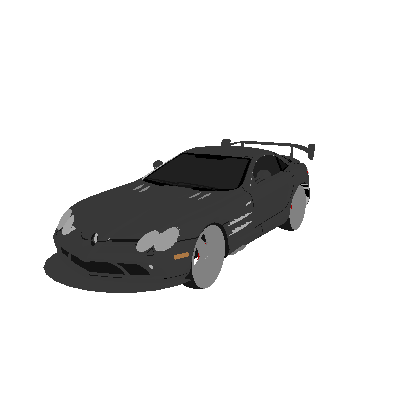}  &
     \includegraphics[trim= 0.0cm 0cm 0cm 0cm,clip, width=0.142\linewidth]{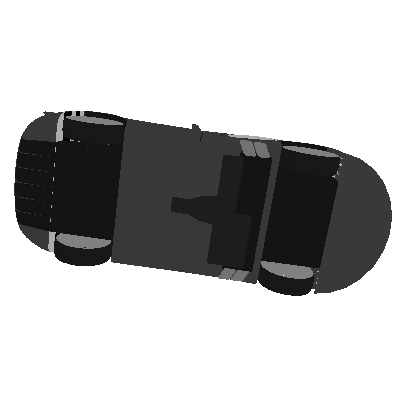}  &
     \includegraphics[trim= 0.0cm 0cm 0cm 0cm,clip, width=0.142\linewidth]{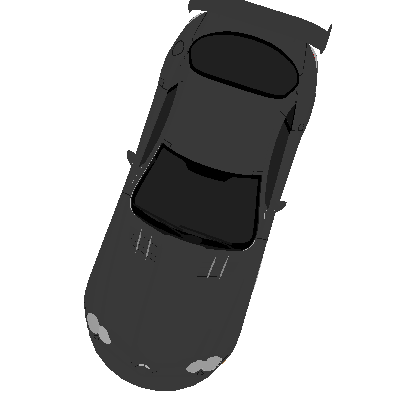}  &
     \includegraphics[trim= 0.0cm 0cm 0cm 0cm,clip, width=0.142\linewidth]{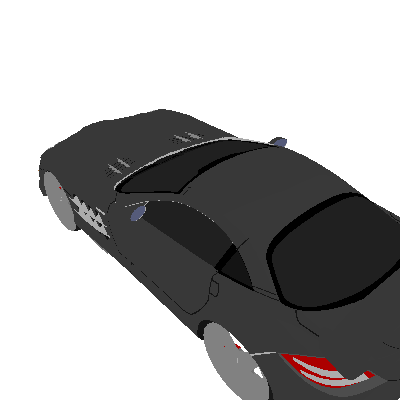}  &
     \includegraphics[trim= 0.0cm 0cm 0cm 0cm,clip, width=0.142\linewidth]{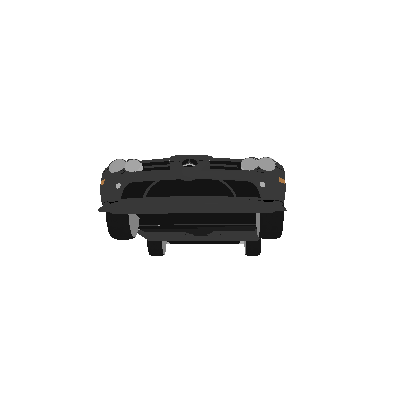}  &
     \includegraphics[trim= 0.0cm 0cm 0cm 0cm,clip, width=0.142\linewidth]{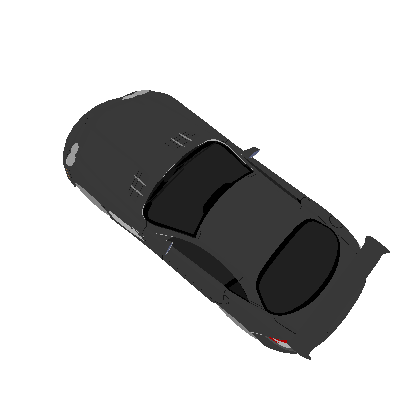}  &
     \includegraphics[trim= 0.0cm 0cm 0cm 0cm,clip, width=0.142\linewidth]{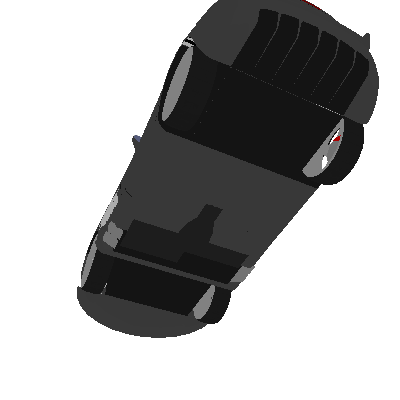}  \\ \midrule

    \multicolumn{7}{c}{\textbf{Train/Test Track}} \\
     \includegraphics[trim= 0.0cm 0cm 0cm 0cm,clip, width=0.142\linewidth]{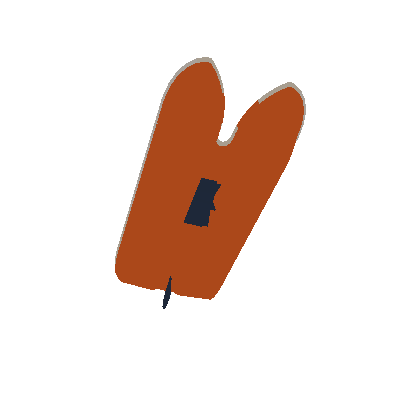}  &
     \includegraphics[trim= 0.0cm 0cm 0cm 0cm,clip, width=0.142\linewidth]{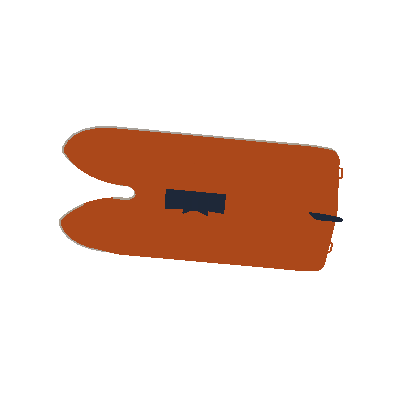}  &
     \includegraphics[trim= 0.0cm 0cm 0cm 0cm,clip, width=0.142\linewidth]{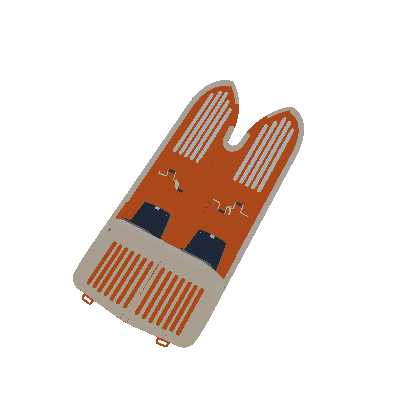}  &
     \includegraphics[trim= 0.0cm 0cm 0cm 0cm,clip, width=0.142\linewidth]{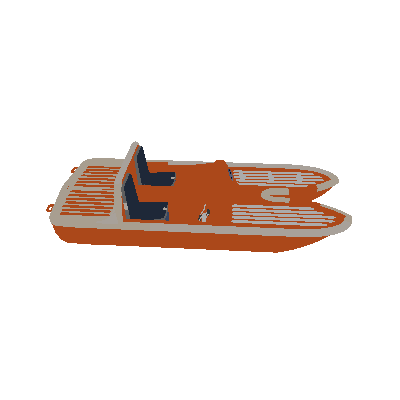}  &
     \includegraphics[trim= 0.0cm 0cm 0cm 0cm,clip, width=0.142\linewidth]{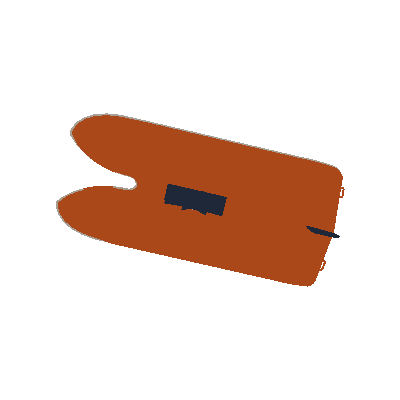}  &
     \includegraphics[trim= 0.0cm 0cm 0cm 0cm,clip, width=0.142\linewidth]{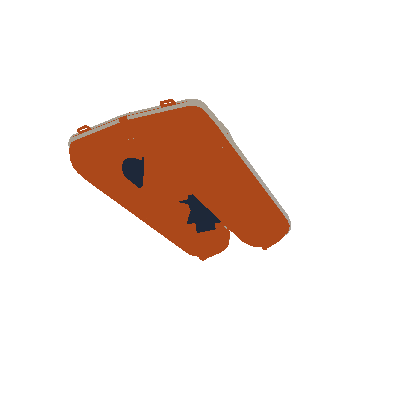}  &
     \includegraphics[trim= 0.0cm 0cm 0cm 0cm,clip, width=0.142\linewidth]{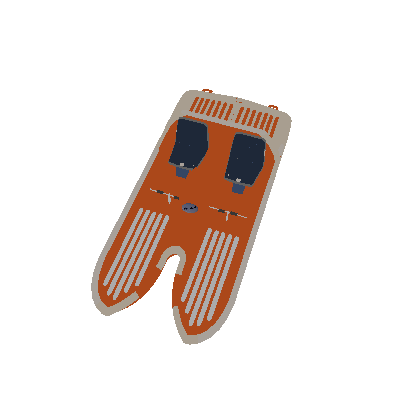}  \\ 
    \multicolumn{7}{c}{\textbf{OOD Hard Test Track}} \\
     \includegraphics[trim= 0.0cm 0cm 0cm 0cm,clip, width=0.142\linewidth]{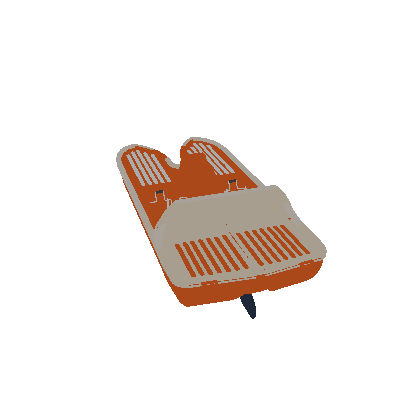}  &
     \includegraphics[trim= 0.0cm 0cm 0cm 0cm,clip, width=0.142\linewidth]{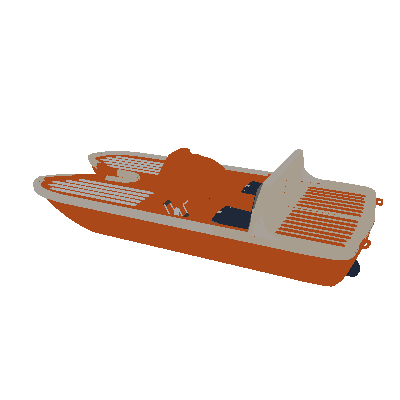}  &
     \includegraphics[trim= 0.0cm 0cm 0cm 0cm,clip, width=0.142\linewidth]{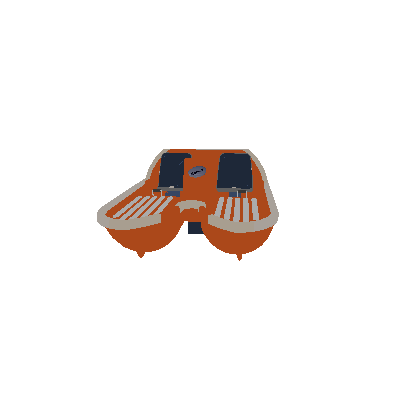}  &
     \includegraphics[trim= 0.0cm 0cm 0cm 0cm,clip, width=0.142\linewidth]{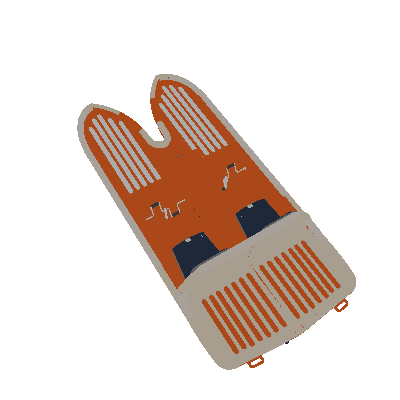}  &
     \includegraphics[trim= 0.0cm 0cm 0cm 0cm,clip, width=0.142\linewidth]{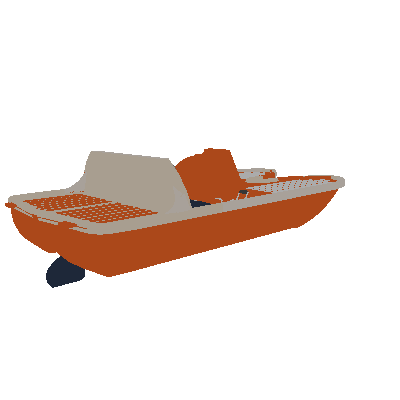}  &
     \includegraphics[trim= 0.0cm 0cm 0cm 0cm,clip, width=0.142\linewidth]{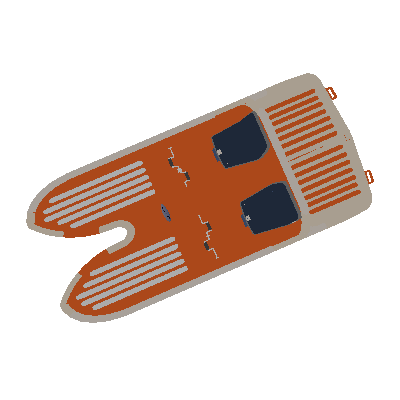}  &
     \includegraphics[trim= 0.0cm 0cm 0cm 0cm,clip, width=0.142\linewidth]{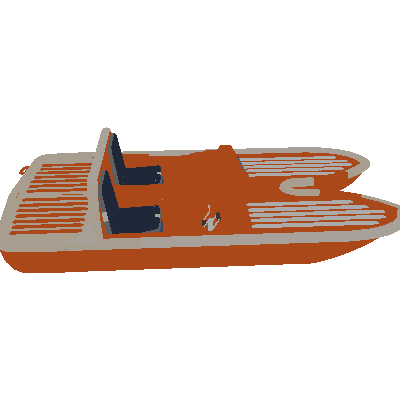}  \\

\bottomrule
\end{tabular}
} %
    \caption{\textbf{SPARF Splits}. SPARF has three main splits for every 3D shape: training views (400 views), test views (20 views), and OOD ``hard" views (10 views) as can be shown in the examples above.    
    }
    \label{fig:ood}
\end{figure*}

\begin{figure*}
    \centering
\tabcolsep=0.03cm
\resizebox{0.9\linewidth}{!}{
\begin{tabular}{ccc|ccc}
      \multicolumn{3}{c}{Ground Truth Whole SRFs } &       \multicolumn{3}{c}{Generated SRFs ($512^3$ resolution) } \\
     \includegraphics[trim= 0.0cm 0cm 0cm 0cm,clip, width=0.16\linewidth]{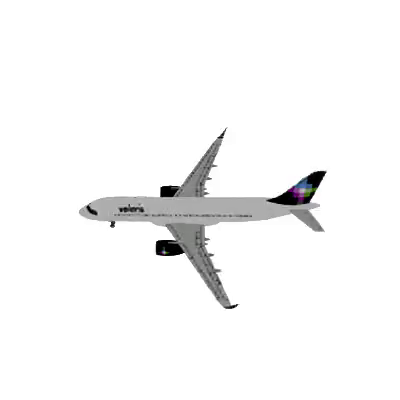}  &
     \includegraphics[trim= 0.0cm 0cm 0cm 0cm,clip, width=0.16\linewidth]{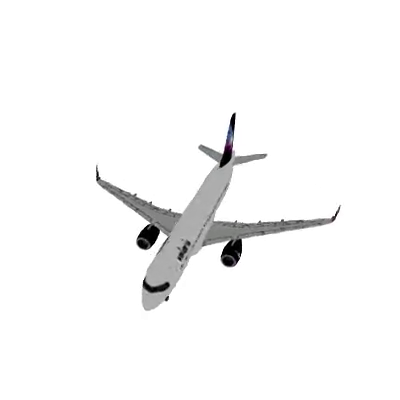}  &
     \includegraphics[trim= 0.0cm 0cm 0cm 0cm,clip, width=0.16\linewidth]{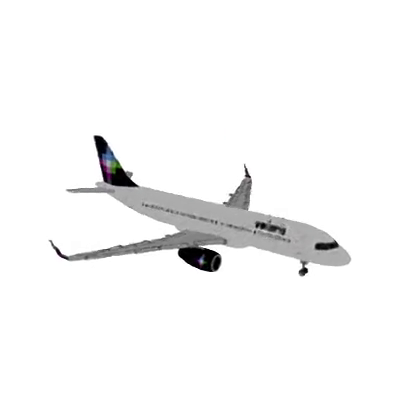}  &
     \includegraphics[trim= 0.0cm 0cm 0cm 0cm,clip, width=0.16\linewidth]{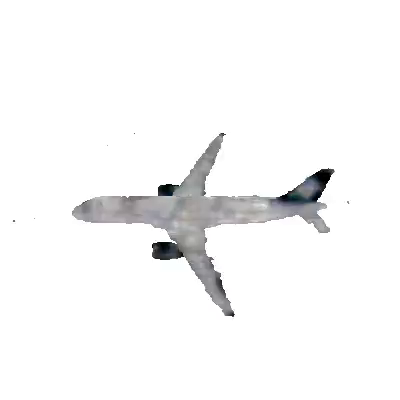}  &
     \includegraphics[trim= 0.0cm 0cm 0cm 0cm,clip, width=0.16\linewidth]{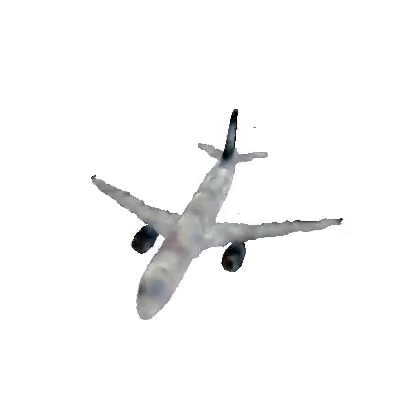}  &
     \includegraphics[trim= 0.0cm 0cm 0cm 0cm,clip, width=0.16\linewidth]{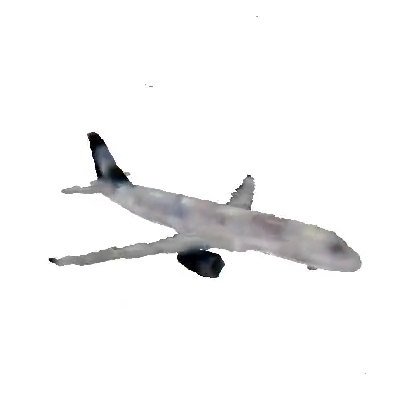}  \\ \hline
     \includegraphics[trim= 0.0cm 0cm 0cm 0cm,clip, width=0.16\linewidth]{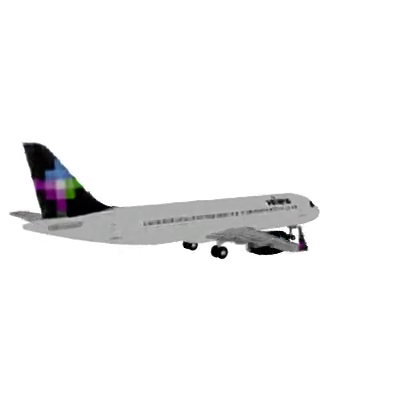}  &
     \includegraphics[trim= 0.0cm 0cm 0cm 0cm,clip, width=0.16\linewidth]{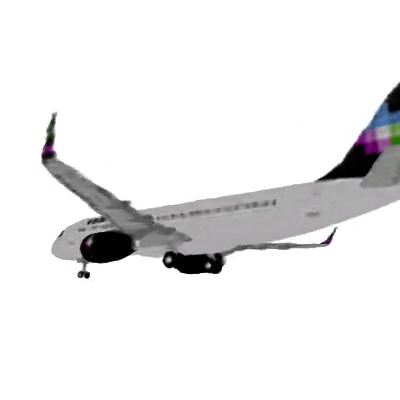}  &
     \includegraphics[trim= 0.0cm 0cm 0cm 0cm,clip, width=0.16\linewidth]{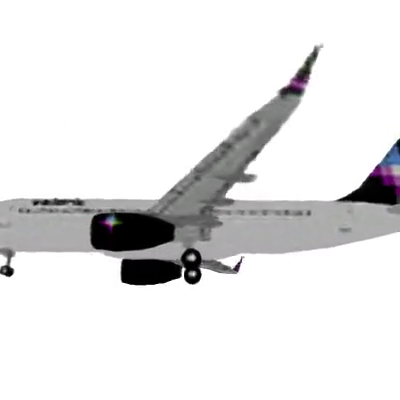}  &
     \includegraphics[trim= 0.0cm 0cm 0cm 0cm,clip, width=0.16\linewidth]{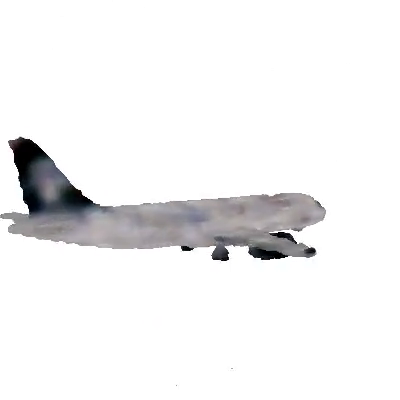}  &
     \includegraphics[trim= 0.0cm 0cm 0cm 0cm,clip, width=0.16\linewidth]{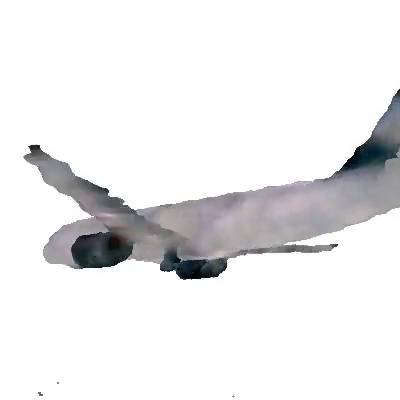}  &
     \includegraphics[trim= 0.0cm 0cm 0cm 0cm,clip, width=0.16\linewidth]{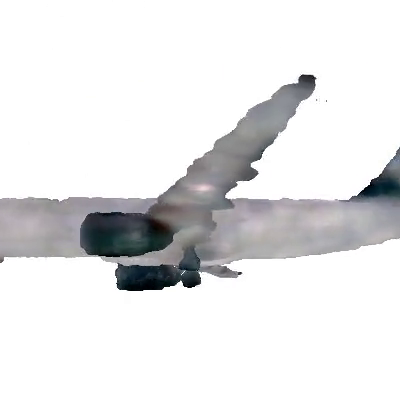}  \\ \hline
     
     \includegraphics[trim= 0.0cm 0cm 0cm 0cm,clip, width=0.16\linewidth]{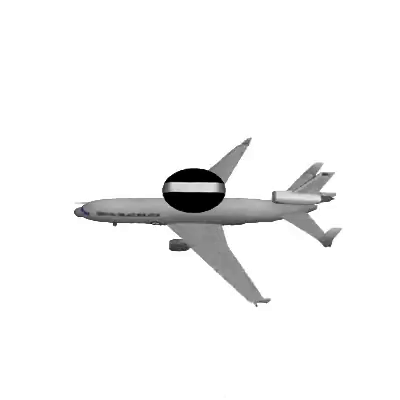}  &
     \includegraphics[trim= 0.0cm 0cm 0cm 0cm,clip, width=0.16\linewidth]{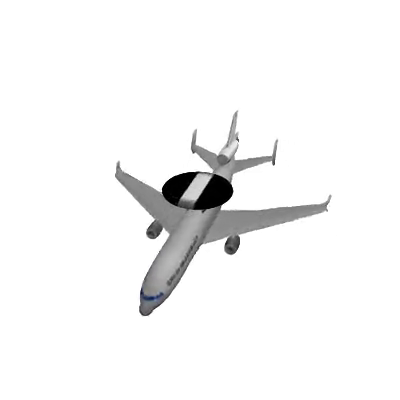}  &
     \includegraphics[trim= 0.0cm 0cm 0cm 0cm,clip, width=0.16\linewidth]{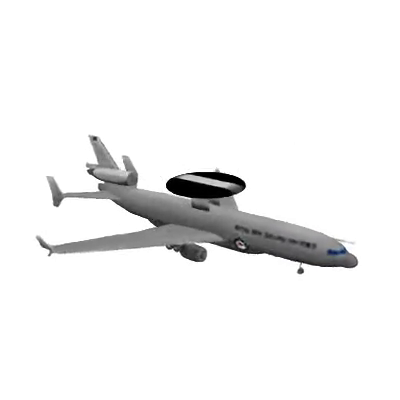}  &
     \includegraphics[trim= 0.0cm 0cm 0cm 0cm,clip, width=0.16\linewidth]{images/quality/s1/0/out.png}  &
     \includegraphics[trim= 0.0cm 0cm 0cm 0cm,clip, width=0.16\linewidth]{images/quality/s1/20/out.png}  &
     \includegraphics[trim= 0.0cm 0cm 0cm 0cm,clip, width=0.16\linewidth]{images/quality/s1/40/out.png}  \\ \hline
     
          \includegraphics[trim= 0.0cm 0cm 0cm 0cm,clip, width=0.16\linewidth]{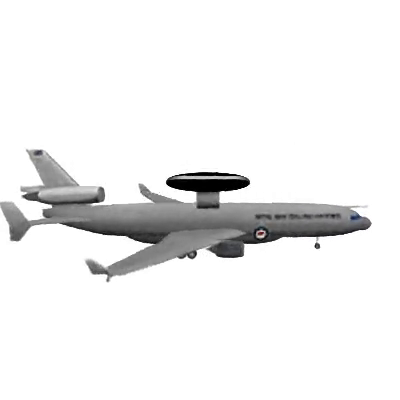}  &
     \includegraphics[trim= 0.0cm 0cm 0cm 0cm,clip, width=0.16\linewidth]{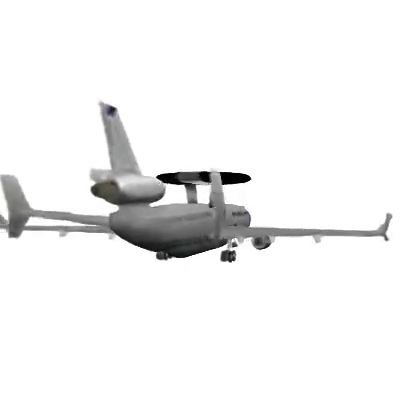}  &
     \includegraphics[trim= 0.0cm 0cm 0cm 0cm,clip, width=0.16\linewidth]{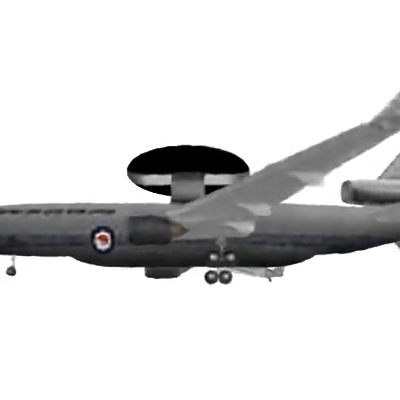}  &
     \includegraphics[trim= 0.0cm 0cm 0cm 0cm,clip, width=0.16\linewidth]{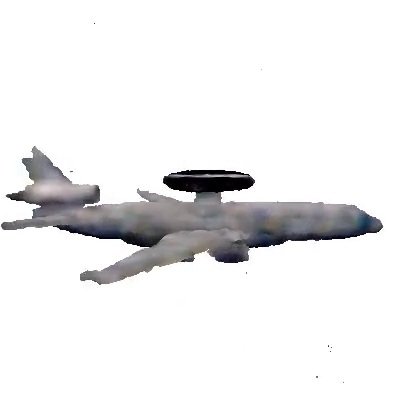}  &
     \includegraphics[trim= 0.0cm 0cm 0cm 0cm,clip, width=0.16\linewidth]{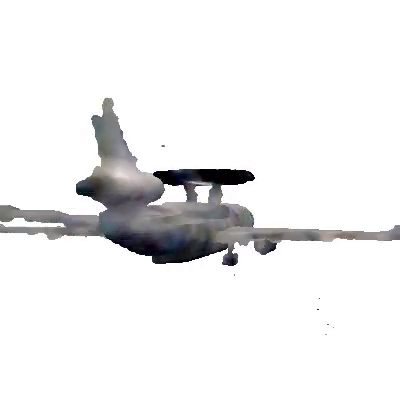}  &
     \includegraphics[trim= 0.0cm 0cm 0cm 0cm,clip, width=0.16\linewidth]{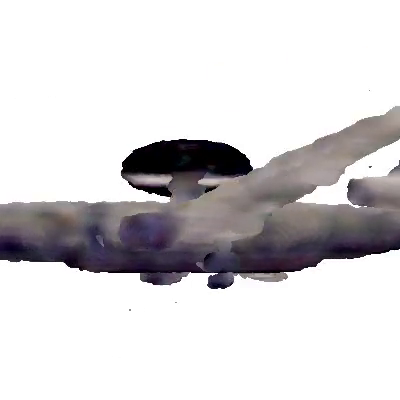}  \\ \hline
     
     \includegraphics[trim= 0.0cm 0cm 0cm 0cm,clip, width=0.16\linewidth]{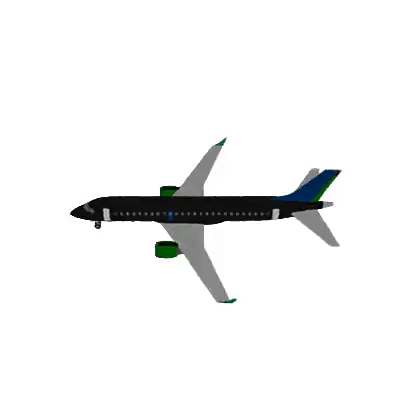}  &
     \includegraphics[trim= 0.0cm 0cm 0cm 0cm,clip, width=0.16\linewidth]{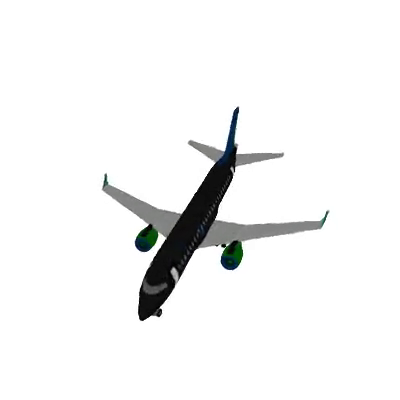}  &
     \includegraphics[trim= 0.0cm 0cm 0cm 0cm,clip, width=0.16\linewidth]{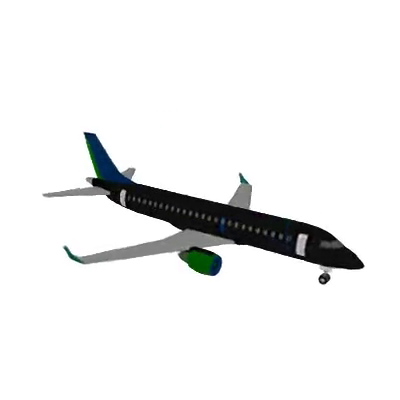}  &
     \includegraphics[trim= 0.0cm 0cm 0cm 0cm,clip, width=0.16\linewidth]{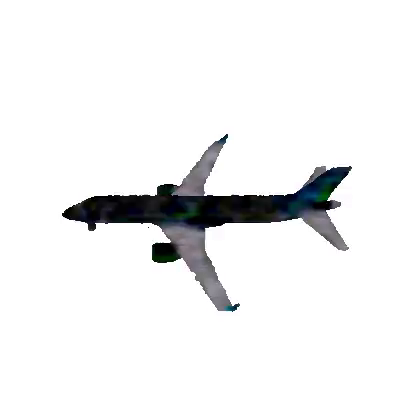}  &
     \includegraphics[trim= 0.0cm 0cm 0cm 0cm,clip, width=0.16\linewidth]{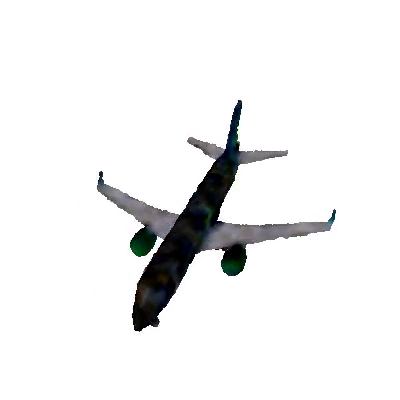}  &
     \includegraphics[trim= 0.0cm 0cm 0cm 0cm,clip, width=0.16\linewidth]{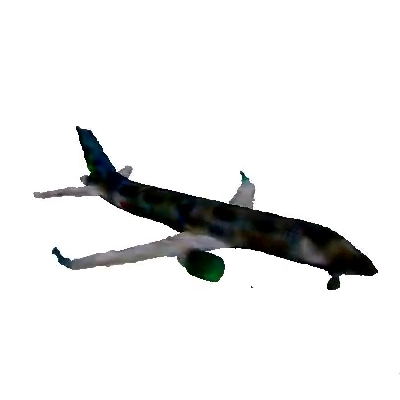}  \\ \hline
     
          \includegraphics[trim= 0.0cm 0cm 0cm 0cm,clip, width=0.16\linewidth]{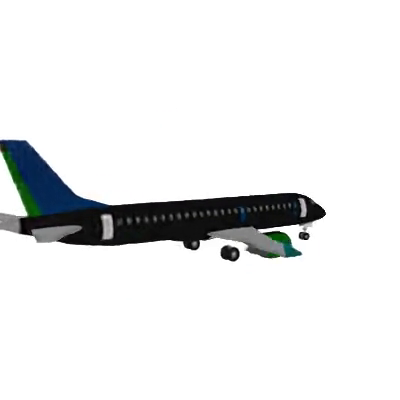}  &
     \includegraphics[trim= 0.0cm 0cm 0cm 0cm,clip, width=0.16\linewidth]{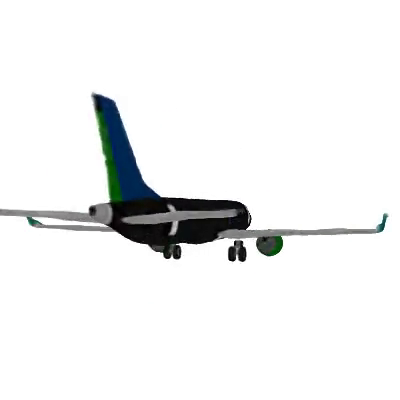}  &
     \includegraphics[trim= 0.0cm 0cm 0cm 0cm,clip, width=0.16\linewidth]{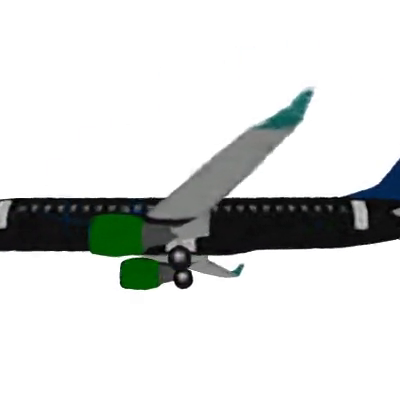}  &
     \includegraphics[trim= 0.0cm 0cm 0cm 0cm,clip, width=0.16\linewidth]{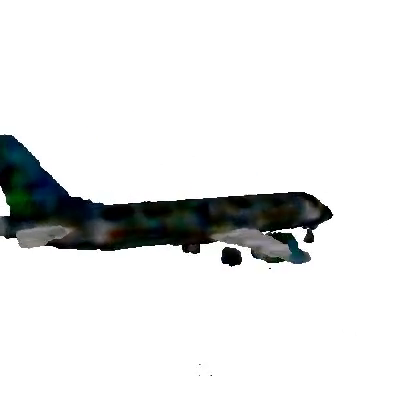}  &
     \includegraphics[trim= 0.0cm 0cm 0cm 0cm,clip, width=0.16\linewidth]{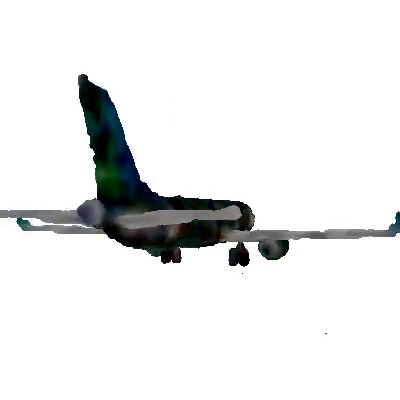}  &
     \includegraphics[trim= 0.0cm 0cm 0cm 0cm,clip, width=0.16\linewidth]{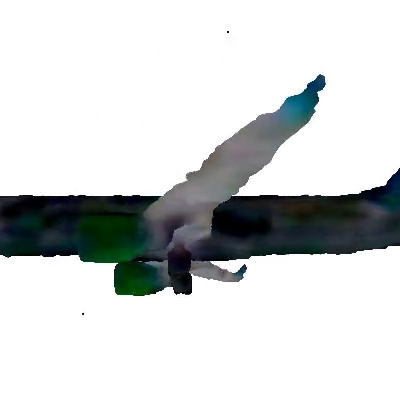}  \\
 \bottomrule
\end{tabular}
} %
    \caption{\textbf{SuRFNet: Generating High-Resolution Radiance Fields}. We show some volume-rendered sequences based on our SuRFNet voxel radiance field outputs (512 resolution), given only 3 images of each shape. This demonstrates the capability of SuRFNet to generate high-resolution sparse voxel SRFs. Note that, here, SURFNet is overfitting on a small dataset in these examples and is not meant for shape generalization. 
    }
    \label{figsup:interpolation}
\end{figure*}

\begin{figure*} [t] 
\centering
\tabcolsep=0.03cm
\resizebox{\linewidth}{!}{
\begin{tabular}{c|ccccc}
Input View & PixelNeRF \cite{pixelnerf} & VisionNeRF \cite{visionnerf} & \textbf{SuRFNet (ours)} &  Ground Truth   \\
\includegraphics[trim= 0.0cm 0cm 0cm 0cm,clip, width=0.2\linewidth]{images/baselines/s0/ourinput.png} &
    \includegraphics[trim= 0.0cm 0cm 0cm 0cm,clip, width=0.2\linewidth]{images/baselines/s0/20/pixel.png}  &
    \includegraphics[trim= 0.0cm 0cm 0cm 0cm,clip, width=0.2\linewidth]{images/baselines/s0/20/avision.png}  &
    \includegraphics[trim= 0.0cm 0cm 0cm 0cm,clip, width=0.2\linewidth]{images/baselines/s0/20/ours.png}  &
    \includegraphics[trim= 0.0cm 0cm 0cm 0cm,clip, width=0.2\linewidth]{images/baselines/s0/20/gt.png}  \\
    \includegraphics[trim= 0.0cm 0cm 0cm 0cm,clip, width=0.2\linewidth]{images/baselines/s0/ourinput.png} &
    \includegraphics[trim= 0.0cm 0cm 0cm 0cm,clip, width=0.2\linewidth]{images/baselines/s0/40/pixel.png}  &
    \includegraphics[trim= 0.0cm 0cm 0cm 0cm,clip, width=0.2\linewidth]{images/baselines/s0/40/avision.png}  &
    \includegraphics[trim= 0.0cm 0cm 0cm 0cm,clip, width=0.2\linewidth]{images/baselines/s0/40/ours.png}  &
    \includegraphics[trim= 0.0cm 0cm 0cm 0cm,clip, width=0.2\linewidth]{images/baselines/s0/40/gt.png}  \\
    \includegraphics[trim= 0.0cm 0cm 0cm 0cm,clip, width=0.2\linewidth]{images/baselines/s0/ourinput.png} &
    \includegraphics[trim= 0.0cm 0cm 0cm 0cm,clip, width=0.2\linewidth]{images/baselines/s0/60/pixel.png}  &
    \includegraphics[trim= 0.0cm 0cm 0cm 0cm,clip, width=0.2\linewidth]{images/baselines/s0/60/avision.png}  &
    \includegraphics[trim= 0.0cm 0cm 0cm 0cm,clip, width=0.2\linewidth]{images/baselines/s0/60/ours.png}  &
    \includegraphics[trim= 0.0cm 0cm 0cm 0cm,clip, width=0.2\linewidth]{images/baselines/s0/60/gt.png}  \\
    \includegraphics[trim= 0.0cm 0cm 0cm 0cm,clip, width=0.2\linewidth]{images/baselines/s0/ourinput.png} &
    \includegraphics[trim= 0.0cm 0cm 0cm 0cm,clip, width=0.2\linewidth]{images/baselines/s0/80/pixel.png}  &
    \includegraphics[trim= 0.0cm 0cm 0cm 0cm,clip, width=0.2\linewidth]{images/baselines/s0/80/avision.png}  &
    \includegraphics[trim= 0.0cm 0cm 0cm 0cm,clip, width=0.2\linewidth]{images/baselines/s0/80/ours.png}  &
    \includegraphics[trim= 0.0cm 0cm 0cm 0cm,clip, width=0.2\linewidth]{images/baselines/s0/80/gt.png}  \\
    \includegraphics[trim= 0.0cm 0cm 0cm 0cm,clip, width=0.2\linewidth]{images/baselines/s0/ourinput.png} &
    \includegraphics[trim= 0.0cm 0cm 0cm 0cm,clip, width=0.2\linewidth]{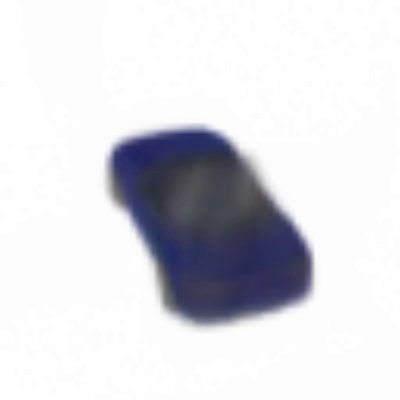}  &
    \includegraphics[trim= 0.0cm 0cm 0cm 0cm,clip, width=0.2\linewidth]{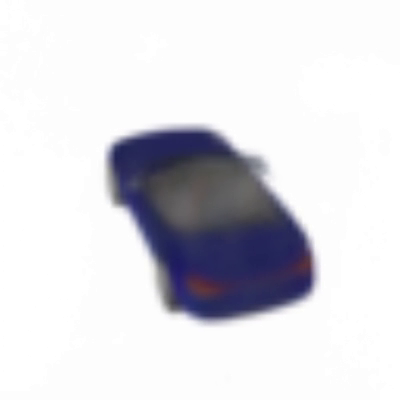}  &
    \includegraphics[trim= 0.0cm 0cm 0cm 0cm,clip, width=0.2\linewidth]{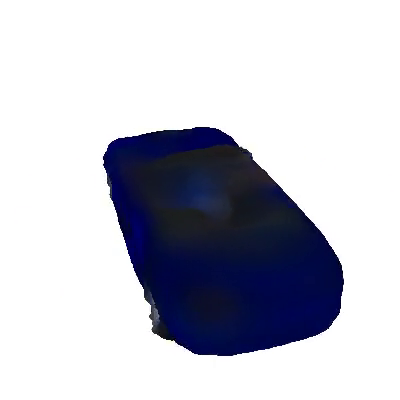}  &
    \includegraphics[trim= 0.0cm 0cm 0cm 0cm,clip, width=0.2\linewidth]{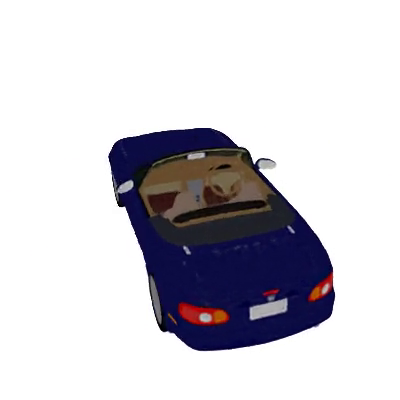}  \\   
 \bottomrule
\end{tabular}
} %
    \caption{\textbf{Qualititve Comparisons 1}. We show different render from our SuRFNet outputs generated from a single image compared to other methods (pixel-Nerf \cite{pixelnerf}, and VisionNerf \cite{visionnerf} ) and whole SRF ''GT" renderings. Note that the predicted views are outside the training views distribution (zoomed in randomly). This test highlights the weakness of the 2D-based baselines \cite{pixelnerf,visionnerf} outside the training track, while our 3D approach maintains multi-view consistency everywhere.
    }
    \label{fig:comparison1}
\end{figure*}

\begin{figure*} [t] 
\centering
\tabcolsep=0.03cm
\resizebox{0.97\linewidth}{!}{
\begin{tabular}{c|ccccc}
Input Views & PixelNeRF \cite{pixelnerf} & VisionNeRF \cite{visionnerf} & \textbf{SuRFNet (ours)} &  Ground Truth   \\
    \includegraphics[trim= 0.0cm 0cm 0cm 0cm,clip, width=0.2\linewidth]{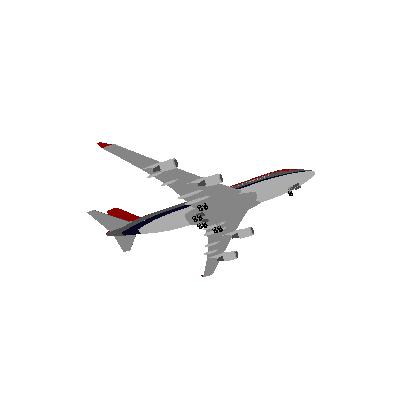} &
    \includegraphics[trim= 0.0cm 0cm 0cm 0cm,clip, width=0.2\linewidth]{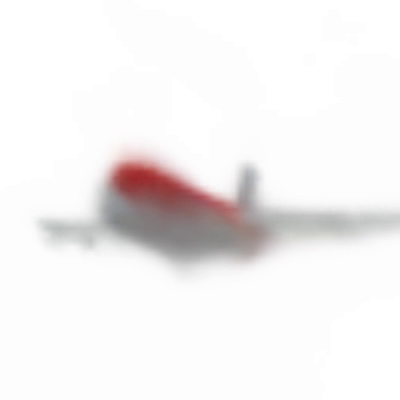}  &
    \includegraphics[trim= 0.0cm 0cm 0cm 0cm,clip, width=0.2\linewidth]{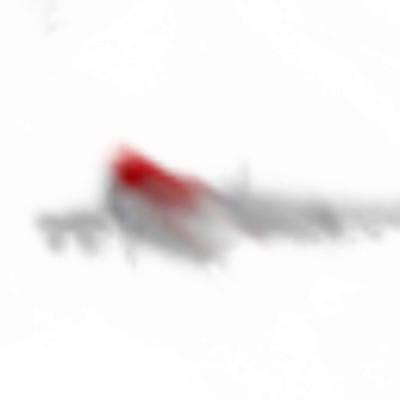}  &
    \includegraphics[trim= 0.0cm 0cm 0cm 0cm,clip, width=0.2\linewidth]{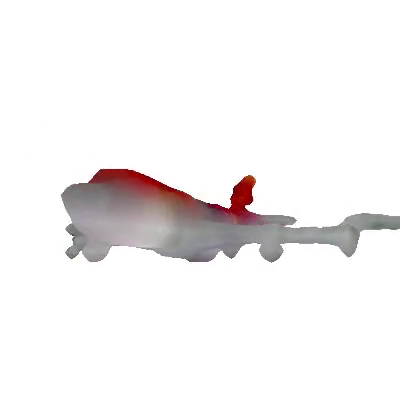}  &
    \includegraphics[trim= 0.0cm 0cm 0cm 0cm,clip, width=0.2\linewidth]{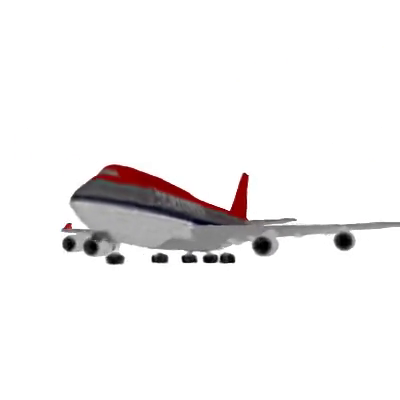}  \\
\includegraphics[trim= 0.0cm 0cm 0cm 0cm,clip, width=0.2\linewidth]{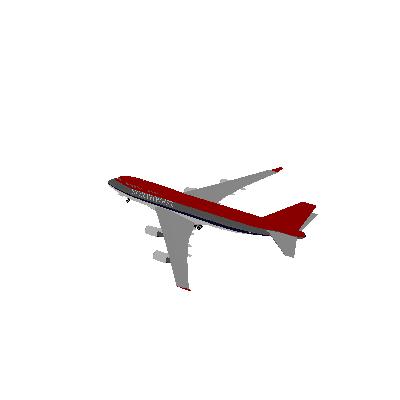} &
    \includegraphics[trim= 0.0cm 0cm 0cm 0cm,clip, width=0.2\linewidth]{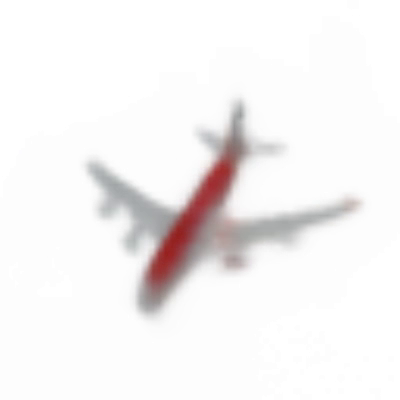}  &
    \includegraphics[trim= 0.0cm 0cm 0cm 0cm,clip, width=0.2\linewidth]{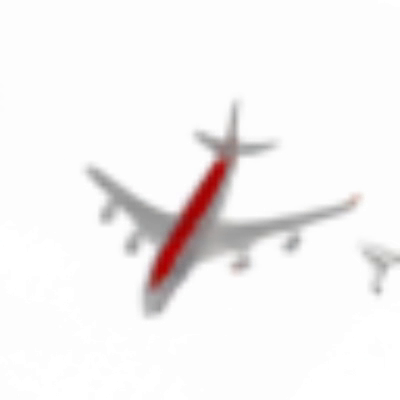}  &
    \includegraphics[trim= 0.0cm 0cm 0cm 0cm,clip, width=0.2\linewidth]{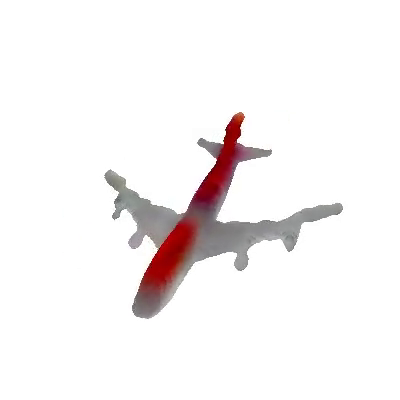}  &
    \includegraphics[trim= 0.0cm 0cm 0cm 0cm,clip, width=0.2\linewidth]{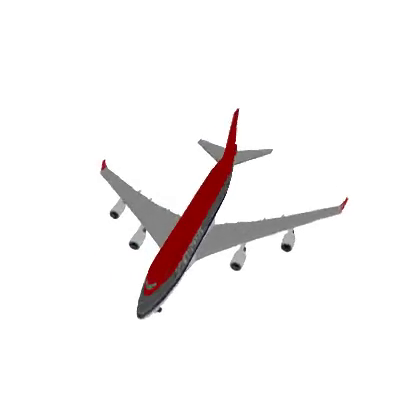}  \\
    \includegraphics[trim= 0.0cm 0cm 0cm 0cm,clip, width=0.2\linewidth]{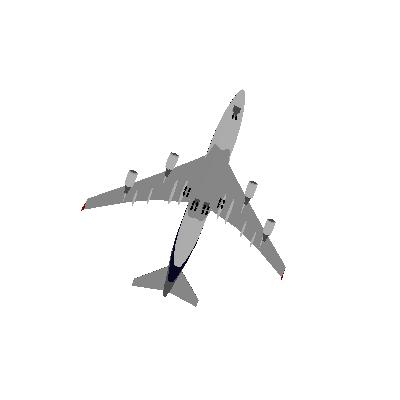} &
    \includegraphics[trim= 0.0cm 0cm 0cm 0cm,clip, width=0.2\linewidth]{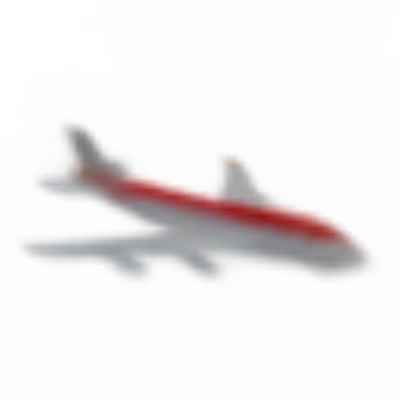}  &
    \includegraphics[trim= 0.0cm 0cm 0cm 0cm,clip, width=0.2\linewidth]{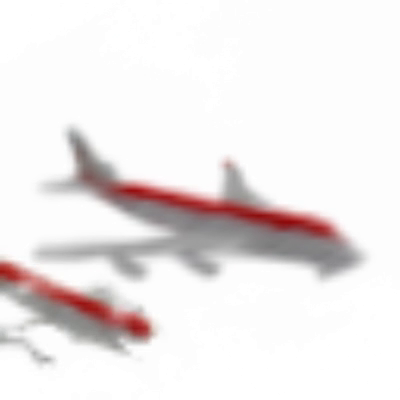}  &
    \includegraphics[trim= 0.0cm 0cm 0cm 0cm,clip, width=0.2\linewidth]{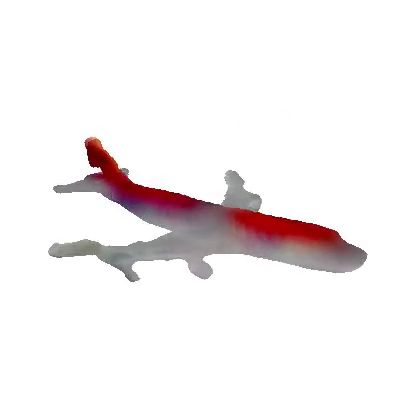}  &
    \includegraphics[trim= 0.0cm 0cm 0cm 0cm,clip, width=0.2\linewidth]{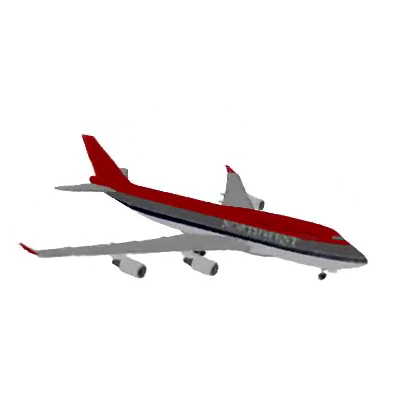}  \\
     &
    \includegraphics[trim= 0.0cm 0cm 0cm 0cm,clip, width=0.2\linewidth]{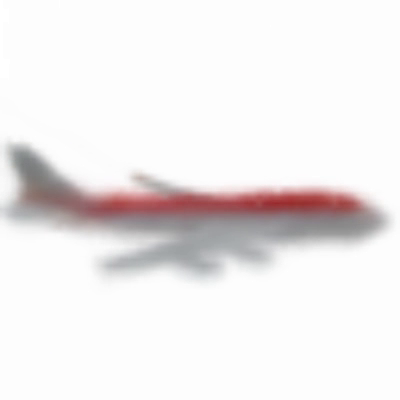}  &
    \includegraphics[trim= 0.0cm 0cm 0cm 0cm,clip, width=0.2\linewidth]{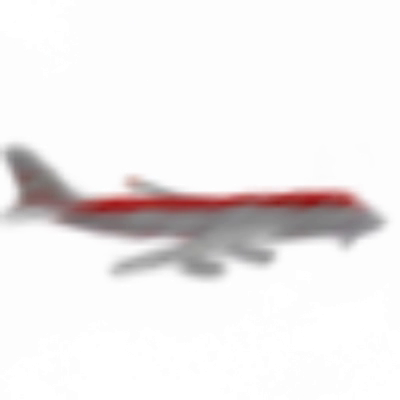}  &
    \includegraphics[trim= 0.0cm 0cm 0cm 0cm,clip, width=0.2\linewidth]{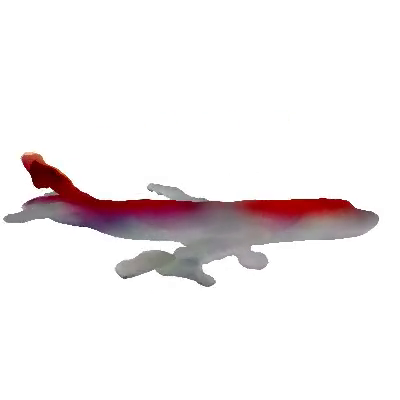}  &
    \includegraphics[trim= 0.0cm 0cm 0cm 0cm,clip, width=0.2\linewidth]{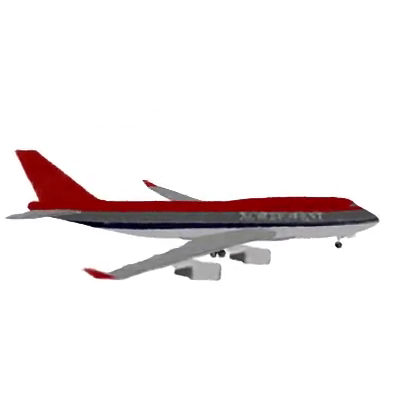}  \\
    &
    \includegraphics[trim= 0.0cm 0cm 0cm 0cm,clip, width=0.2\linewidth]{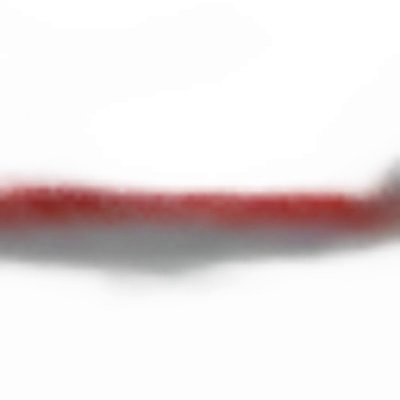}  &
    \includegraphics[trim= 0.0cm 0cm 0cm 0cm,clip, width=0.2\linewidth]{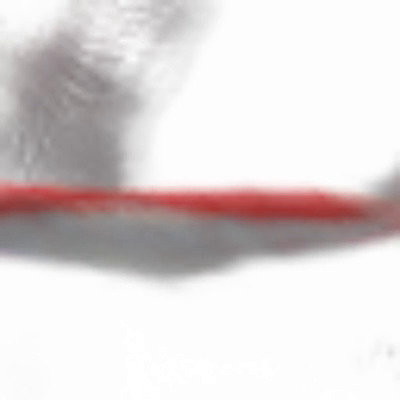}  &
    \includegraphics[trim= 0.0cm 0cm 0cm 0cm,clip, width=0.2\linewidth]{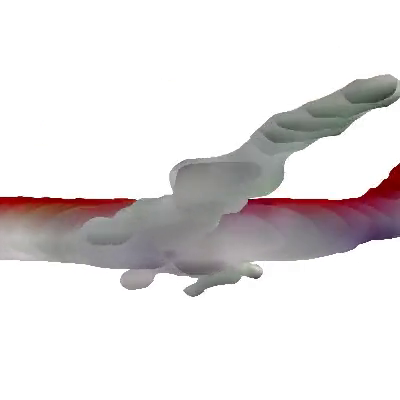}  &
    \includegraphics[trim= 0.0cm 0cm 0cm 0cm,clip, width=0.2\linewidth]{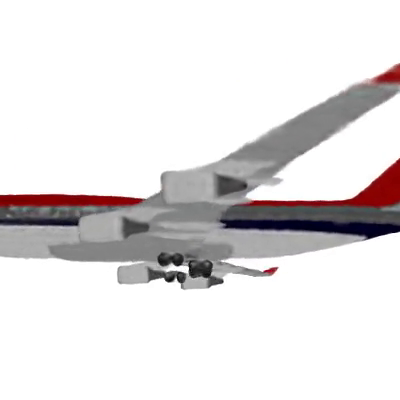}  \\
    \ &
    \includegraphics[trim= 0.0cm 0cm 0cm 0cm,clip, width=0.2\linewidth]{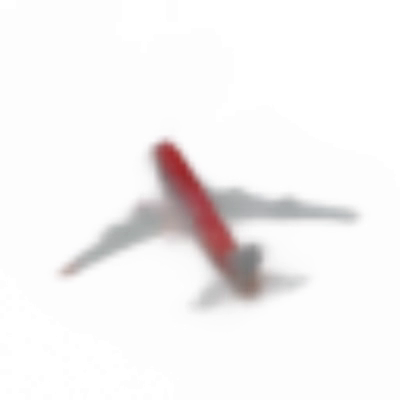}  &
    \includegraphics[trim= 0.0cm 0cm 0cm 0cm,clip, width=0.2\linewidth]{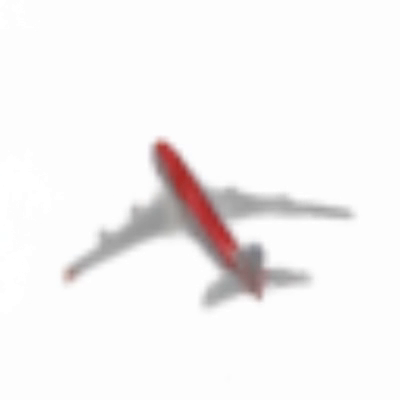}  &
    \includegraphics[trim= 0.0cm 0cm 0cm 0cm,clip, width=0.2\linewidth]{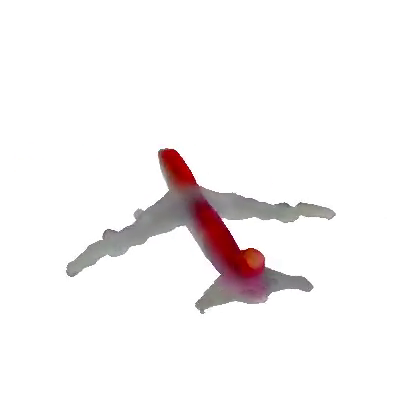}  &
    \includegraphics[trim= 0.0cm 0cm 0cm 0cm,clip, width=0.2\linewidth]{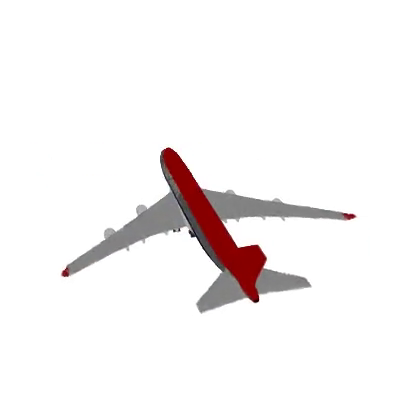}  \\   
 \bottomrule
\end{tabular}
} %
    \caption{\textbf{Qualititve Comparisons 2}. We show different render from our SuRFNet outputs generated from 3 input images compared to other methods (pixel-Nerf \cite{pixelnerf}, and VisionNerf \cite{visionnerf} ) and whole SRF ''GT" renderings. Note that the predicted views are outside the training views distribution (zoomed in randomly). This test highlights the weakness of the 2D-based baselines \cite{pixelnerf,visionnerf} outside the training track, while our 3D approach maintains multi-view consistency everywhere.
    }
    \label{fig:comparison2}
\end{figure*}

\begin{figure*}[t]
    \centering
\tabcolsep=0.03cm
\resizebox{0.99\linewidth}{!}{
\begin{tabular}{c|ccccc}
Real & \multicolumn{5}{c}{Generated} \\
         \includegraphics[trim= 0.0cm 16cm 0cm 10cm,clip, width=0.16\linewidth]{images/real/s0/co.jpg}  &
     \includegraphics[trim= 0.0cm 9cm 0cm 9cm,clip, width=0.16\linewidth]{images/real/s0/0/srf_rot.png}  &
          \includegraphics[trim= 0.0cm 9cm 0cm 9cm,clip, width=0.16\linewidth]{images/real/s0/20/srf_rot.png} & 
                    \includegraphics[trim= 0.0cm 9cm 0cm 9cm,clip, width=0.16\linewidth]{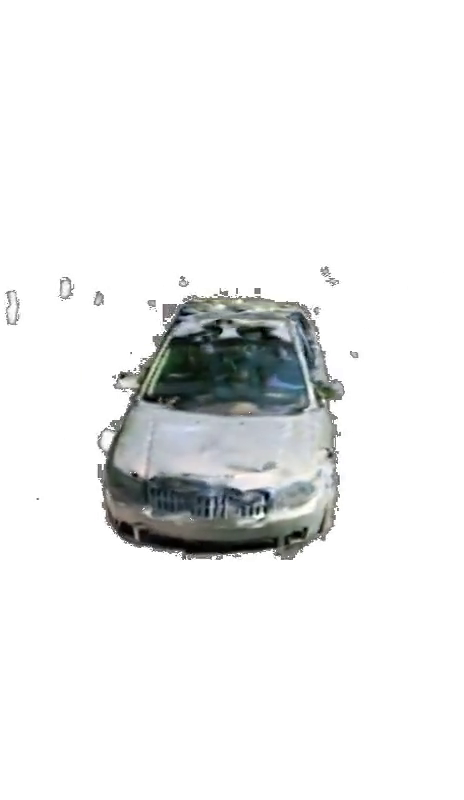} & 
          \includegraphics[trim= 0.0cm 9cm 0cm 9cm,clip, width=0.16\linewidth]{images/real/s0/120/srf_rot.png}  &
          \includegraphics[trim= 0.0cm 9cm 0cm 9cm,clip, width=0.16\linewidth]{images/real/s0/180/srf_rot.png}  \\
         \includegraphics[trim= 0.0cm 16cm 0cm 10cm,clip, width=0.16\linewidth]{images/real/s0/co.jpg}  &
     \includegraphics[trim= 5cm 6cm 5cm 5.5cm,clip, width=0.16\linewidth]{images/real/s0/0/pixel.png}  &
          \includegraphics[trim= 5cm 6cm 5cm 5.5cm,clip, width=0.16\linewidth]{images/real/s0/20/pixel.png} & 
                 \includegraphics[trim= 5cm 6cm 5cm 5.5cm,clip, width=0.16\linewidth]{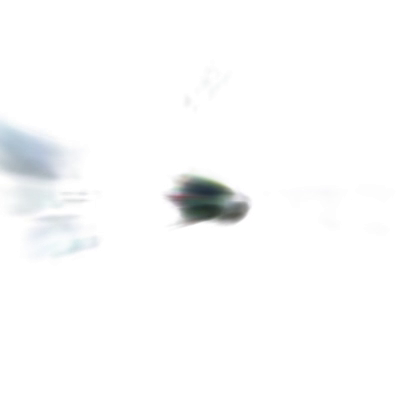} & 
          \includegraphics[trim= 5cm 6cm 5cm 5.5cm,clip, width=0.16\linewidth]{images/real/s0/120/pixel.png}  &
          \includegraphics[trim= 5cm 6cm 5cm 5.5cm,clip, width=0.16\linewidth]{images/real/s0/180/pixel.png}  \\ \midrule

     \includegraphics[trim= 4.0cm 8cm 4cm 1.0cm,clip, width=0.16\linewidth]{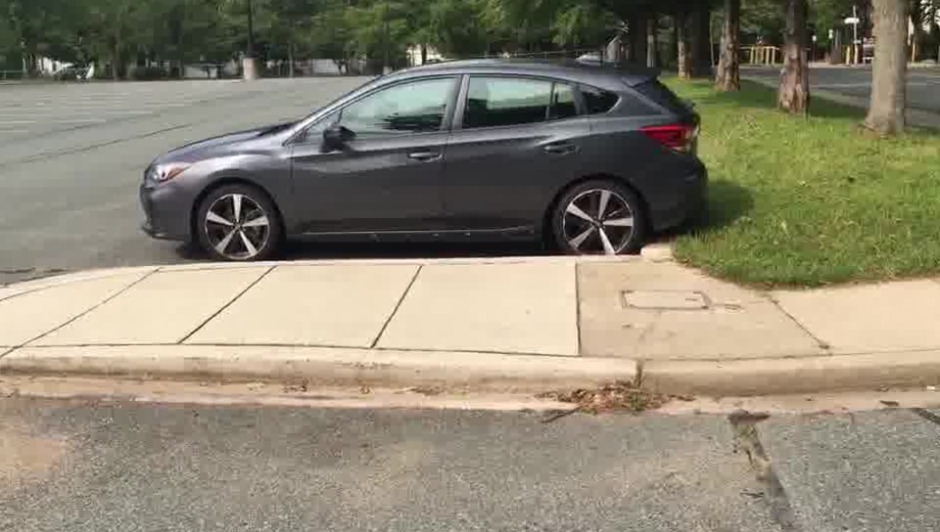}  &
     \includegraphics[trim= 0.5cm 3.5cm 0.5cm 4cm,clip, width=0.16\linewidth]{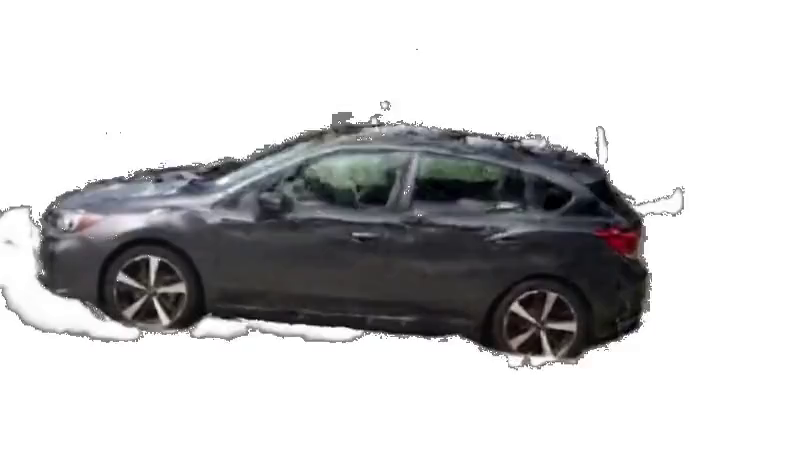}  &
          \includegraphics[trim= 0.5cm 3.5cm 0.5cm 4cm,clip, width=0.16\linewidth]{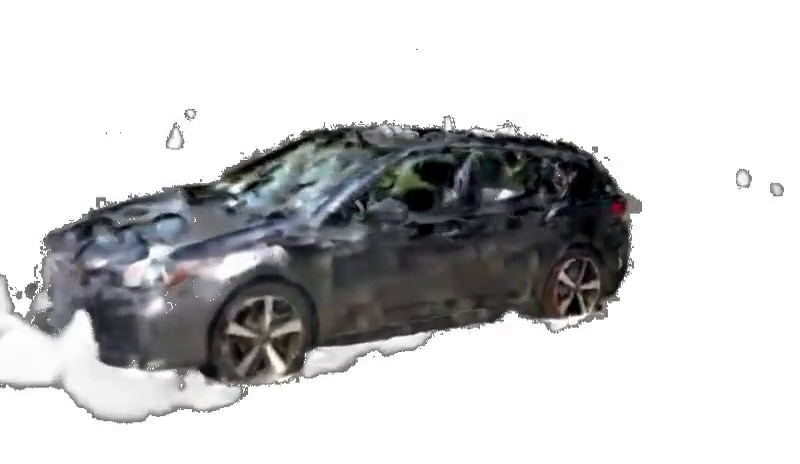} & 
                    \includegraphics[trim= 0.5cm 3.5cm 0.5cm 4cm,clip, width=0.16\linewidth]{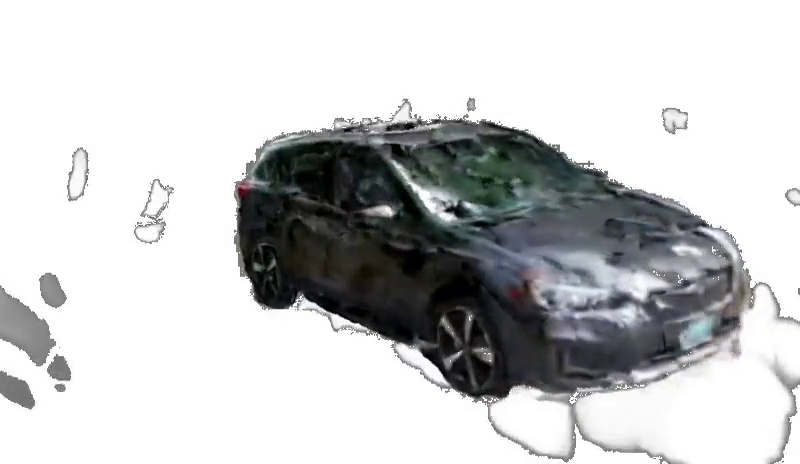} & 
          \includegraphics[trim= 0.5cm 3.5cm 0.5cm 4cm,clip, width=0.16\linewidth]{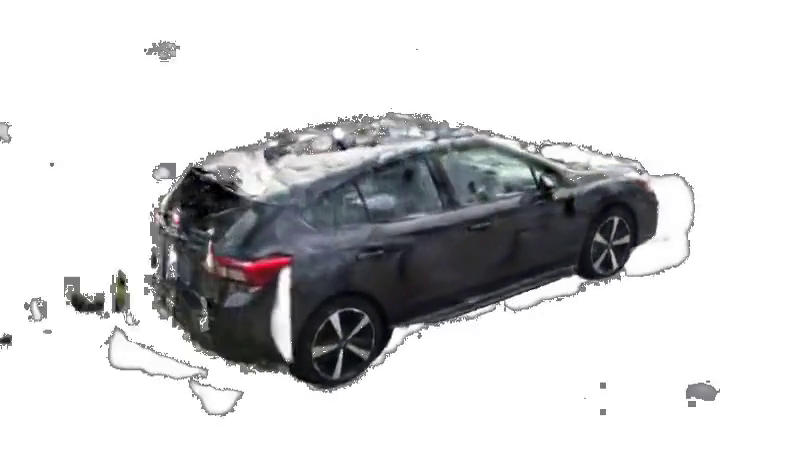}  &
          \includegraphics[trim= 0.5cm 3.5cm 0.5cm 4cm,clip, width=0.16\linewidth]{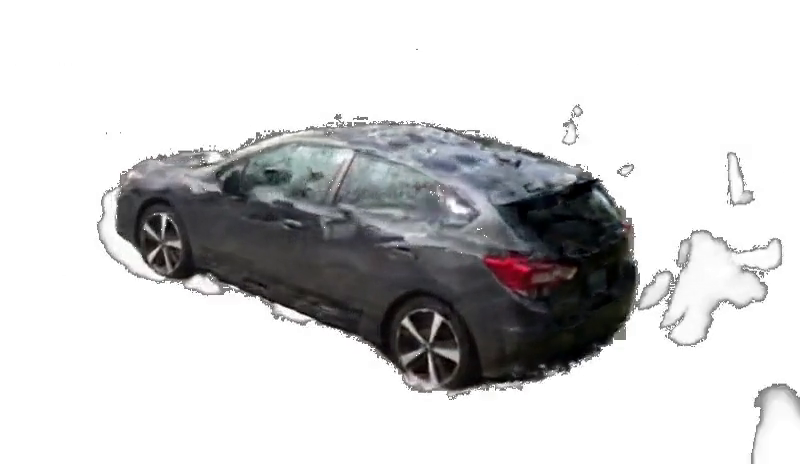}  \\
     \includegraphics[trim= 4.0cm 8cm 4cm 1.0cm,clip, width=0.16\linewidth]{images/real/s1/co.jpg}  &
     \includegraphics[trim= 2.0cm 5.5cm 2.0cm 4.5cm,clip, width=0.16\linewidth]{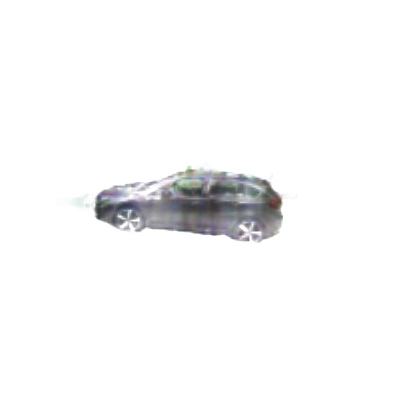}  &
          \includegraphics[trim= 2.0cm 5.5cm 2.0cm 4.5cm,clip, width=0.16\linewidth]{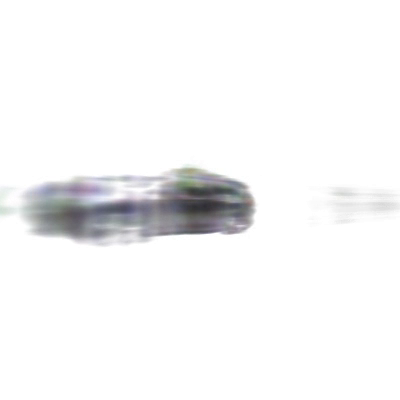} & 
                    \includegraphics[trim= 2.0cm 5.5cm 2.0cm 4.5cm,clip, width=0.16\linewidth]{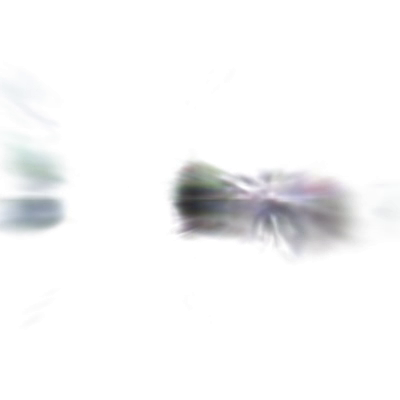} & 
          \includegraphics[trim= 2.0cm 5.5cm 2.0cm 4.5cm,clip, width=0.16\linewidth]{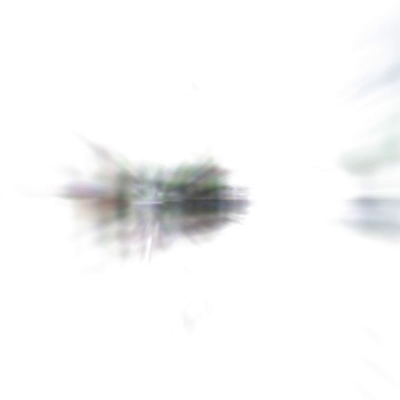}  &
          \includegraphics[trim= 2.0cm 5.5cm 2.0cm 4.5cm,clip, width=0.16\linewidth]{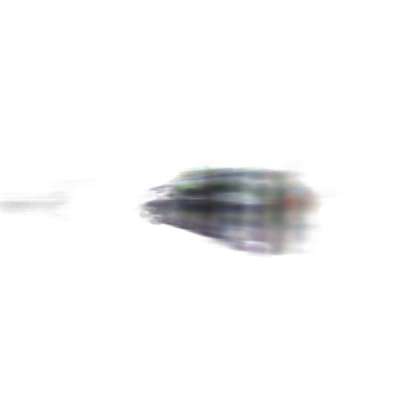}  \\ \midrule

     \includegraphics[trim= 1.0cm 2cm 1cm 2.0cm,clip, width=0.16\linewidth]{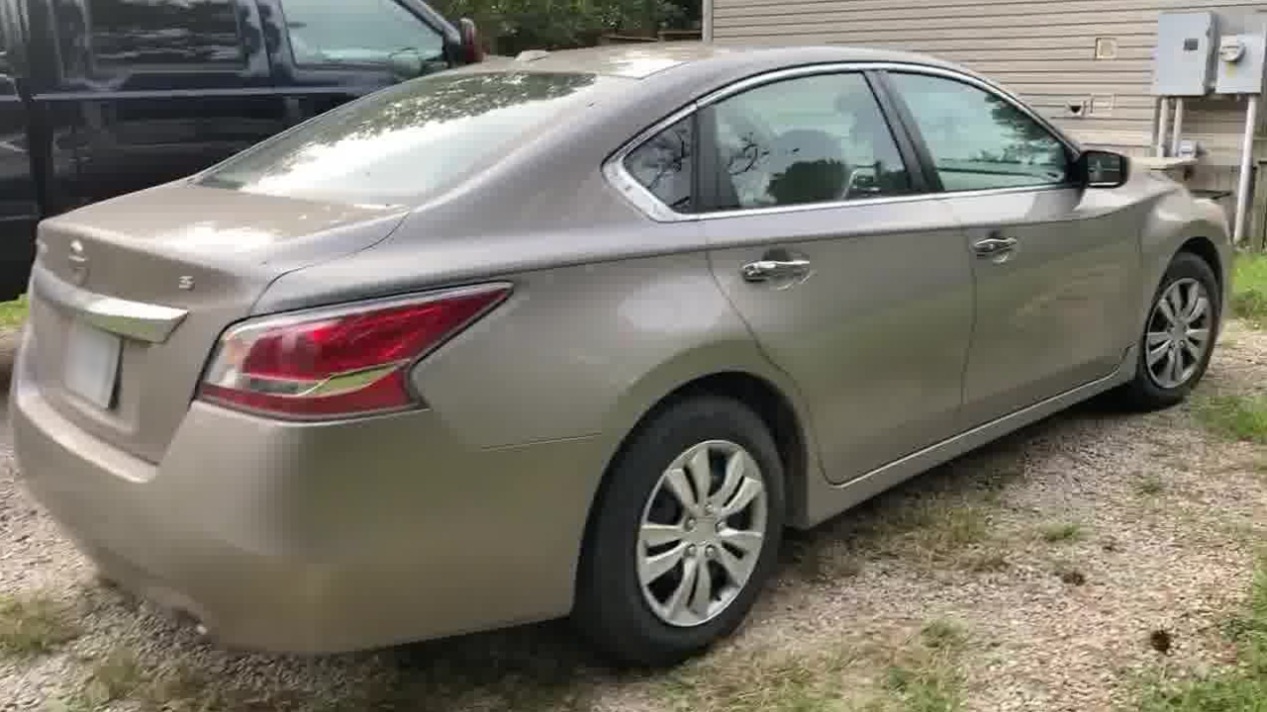}  &
     \includegraphics[trim= 0.5cm 0.5cm 0.5cm 0cm,clip, width=0.16\linewidth]{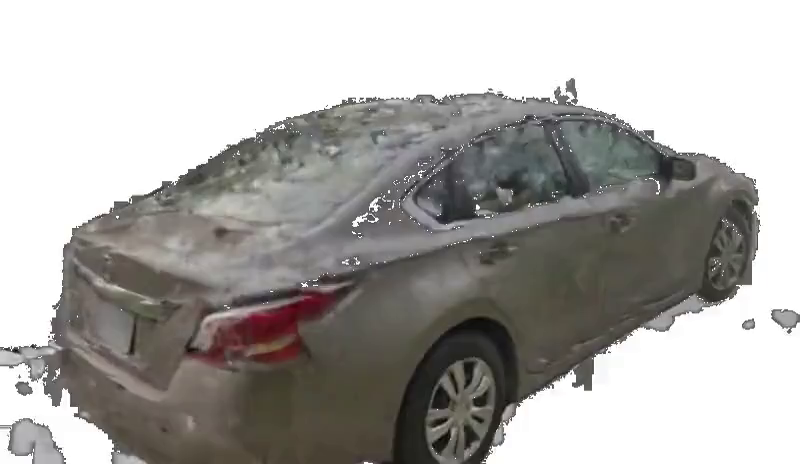}  &
          \includegraphics[trim= 0.5cm 0.5cm 0.5cm 0cm,clip, width=0.16\linewidth]{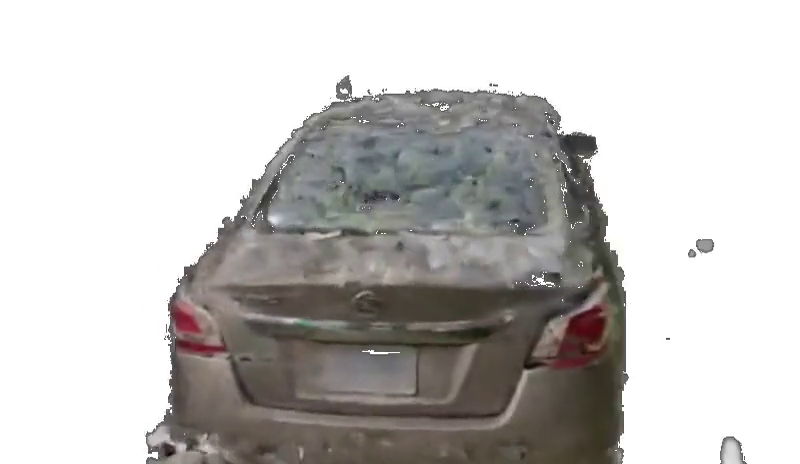} & 
                    \includegraphics[trim= 0.5cm 0.5cm 0.5cm 0cm,clip, width=0.16\linewidth]{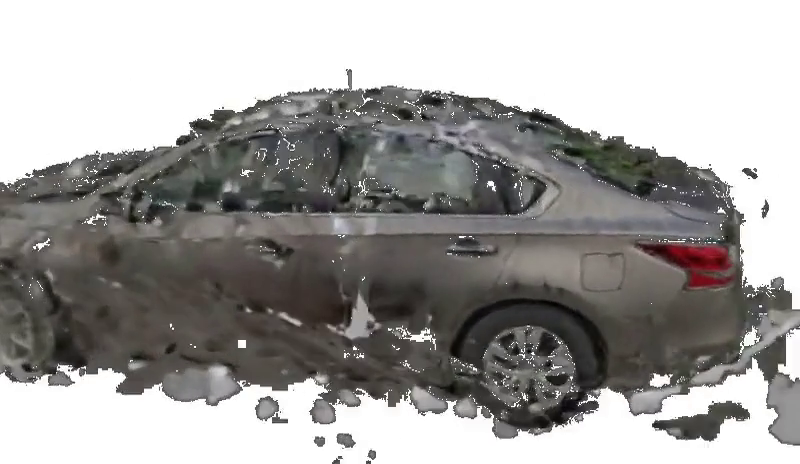} & 
          \includegraphics[trim= 0.5cm 0.5cm 0.5cm 0cm,clip, width=0.16\linewidth]{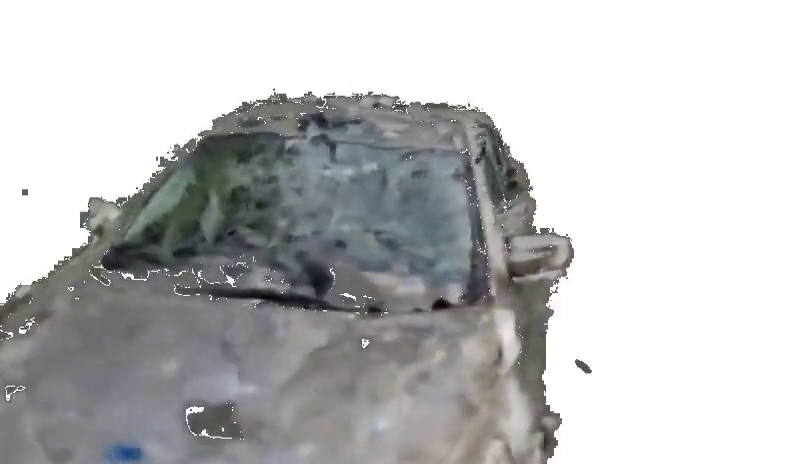}  &
          \includegraphics[trim= 0.5cm 0.5cm 0.5cm 0cm,clip, width=0.16\linewidth]{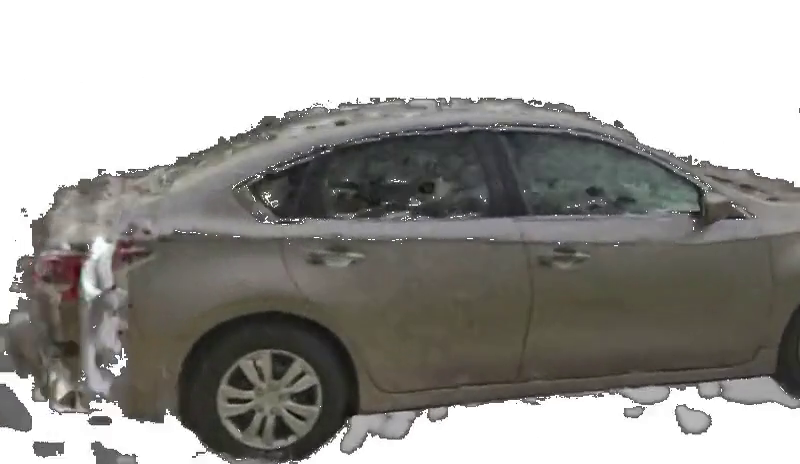}  \\
     \includegraphics[trim= 1.0cm 2cm 1cm 2.0cm,clip, width=0.16\linewidth]{images/real/s2/co.jpg}  &
     \includegraphics[trim= 2.0cm 5.5cm 2.0cm 4.5cm,clip, width=0.16\linewidth]{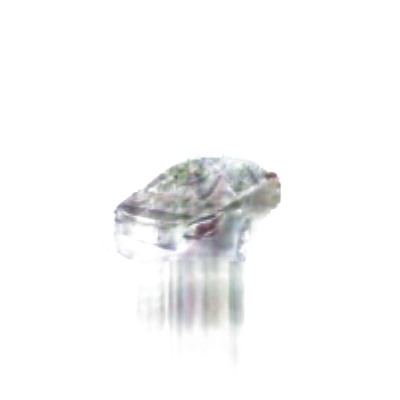}  &
          \includegraphics[trim= 2.0cm 5.5cm 2.0cm 4.5cm,clip, width=0.16\linewidth]{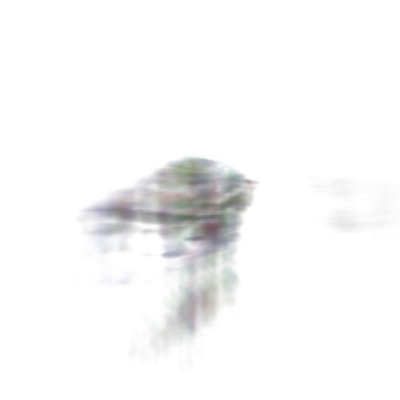} & 
                    \includegraphics[trim= 2.0cm 5.5cm 2.0cm 4.5cm,clip, width=0.16\linewidth]{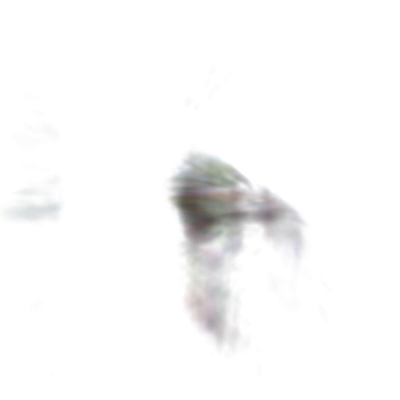} & 
          \includegraphics[trim= 2.0cm 5.5cm 2.0cm 4.5cm,clip, width=0.16\linewidth]{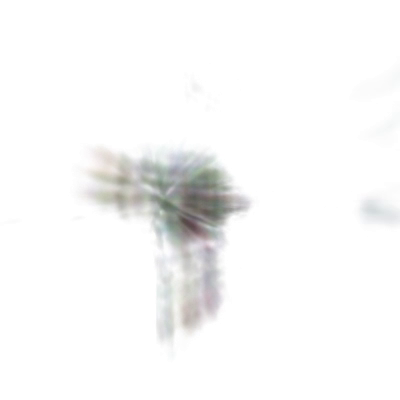}  &
          \includegraphics[trim= 2.0cm 5.5cm 2.0cm 4.5cm,clip, width=0.16\linewidth]{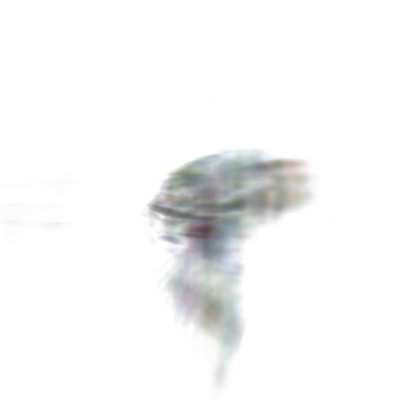}  \\ \midrule

     \includegraphics[trim= 1.0cm 2cm 1cm 2.0cm,clip, width=0.16\linewidth]{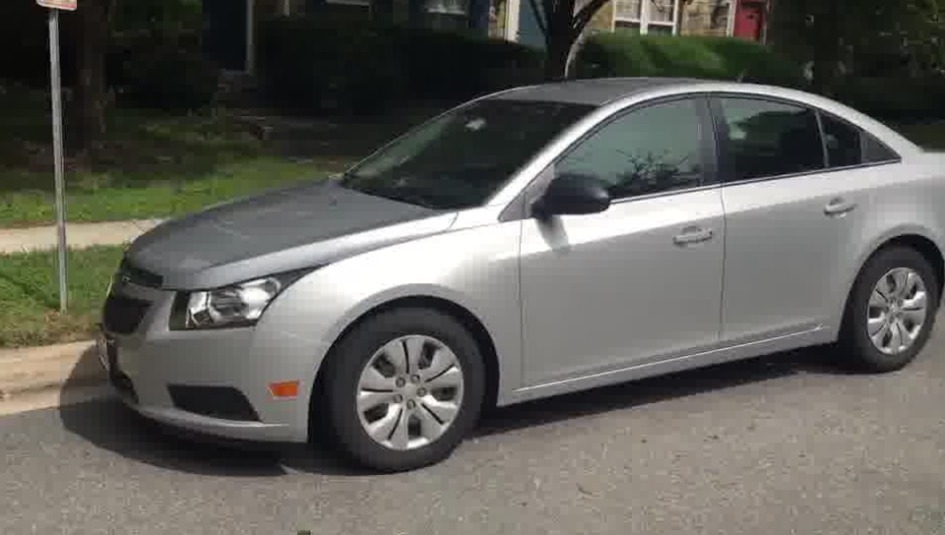}  &
     \includegraphics[trim= 0.5cm 0.5cm 0.5cm 0cm,clip, width=0.16\linewidth]{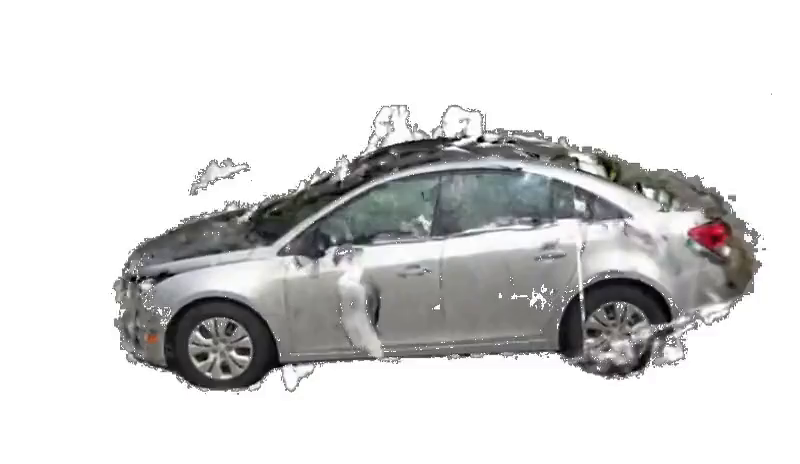}  &
          \includegraphics[trim= 0.5cm 0.5cm 0.5cm 0cm,clip, width=0.16\linewidth]{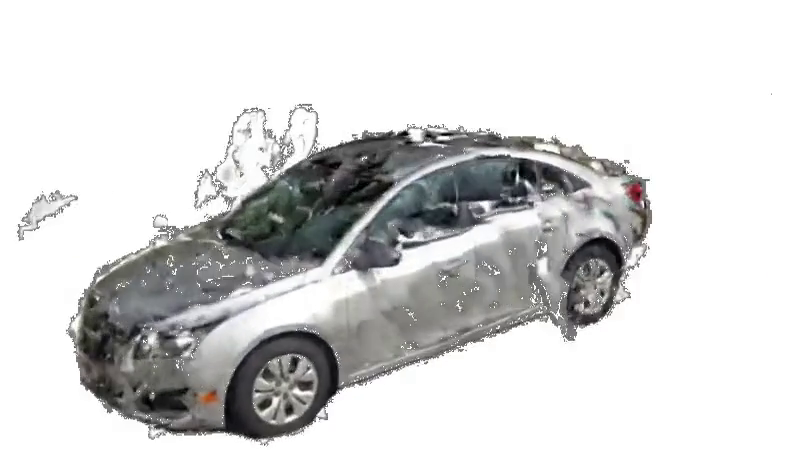} & 
          \includegraphics[trim= 0.5cm 0.5cm 0.5cm 0cm,clip, width=0.16\linewidth]{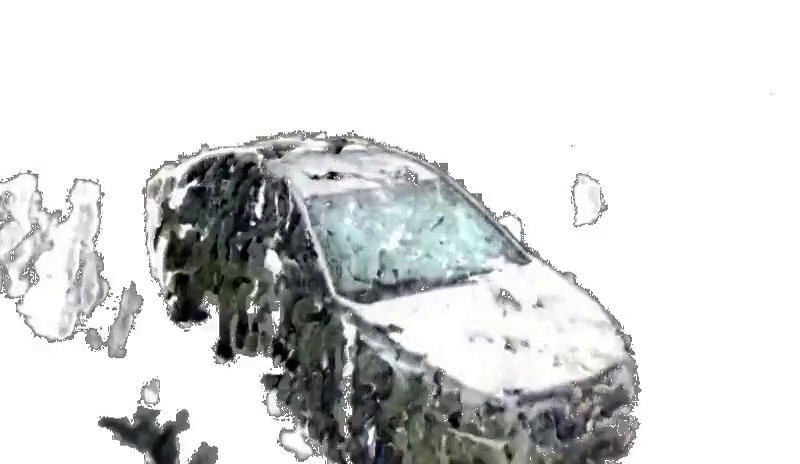} & 
          \includegraphics[trim= 0.5cm 0.5cm 0.5cm 0cm,clip, width=0.16\linewidth]{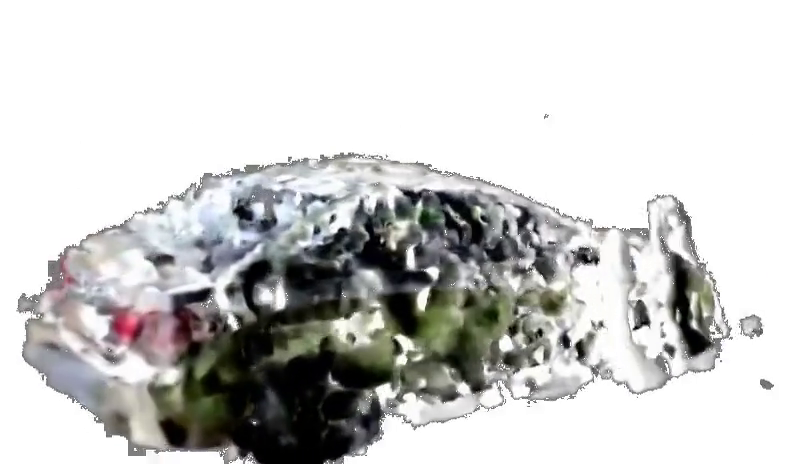}  &
          \includegraphics[trim= 0.5cm 0.5cm 0.5cm 0cm,clip, width=0.16\linewidth]{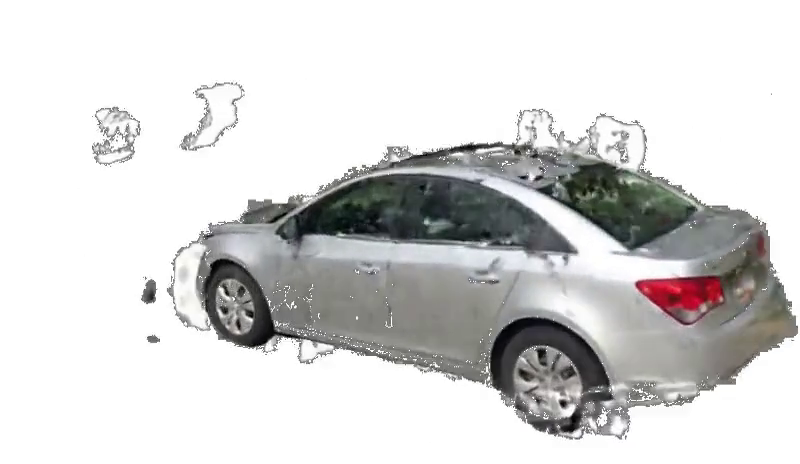}  \\
     \includegraphics[trim= 1.0cm 2cm 1cm 2.0cm,clip, width=0.16\linewidth]{images/real/s3/co.jpg}  &
     \includegraphics[trim= 3.0cm 5.5cm 3.0cm 4.0cm,clip, width=0.16\linewidth]{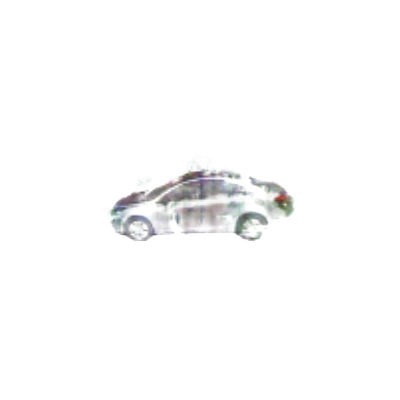}  &
          \includegraphics[trim= 3.0cm 5.5cm 3.0cm 4.0cm,clip, width=0.16\linewidth]{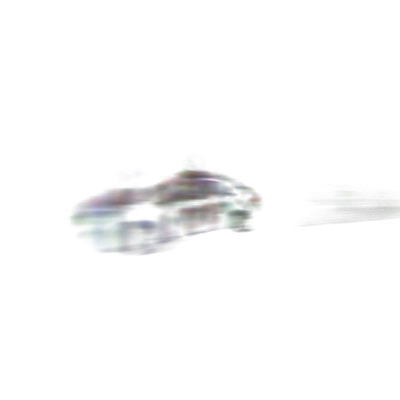} & 
          \includegraphics[trim= 3.0cm 5.5cm 3.0cm 4.0cm,clip, width=0.16\linewidth]{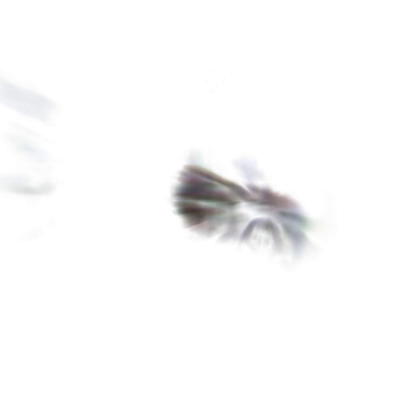} & 
          \includegraphics[trim= 3.0cm 5.5cm 3.0cm 4.0cm,clip, width=0.16\linewidth]{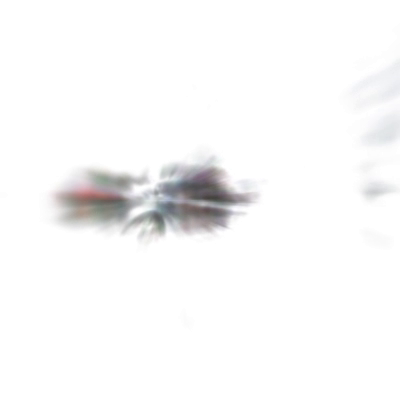}  &
          \includegraphics[trim= 3.0cm 5.5cm 3.0cm 4.0cm,clip, width=0.16\linewidth]{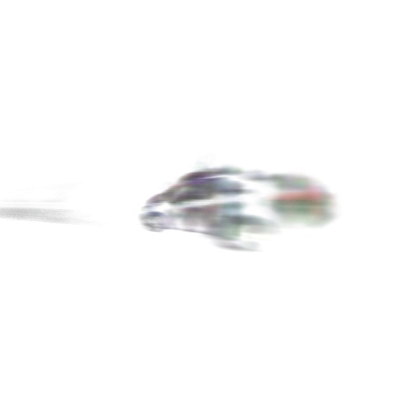}  \\

 \bottomrule
\end{tabular}
}
    \caption{\textbf{Real images} We show real images of Co3D \cite{co3d} (\textit{left}) and the corresponding generated views from our SRFs (\textit{rows' top part}) and pretrained PixelNeRF \cite{pixelnerf} (\textit{rows' bottom part}).
    }
    \label{fig:sup-co3d}
\end{figure*}

Additional results of normal test tracks benchmark of SPARF are presented in Table \ref{tbl:test}. Please see figures \ref{fig:cameras} and \ref{fig:ood} for differences between the normal train/test track and the OOD hard track .More comparisons and generations are provided in Figures \ref{fig:comparison1},\ref{fig:comparison2},\ref{figsup:interpolation}. Regarding real images, we ahow more Co3D images and their SRFs and PixelNeRF reconstruction in \figLabel{\ref{fig:sup-co3d}}. Note that the goal of this figure is to demonstrate the quality of the renderings form proper SRFs (similar ot the ones used in training SuRFNet), compared to a 2D-based network ( like PixelNeRF \cite{pixelnerf}). It is not used to evaluate generalization ability from few input views on real images, but to show the potential of training on real images. 

one important aspect to consider is the \textit{3D consistency} of our SuRFNet renderings compared to the other 2D methods, especially when moving \textit{out-of-distribution of the training views}. This is one of the most important aspects we investigate in our work that previous works in the literature have overlooked.  Figure \ref{fig:arrow} demonstrates that as the testing rendered views move outside of the training distribution (going right), SuRFNet generally produces more consistent renderings than all previous 2D-based methods [\cite{pixelnerf,visionnerf}].

\begin{figure*}
    \centering
        \tabcolsep=0.03cm
\resizebox{0.9\linewidth}{!}{
\begin{tabular}{c|c|c}
SPARF (ours) & SRN \cite{srn}  & NMR \cite{dvr} \\ \hline
     \includegraphics[trim= 0.0cm 0cm 0cm 0cm,clip, width=0.33\linewidth]{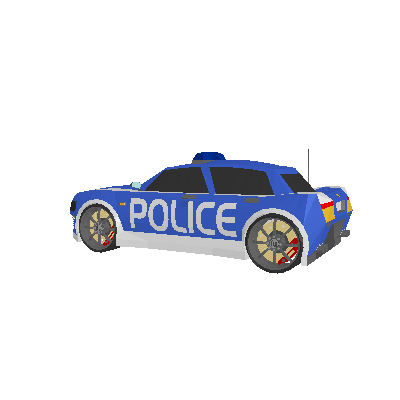} &
     \includegraphics[trim= 0.0cm 0cm 0cm 0cm,clip, width=0.33\linewidth]{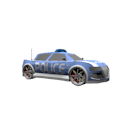}&
     \includegraphics[trim= 0.0cm 0cm 0cm 0cm,clip, width=0.33\linewidth]{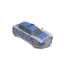} \\
     
     \includegraphics[trim= 0.0cm 0cm 0cm 0cm,clip, width=0.33\linewidth]{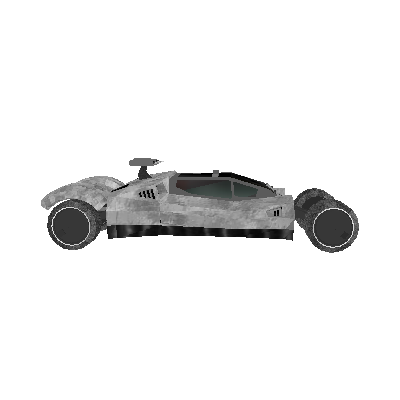} &
     \includegraphics[trim= 0.0cm 0cm 0cm 0cm,clip, width=0.33\linewidth]{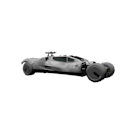}&
     \includegraphics[trim= 0.0cm 0cm 0cm 0cm,clip, width=0.33\linewidth]{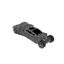} \\

     \includegraphics[trim= 0.0cm 0cm 0cm 0cm,clip, width=0.33\linewidth]{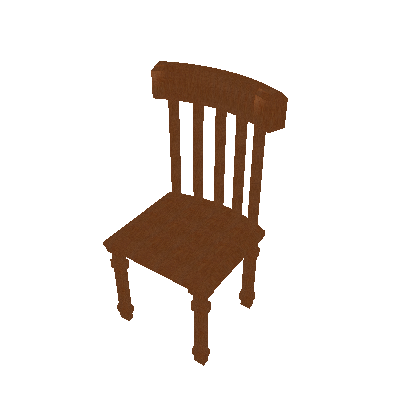} &
     \includegraphics[trim= 0.0cm 0cm 0cm 0cm,clip, width=0.33\linewidth]{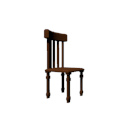}&
     \includegraphics[trim= 0.0cm 0cm 0cm 0cm,clip, width=0.33\linewidth]{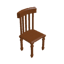} \\

     \includegraphics[trim= 0.0cm 0cm 0cm 0cm,clip, width=0.33\linewidth]{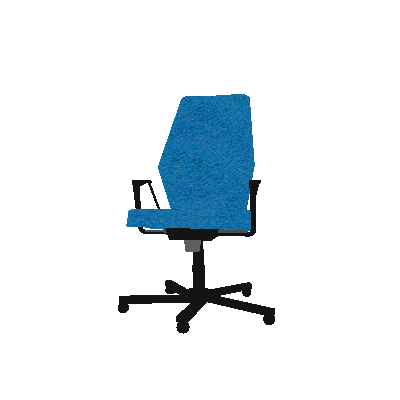} &
     \includegraphics[trim= 0.0cm 0cm 0cm 0cm,clip, width=0.33\linewidth]{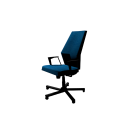}&
     \includegraphics[trim= 0.0cm 0cm 0cm 0cm,clip, width=0.33\linewidth]{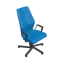} \\

\bottomrule
\end{tabular}
} %
    \caption{\textbf{SPARF \vs other Datasets }. SPARF offers a large-scale high-resolution dataset compared to other posed multi-view datasets.  Note that SRN \cite{srn} has only cars and chairs, while NMR \cite{dvr} and SPARF has 13 classes. 
    }
    \label{figsup:datasets}
\end{figure*}
\begin{figure*}
    \centering
    \includegraphics[trim= 0cm 0cm 0cm 0cm,clip, width=0.19\linewidth]{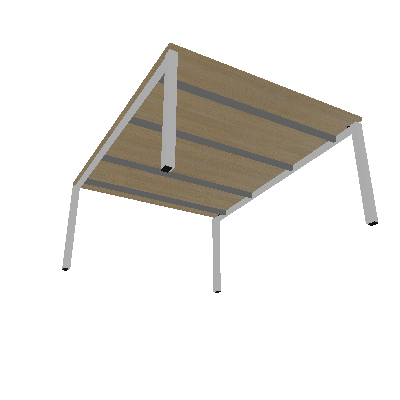}
    \includegraphics[trim= 0cm 0cm 0cm 0cm,clip, width=0.19\linewidth]{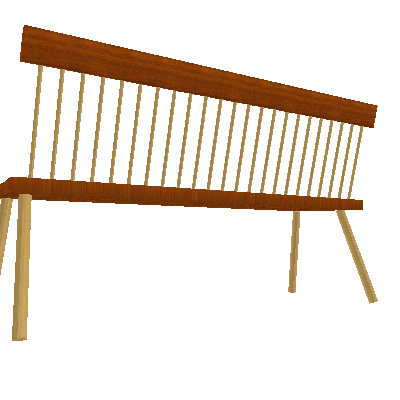}
    \includegraphics[trim= 0cm 0cm 0cm 0cm,clip, width=0.19\linewidth]{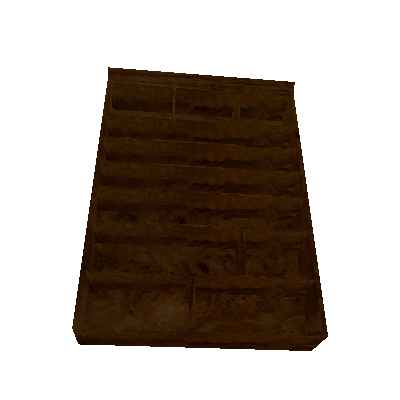}
    \includegraphics[trim= 0cm 0cm 0cm 0cm,clip, width=0.19\linewidth]{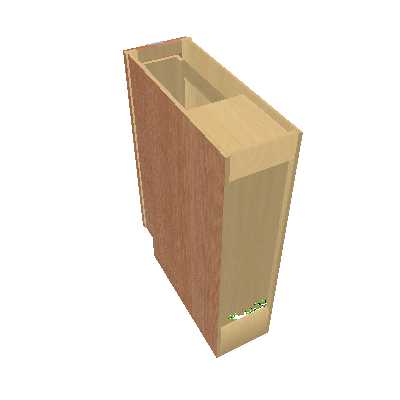}
    \includegraphics[trim= 0cm 0cm 0cm 0cm,clip, width=0.19\linewidth]{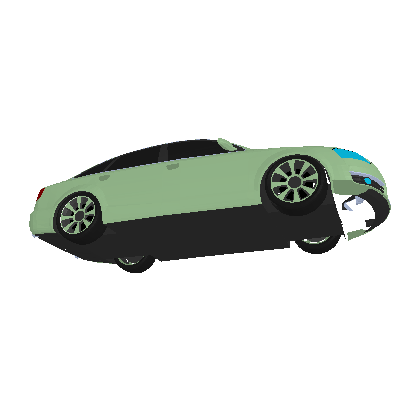}
    \includegraphics[trim= 0cm 0cm 0cm 0cm,clip, width=0.19\linewidth]{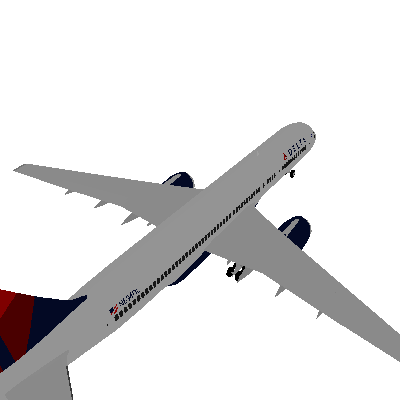}
    \includegraphics[trim= 0cm 0cm 0cm 0cm,clip, width=0.19\linewidth]{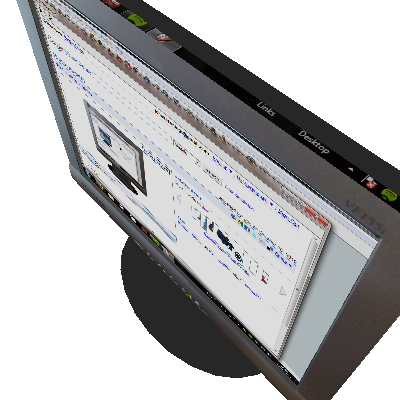}
    \includegraphics[trim= 0cm 0cm 0cm 0cm,clip, width=0.19\linewidth]{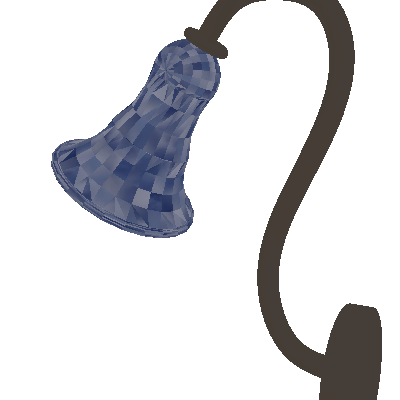}
    \includegraphics[trim= 0cm 0cm 0cm 0cm,clip, width=0.19\linewidth]{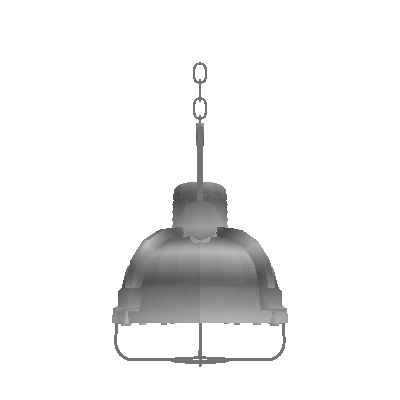}
    \includegraphics[trim= 0cm 0cm 0cm 0cm,clip, width=0.19\linewidth]{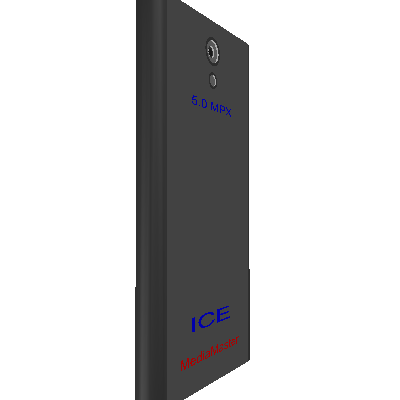}
    \includegraphics[trim= 0cm 0cm 0cm 0cm,clip, width=0.19\linewidth]{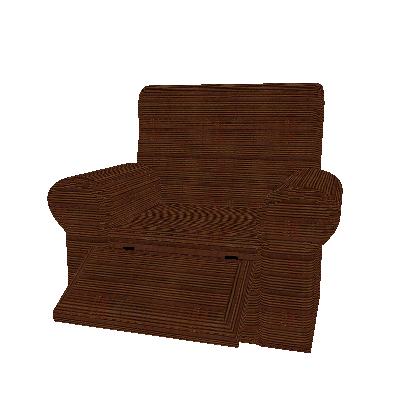}
    \includegraphics[trim= 0cm 0cm 0cm 0cm,clip, width=0.19\linewidth]{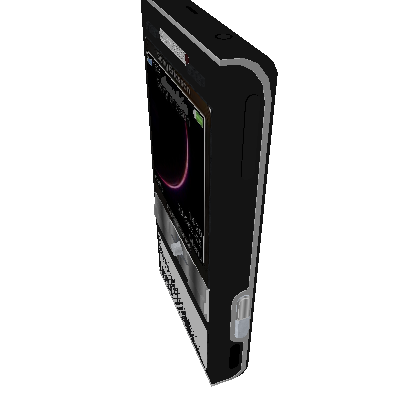}
    \includegraphics[trim= 0cm 0cm 0cm 0cm,clip, width=0.19\linewidth]{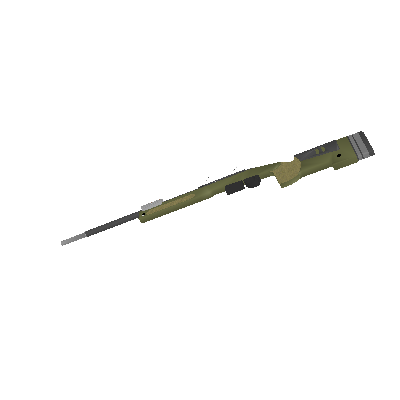}
    \includegraphics[trim= 0cm 0cm 0cm 0cm,clip, width=0.19\linewidth]{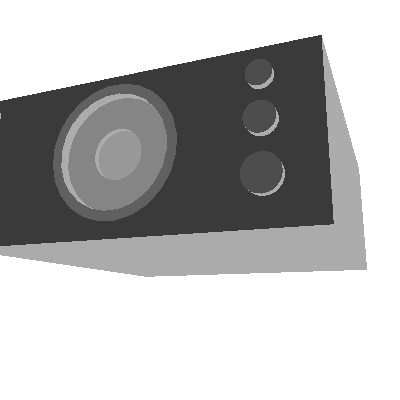}
    \includegraphics[trim= 0cm 0cm 0cm 0cm,clip, width=0.19\linewidth]{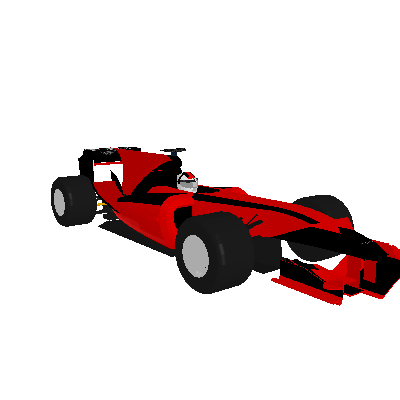}
    \includegraphics[trim= 0cm 0cm 0cm 0cm,clip, width=0.19\linewidth]{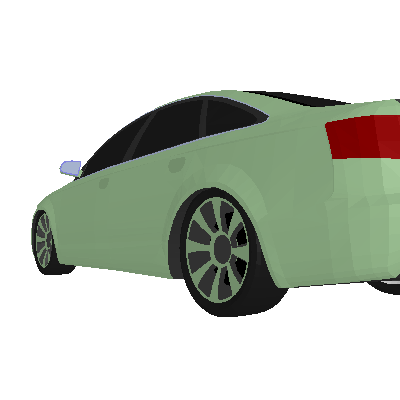}
    \includegraphics[trim= 0cm 0cm 0cm 0cm,clip, width=0.19\linewidth]{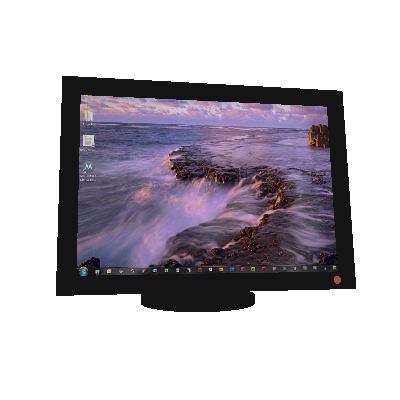}
    \includegraphics[trim= 0cm 0cm 0cm 0cm,clip, width=0.19\linewidth]{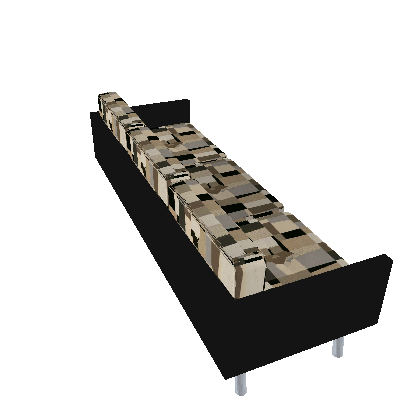}
    \includegraphics[trim= 0cm 0cm 0cm 0cm,clip, width=0.19\linewidth]{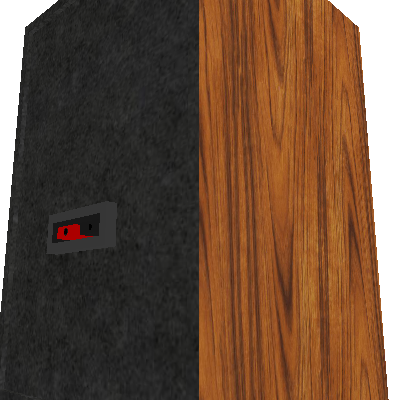}
    \includegraphics[trim= 0cm 0cm 0cm 0cm,clip, width=0.19\linewidth]{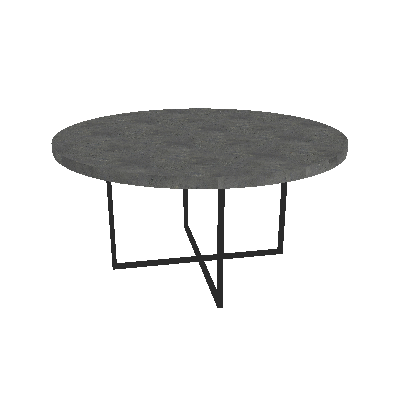}
    \includegraphics[trim= 0cm 0cm 0cm 0cm,clip, width=0.19\linewidth]{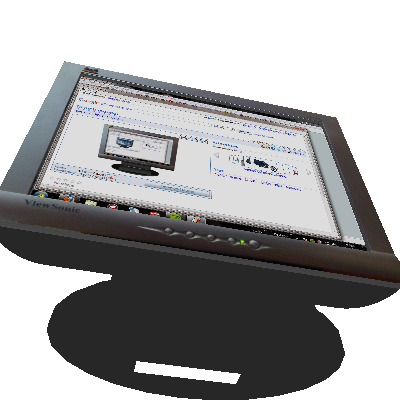}
    \includegraphics[trim= 0cm 0cm 0cm 0cm,clip, width=0.19\linewidth]{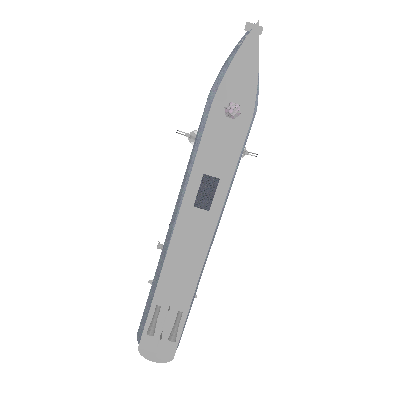}
    \includegraphics[trim= 0cm 0cm 0cm 0cm,clip, width=0.19\linewidth]{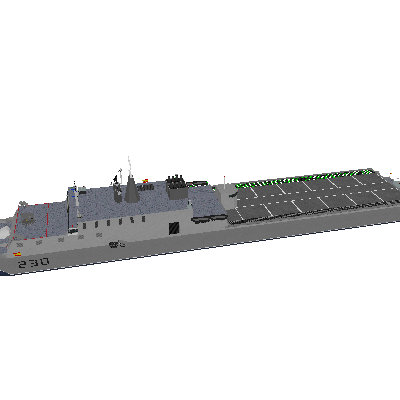}
    \includegraphics[trim= 0cm 0cm 0cm 0cm,clip, width=0.19\linewidth]{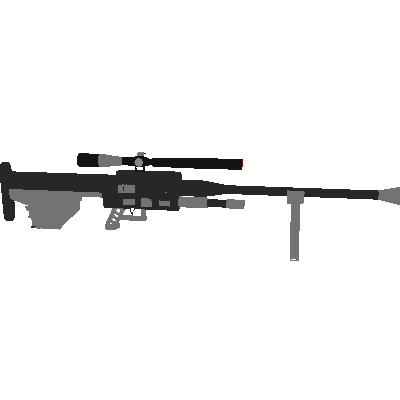}
    \includegraphics[trim= 0cm 0cm 0cm 0cm,clip, width=0.19\linewidth]{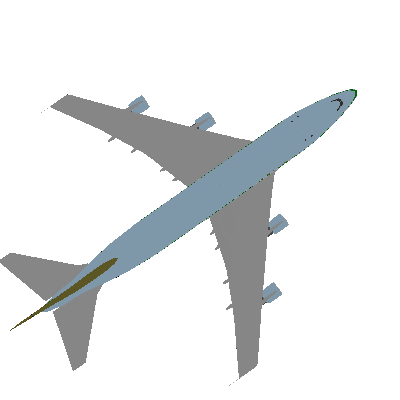}
    \includegraphics[trim= 0cm 0cm 0cm 0cm,clip, width=0.19\linewidth]{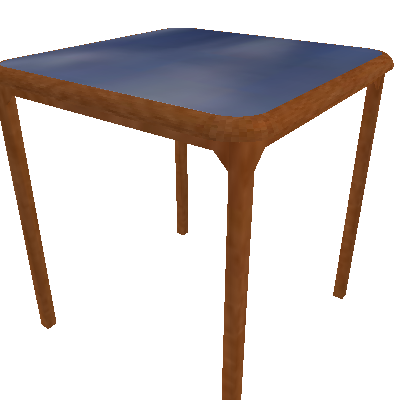}
    \includegraphics[trim= 0cm 0cm 0cm 0cm,clip, width=0.19\linewidth]{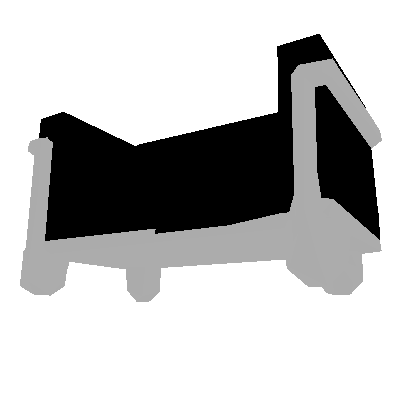}
    \includegraphics[trim= 0cm 0cm 0cm 0cm,clip, width=0.19\linewidth]{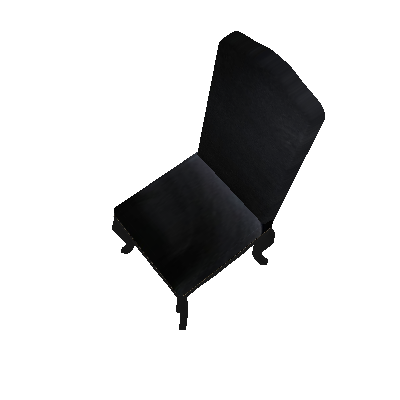}
    \includegraphics[trim= 0cm 0cm 0cm 0cm,clip, width=0.19\linewidth]{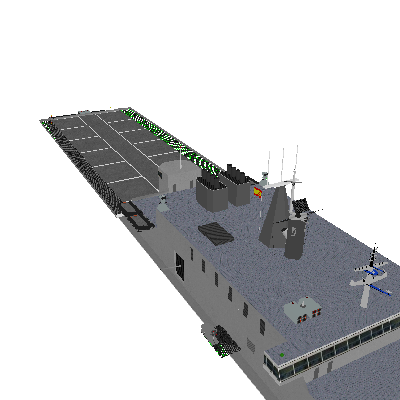}
    \includegraphics[trim= 0cm 0cm 0cm 0cm,clip, width=0.19\linewidth]{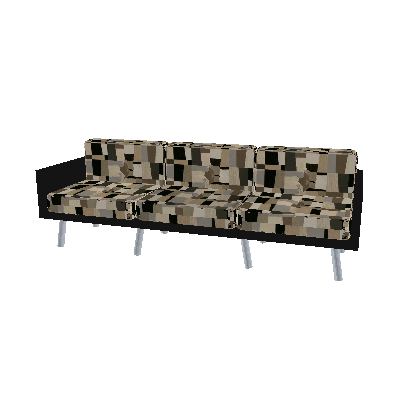}
    \caption{\textbf{SPARF: a Large Dataset for 3D Shapes Radiance Fields and Novel Views Synthesis}.  
    }
    \label{figsup:sparf}
\end{figure*}

\begin{figure*}
    \centering
    \tabcolsep=0.03cm
\resizebox{0.9\linewidth}{!}{
\begin{tabular}{cccccc}
     \includegraphics[trim= 0.0cm 0cm 0cm 0cm,clip, width=0.2\linewidth]{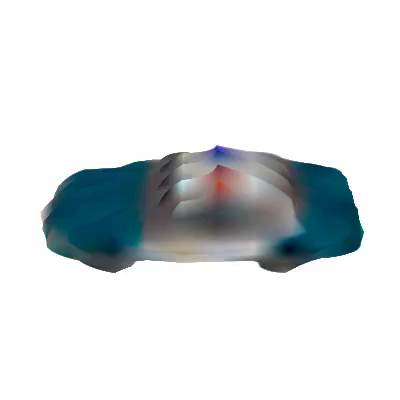}  &
     \includegraphics[trim= 0.0cm 0cm 0cm 0cm,clip, width=0.2\linewidth]{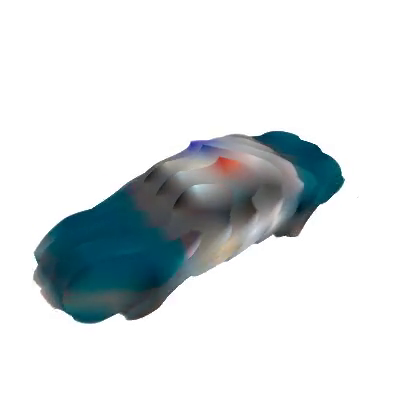}  &
     \includegraphics[trim= 0.0cm 0cm 0cm 0cm,clip, width=0.2\linewidth]{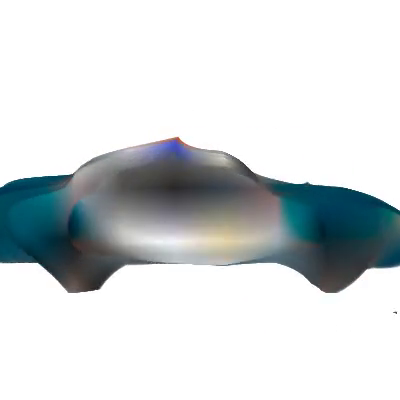}  &
     \includegraphics[trim= 0.0cm 0cm 0cm 0cm,clip, width=0.2\linewidth]{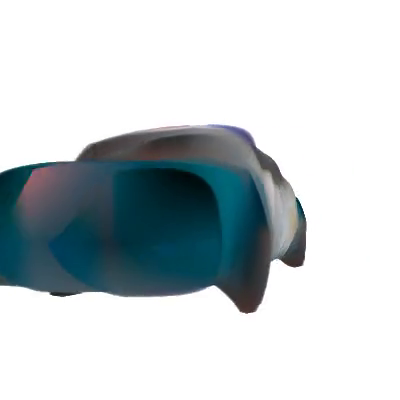}  & 
     \includegraphics[trim= 0.0cm 0cm 0cm 0cm,clip, width=0.2\linewidth]{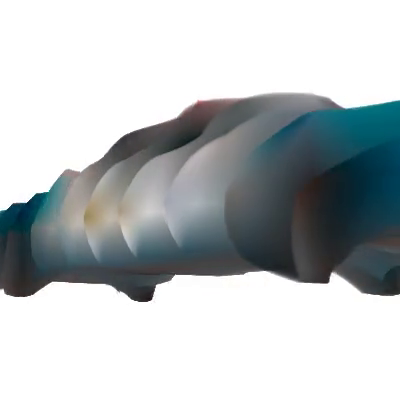}  & 
     \includegraphics[trim= 0.0cm 0cm 0cm 0cm,clip, width=0.2\linewidth]{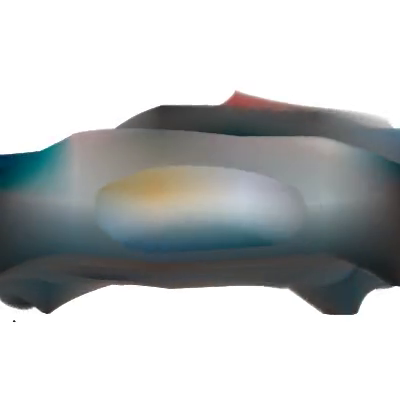}   \\
     \includegraphics[trim= 0.0cm 0cm 0cm 0cm,clip, width=0.2\linewidth]{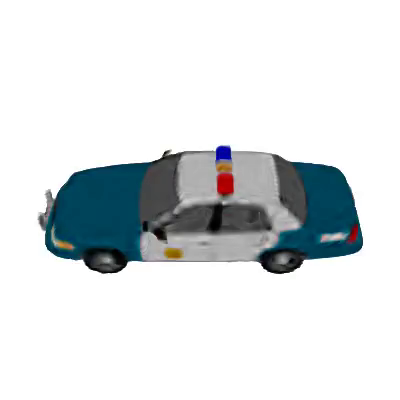}  &
     \includegraphics[trim= 0.0cm 0cm 0cm 0cm,clip, width=0.2\linewidth]{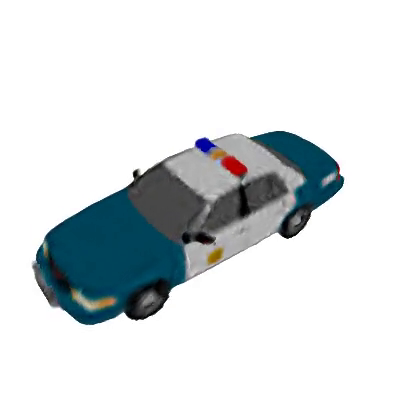}  &
     \includegraphics[trim= 0.0cm 0cm 0cm 0cm,clip, width=0.2\linewidth]{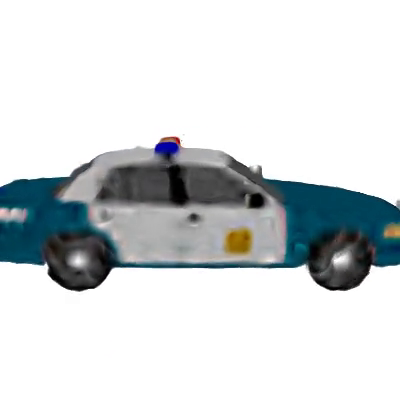}  &
     \includegraphics[trim= 0.0cm 0cm 0cm 0cm,clip, width=0.2\linewidth]{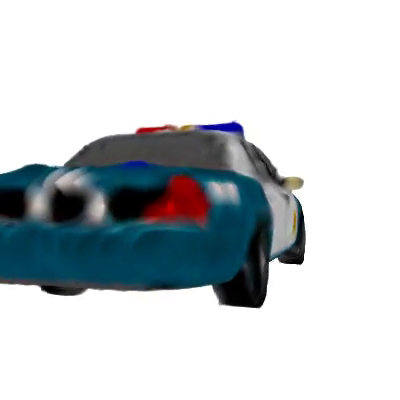}  & 
     \includegraphics[trim= 0.0cm 0cm 0cm 0cm,clip, width=0.2\linewidth]{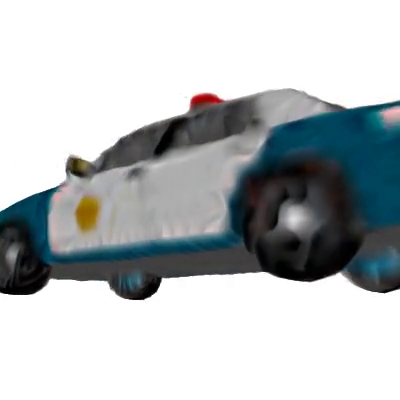}  & 
     \includegraphics[trim= 0.0cm 0cm 0cm 0cm,clip, width=0.2\linewidth]{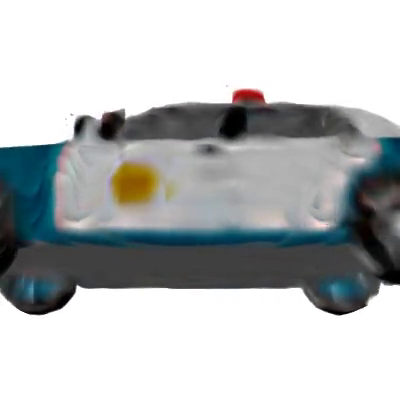}   \\
     \includegraphics[trim= 0.0cm 0cm 0cm 0cm,clip, width=0.2\linewidth]{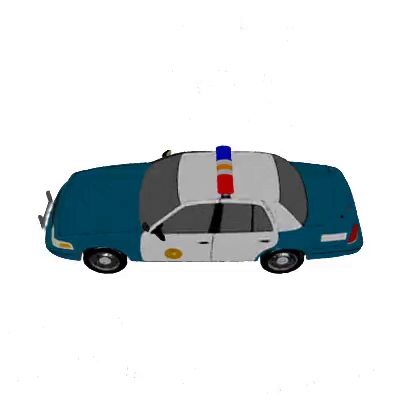}  &
     \includegraphics[trim= 0.0cm 0cm 0cm 0cm,clip, width=0.2\linewidth]{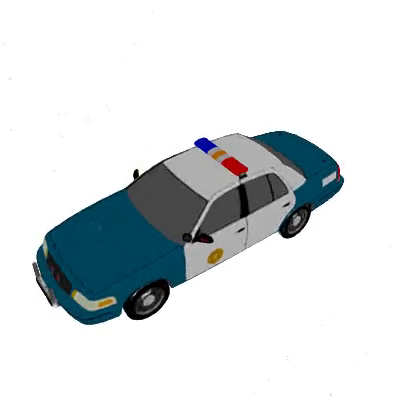}  &
     \includegraphics[trim= 0.0cm 0cm 0cm 0cm,clip, width=0.2\linewidth]{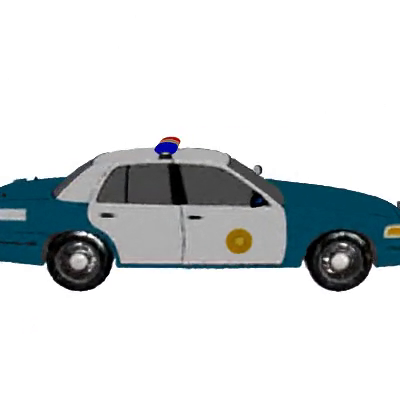}  &
     \includegraphics[trim= 0.0cm 0cm 0cm 0cm,clip, width=0.2\linewidth]{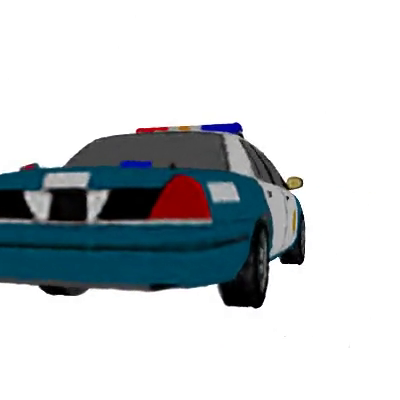}  & 
     \includegraphics[trim= 0.0cm 0cm 0cm 0cm,clip, width=0.2\linewidth]{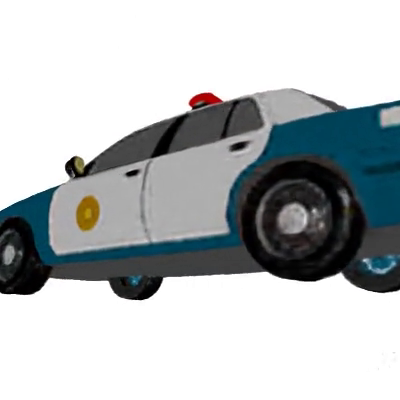}  & 
     \includegraphics[trim= 0.0cm 0cm 0cm 0cm,clip, width=0.2\linewidth]{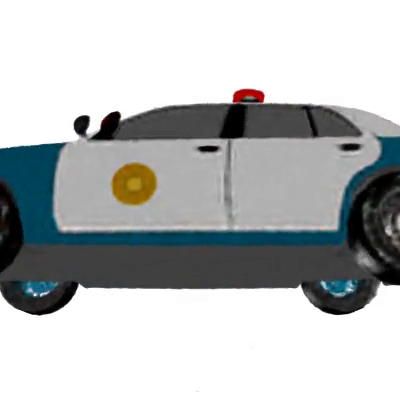}   \\

     \hline
     \includegraphics[trim= 0.0cm 0cm 0cm 0cm,clip, width=0.2\linewidth]{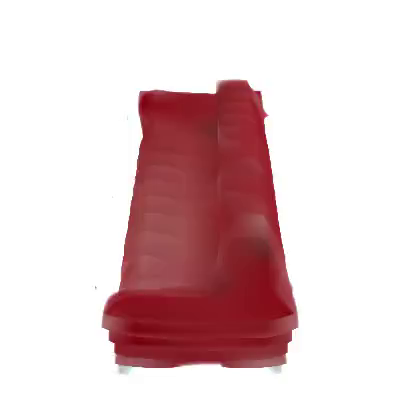}  &
     \includegraphics[trim= 0.0cm 0cm 0cm 0cm,clip, width=0.2\linewidth]{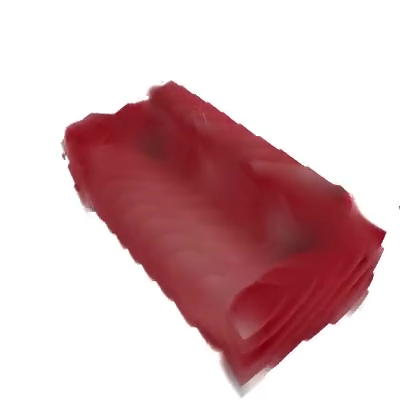}  &
     \includegraphics[trim= 0.0cm 0cm 0cm 0cm,clip, width=0.2\linewidth]{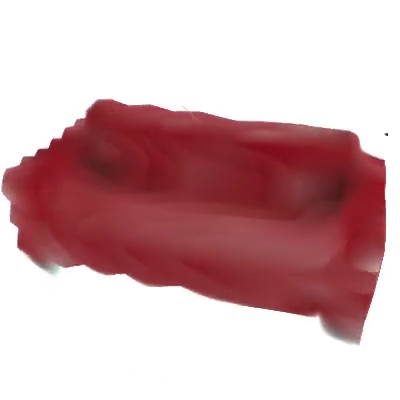}  &
     \includegraphics[trim= 0.0cm 0cm 0cm 0cm,clip, width=0.2\linewidth]{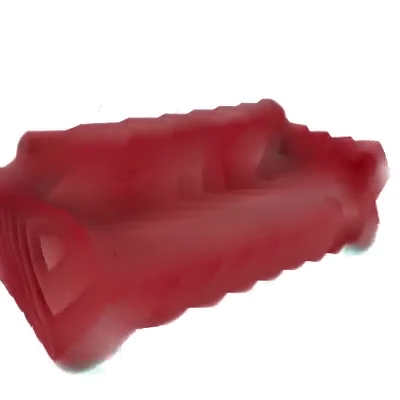}  &
     \includegraphics[trim= 0.0cm 0cm 0cm 0cm,clip, width=0.2\linewidth]{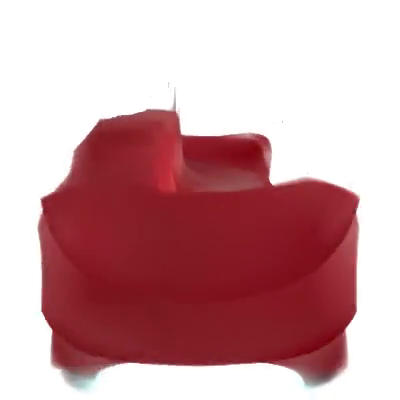}  & 
     \includegraphics[trim= 0.0cm 0cm 0cm 0cm,clip, width=0.2\linewidth]{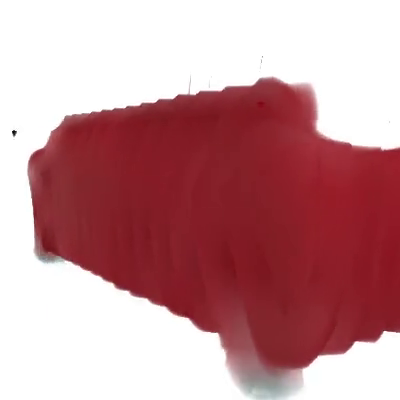}   \\
     \includegraphics[trim= 0.0cm 0cm 0cm 0cm,clip, width=0.2\linewidth]{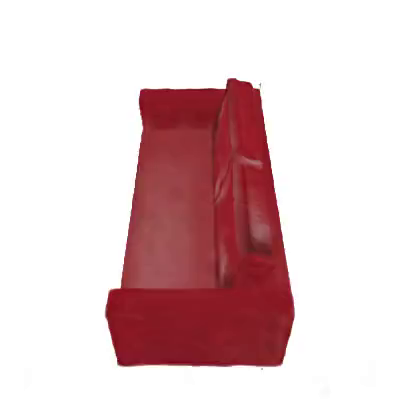}  &
     \includegraphics[trim= 0.0cm 0cm 0cm 0cm,clip, width=0.2\linewidth]{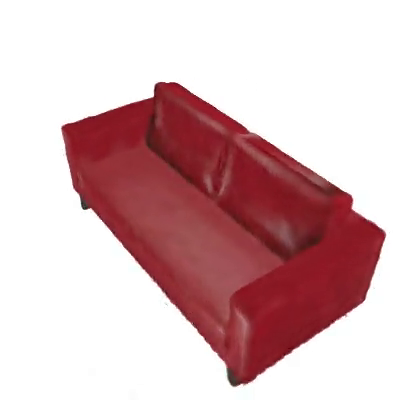}  &
     \includegraphics[trim= 0.0cm 0cm 0cm 0cm,clip, width=0.2\linewidth]{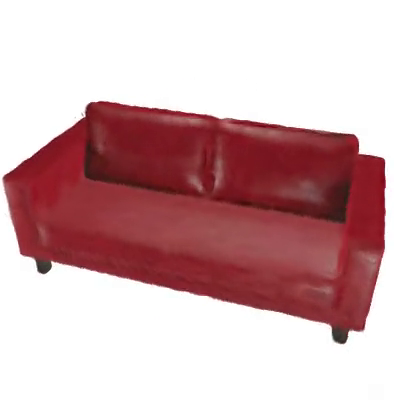}  &
     \includegraphics[trim= 0.0cm 0cm 0cm 0cm,clip, width=0.2\linewidth]{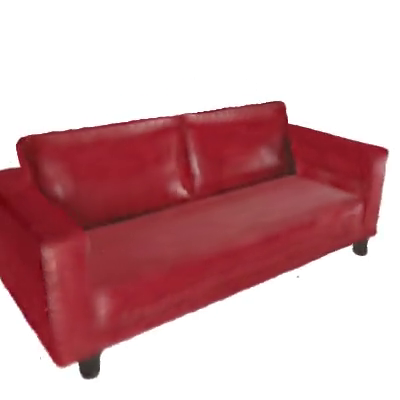}  &
     \includegraphics[trim= 0.0cm 0cm 0cm 0cm,clip, width=0.2\linewidth]{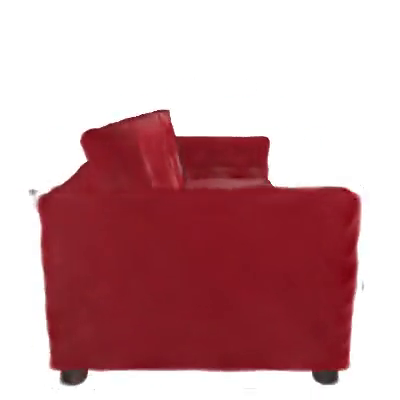}  & 
     \includegraphics[trim= 0.0cm 0cm 0cm 0cm,clip, width=0.2\linewidth]{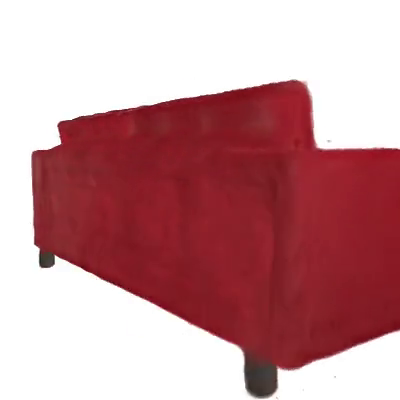}   \\
     \includegraphics[trim= 0.0cm 0cm 0cm 0cm,clip, width=0.2\linewidth]{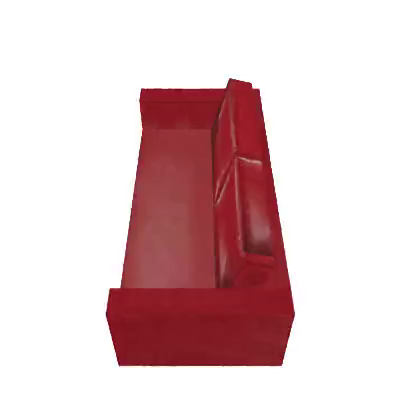}  &
     \includegraphics[trim= 0.0cm 0cm 0cm 0cm,clip, width=0.2\linewidth]{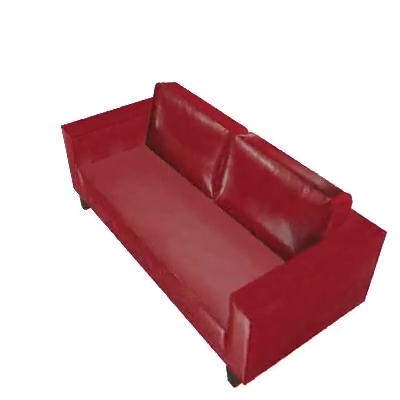}  &
     \includegraphics[trim= 0.0cm 0cm 0cm 0cm,clip, width=0.2\linewidth]{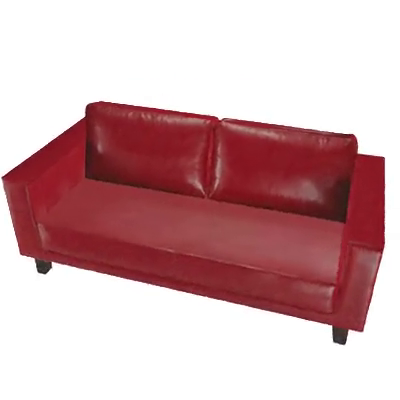}  &
     \includegraphics[trim= 0.0cm 0cm 0cm 0cm,clip, width=0.2\linewidth]{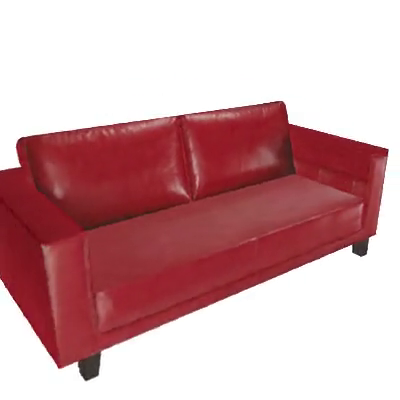}  &
     \includegraphics[trim= 0.0cm 0cm 0cm 0cm,clip, width=0.2\linewidth]{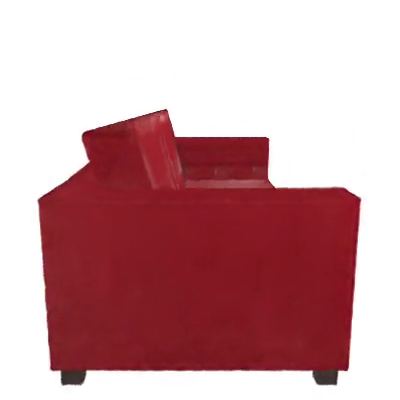}  & 
     \includegraphics[trim= 0.0cm 0cm 0cm 0cm,clip, width=0.2\linewidth]{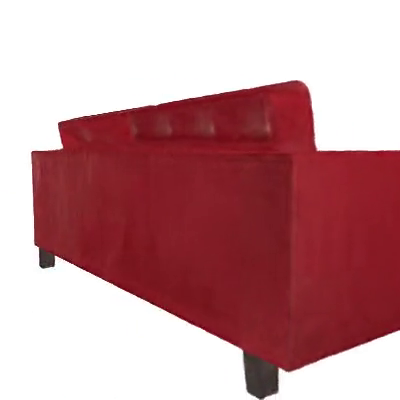}   \\
\bottomrule
\end{tabular}
} %
    \caption{\textbf{SRFs: The optimized Sparse Radiance Fields in SPARF 1}.  A total of one million SRFs have been collected in SPARF, including on multiple voxel resolutions: 32 (\textit{\textit{top}}), 128 (\textit{middle}), and 512 (\textit{bottom}) for every 3D shape.
    }
    \label{fig:srf1}
\end{figure*}
\begin{figure*}
    \centering
    \tabcolsep=0.03cm
\resizebox{0.9\linewidth}{!}{
\begin{tabular}{cccccc}
     \includegraphics[trim= 0.0cm 0cm 0cm 0cm,clip, width=0.2\linewidth]{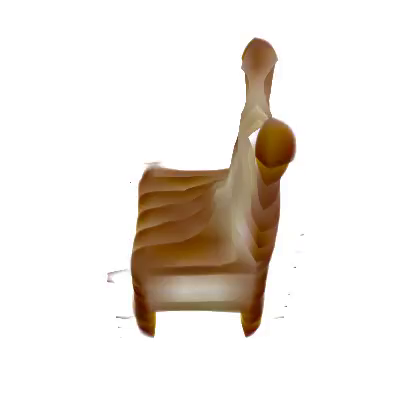}  &
     \includegraphics[trim= 0.0cm 0cm 0cm 0cm,clip, width=0.2\linewidth]{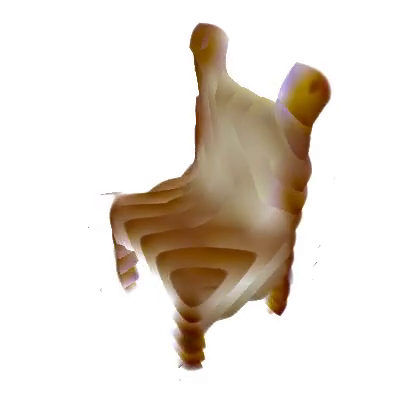}  &
     \includegraphics[trim= 0.0cm 0cm 0cm 0cm,clip, width=0.2\linewidth]{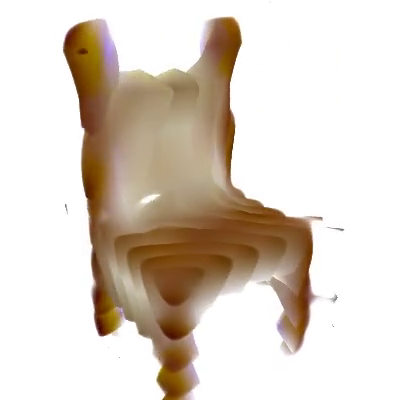}  &
     \includegraphics[trim= 0.0cm 0cm 0cm 0cm,clip, width=0.2\linewidth]{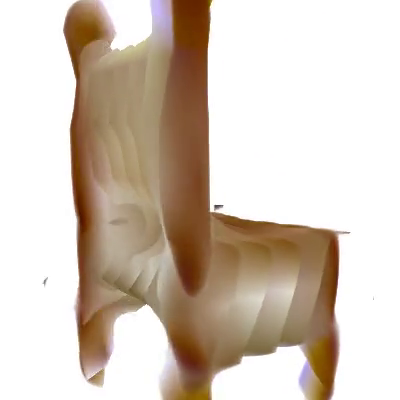}  & 
     \includegraphics[trim= 0.0cm 0cm 0cm 0cm,clip, width=0.2\linewidth]{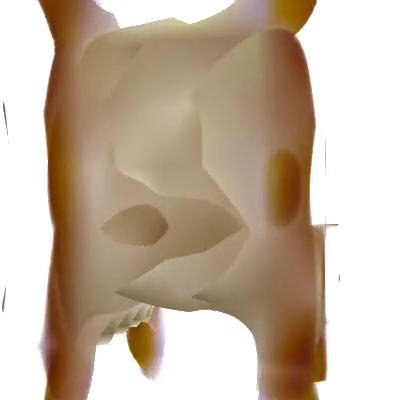}  & 
     \includegraphics[trim= 0.0cm 0cm 0cm 0cm,clip, width=0.2\linewidth]{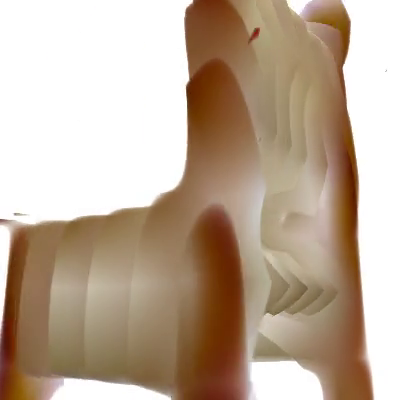}   \\
     \includegraphics[trim= 0.0cm 0cm 0cm 0cm,clip, width=0.2\linewidth]{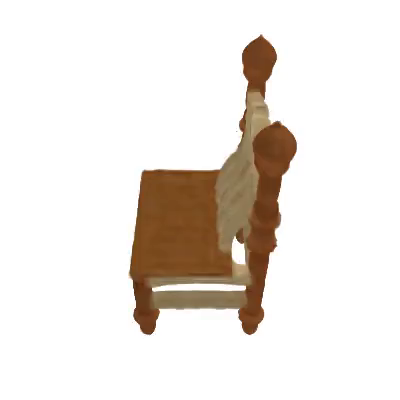}  &
     \includegraphics[trim= 0.0cm 0cm 0cm 0cm,clip, width=0.2\linewidth]{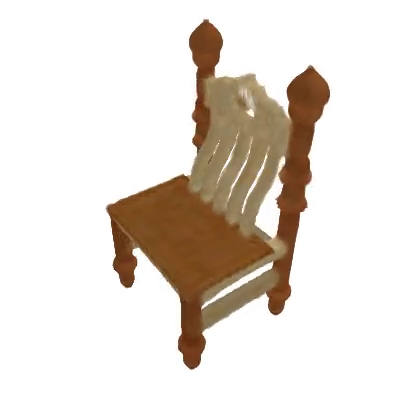}  &
     \includegraphics[trim= 0.0cm 0cm 0cm 0cm,clip, width=0.2\linewidth]{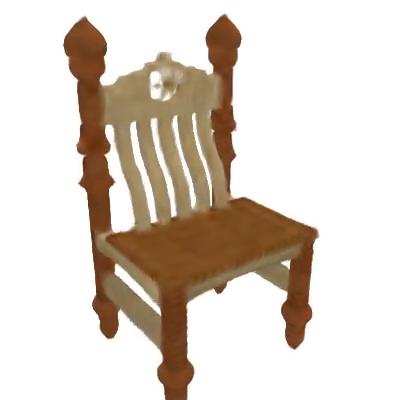}  &
     \includegraphics[trim= 0.0cm 0cm 0cm 0cm,clip, width=0.2\linewidth]{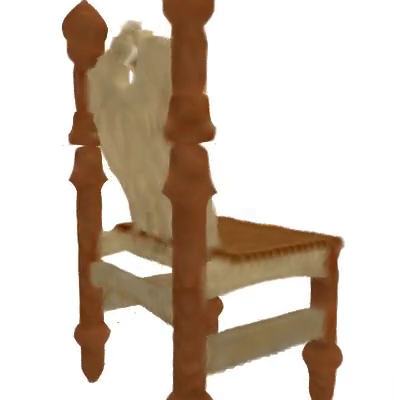}  & 
     \includegraphics[trim= 0.0cm 0cm 0cm 0cm,clip, width=0.2\linewidth]{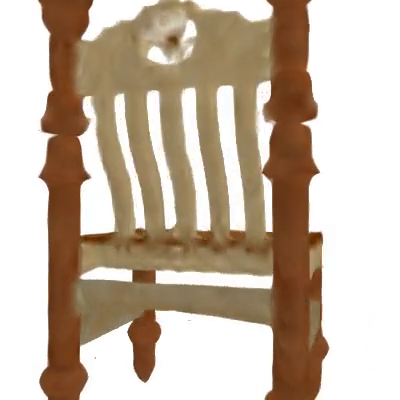}  & 
     \includegraphics[trim= 0.0cm 0cm 0cm 0cm,clip, width=0.2\linewidth]{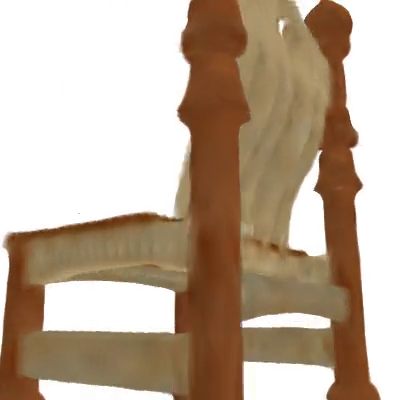}   \\
     \includegraphics[trim= 0.0cm 0cm 0cm 0cm,clip, width=0.2\linewidth]{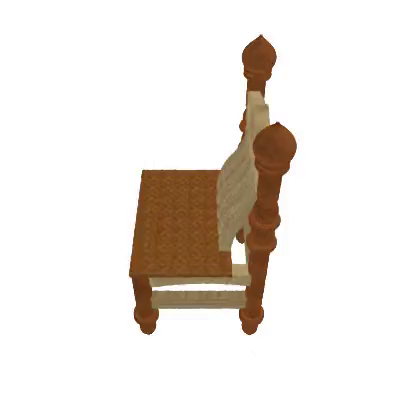}  &
     \includegraphics[trim= 0.0cm 0cm 0cm 0cm,clip, width=0.2\linewidth]{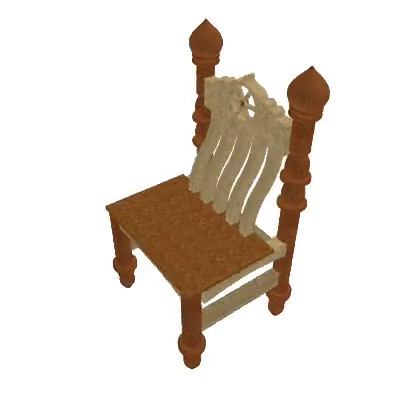}  &
     \includegraphics[trim= 0.0cm 0cm 0cm 0cm,clip, width=0.2\linewidth]{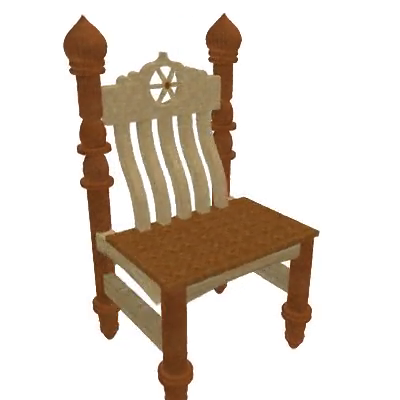}  &
     \includegraphics[trim= 0.0cm 0cm 0cm 0cm,clip, width=0.2\linewidth]{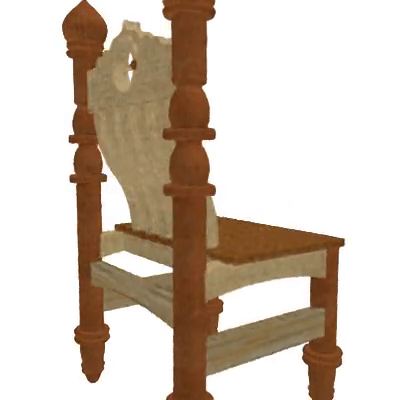}  & 
     \includegraphics[trim= 0.0cm 0cm 0cm 0cm,clip, width=0.2\linewidth]{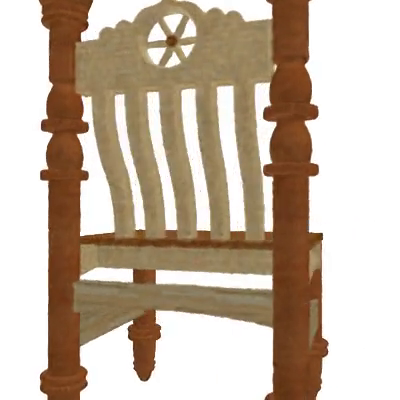}  & 
     \includegraphics[trim= 0.0cm 0cm 0cm 0cm,clip, width=0.2\linewidth]{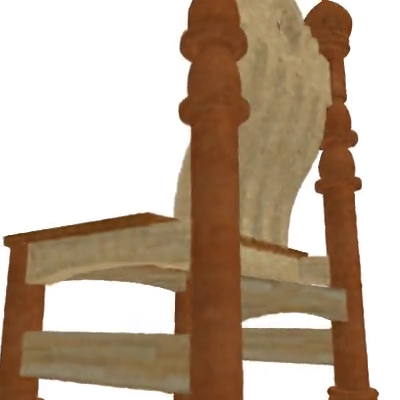}   \\

     \hline
     \includegraphics[trim= 0.0cm 0cm 0cm 0cm,clip, width=0.2\linewidth]{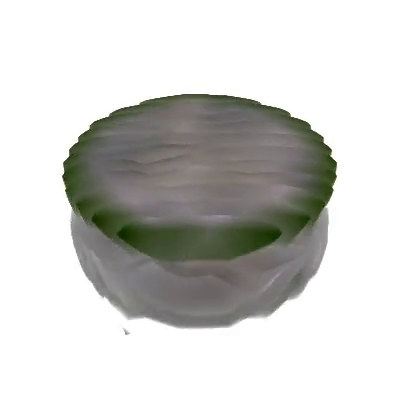}  &
     \includegraphics[trim= 0.0cm 0cm 0cm 0cm,clip, width=0.2\linewidth]{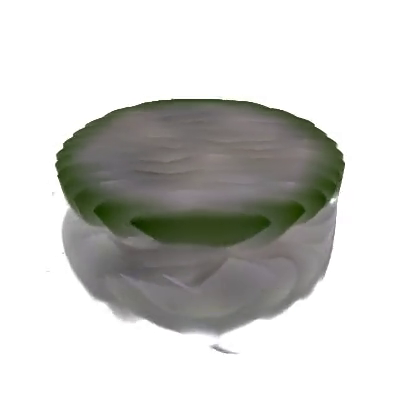}  &
     \includegraphics[trim= 0.0cm 0cm 0cm 0cm,clip, width=0.2\linewidth]{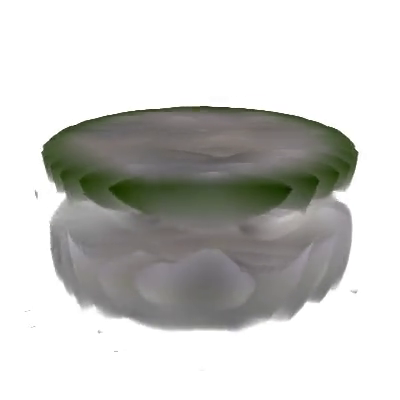}  &
     \includegraphics[trim= 0.0cm 0cm 0cm 0cm,clip, width=0.2\linewidth]{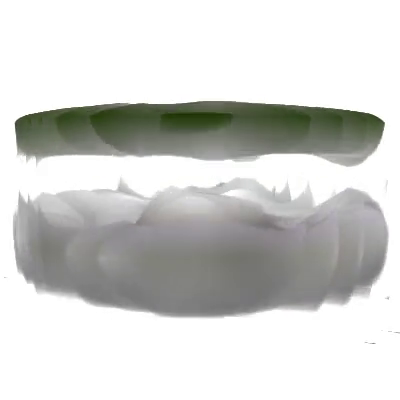}  &
     \includegraphics[trim= 0.0cm 0cm 0cm 0cm,clip, width=0.2\linewidth]{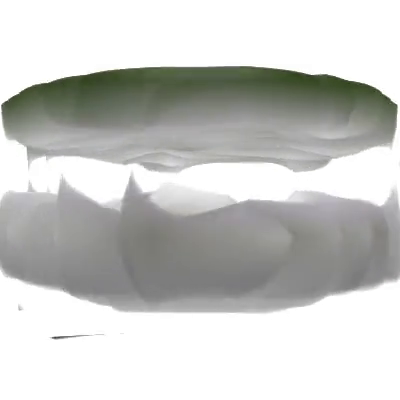}  & 
     \includegraphics[trim= 0.0cm 0cm 0cm 0cm,clip, width=0.2\linewidth]{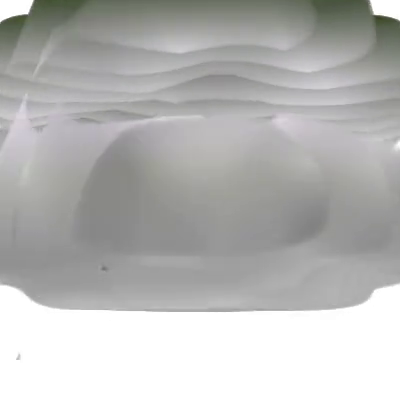}   \\
     \includegraphics[trim= 0.0cm 0cm 0cm 0cm,clip, width=0.2\linewidth]{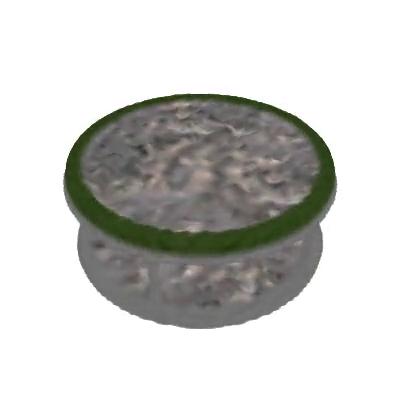}  &
     \includegraphics[trim= 0.0cm 0cm 0cm 0cm,clip, width=0.2\linewidth]{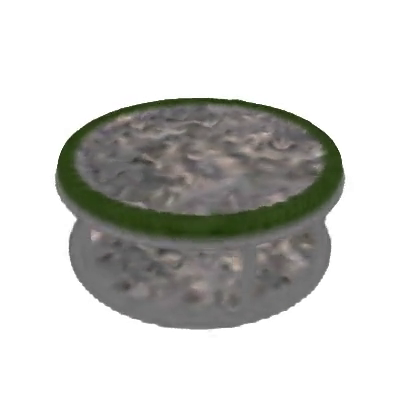}  &
     \includegraphics[trim= 0.0cm 0cm 0cm 0cm,clip, width=0.2\linewidth]{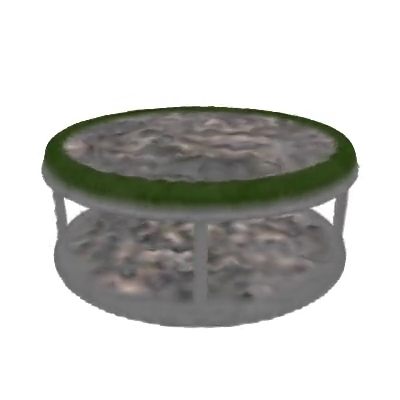}  &
     \includegraphics[trim= 0.0cm 0cm 0cm 0cm,clip, width=0.2\linewidth]{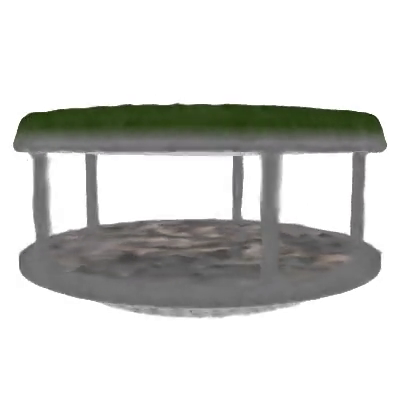}  &
     \includegraphics[trim= 0.0cm 0cm 0cm 0cm,clip, width=0.2\linewidth]{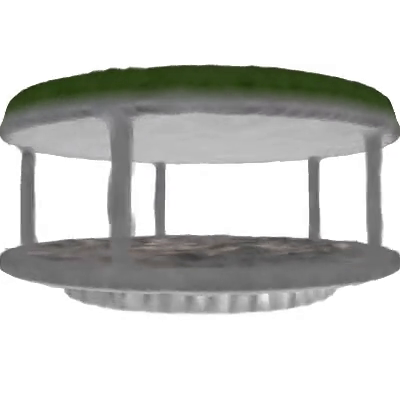}  & 
     \includegraphics[trim= 0.0cm 0cm 0cm 0cm,clip, width=0.2\linewidth]{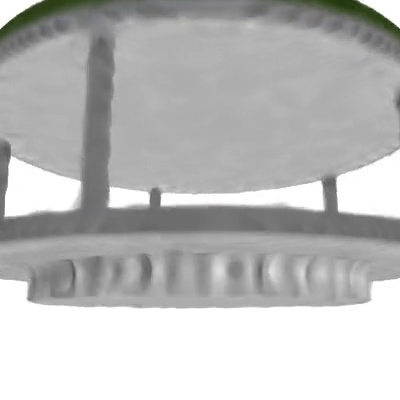}   \\
     \includegraphics[trim= 0.0cm 0cm 0cm 0cm,clip, width=0.2\linewidth]{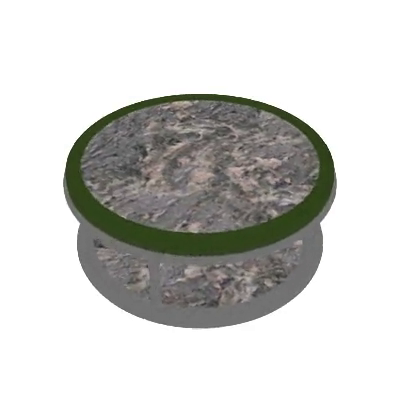}  &
     \includegraphics[trim= 0.0cm 0cm 0cm 0cm,clip, width=0.2\linewidth]{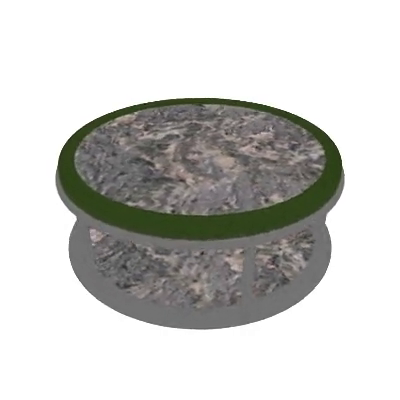}  &
     \includegraphics[trim= 0.0cm 0cm 0cm 0cm,clip, width=0.2\linewidth]{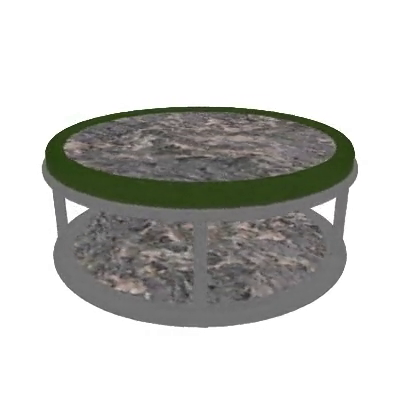}  &
     \includegraphics[trim= 0.0cm 0cm 0cm 0cm,clip, width=0.2\linewidth]{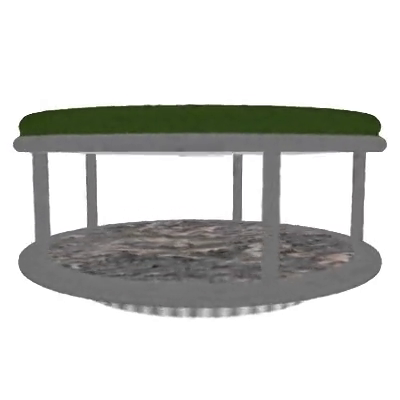}  &
     \includegraphics[trim= 0.0cm 0cm 0cm 0cm,clip, width=0.2\linewidth]{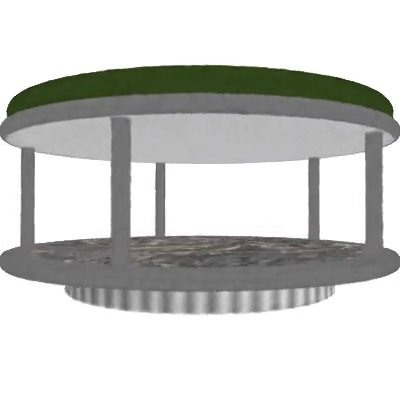}  & 
     \includegraphics[trim= 0.0cm 0cm 0cm 0cm,clip, width=0.2\linewidth]{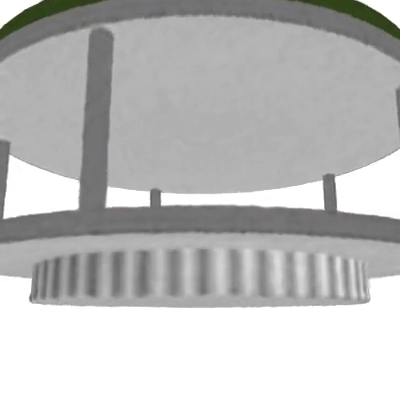}   \\
\bottomrule
\end{tabular}
} %
    \caption{\textbf{SRFs: The optimized Sparse Radiance Fields in SPARF 2}.  A total of one million SRFs have been collected in SPARF, including on multiple voxel resolutions: 32 (\textit{\textit{top}}), 128 (\textit{middle}), and 512 (\textit{bottom}) for every 3D shape.
    }
    \label{fig:srf2}
\end{figure*}


\end{document}